\documentclass[lettersize,journal]{IEEEtran}
\usepackage{ifpdf}
\ifCLASSOPTIONcompsoc
  \usepackage[nocompress]{cite} 
\else
  \usepackage{cite}
\fi
\usepackage{hyperref}
\usepackage{xspace}
\usepackage{graphicx} \graphicspath{{figures/}} 
\usepackage{amsmath,amssymb,nicefrac,bbm,pifont}
\usepackage{cite}
\usepackage{tcolorbox}
\usepackage[linesnumbered,ruled,vlined]{algorithm2e}
\usepackage{acronym}
\usepackage[skip=3pt,font=small]{subcaption}
\usepackage[skip=3pt,font=small]{caption}
\usepackage[dvipsnames,svgnames,x11names,table]{xcolor}
\usepackage[capitalise,noabbrev,nameinlink]{cleveref}
\usepackage{booktabs,tabularx,colortbl,multirow,multicol,array,makecell,tabularray}
\usepackage{overpic,wrapfig}
\usepackage[misc]{ifsym}
\usepackage{enumitem}
\usepackage{orcidlink}
\usepackage{dblfloatfix}
\usepackage{siunitx}
\usepackage{xr}
\usepackage{array}
\usepackage{dsfont}
\usepackage{diagbox}

\usepackage{setspace}
\definecolor{gblue}{HTML}{4285F4}
\definecolor{gred}{HTML}{DB4437}
\definecolor{ggreen}{HTML}{0F9D58}
\definecolor{ggrey}{HTML}{E9E9E9}
\definecolor{lightpurple}{HTML}{d8c7e5}
\definecolor{darkpurple}{HTML}{b18dcc}
\definecolor{gbest}{HTML}{FFFFFF}
\definecolor{gsecond}{HTML}{FFFFFF}
\definecolor{pinkred}{HTML}{ea99a5}
\definecolor{pinkblue}{HTML}{96b4e2}
\definecolor{SkyBlue}{RGB}{135, 206, 235}

\crefname{algorithm}{Alg.}{Algs.}
\Crefname{algocf}{Algorithm}{Algorithms}
\crefname{section}{Sec.}{Secs.}
\Crefname{section}{Section}{Sections}
\crefname{table}{Tab.}{Tabs.}
\Crefname{table}{Table}{Tables}
\crefname{figure}{Fig.}{Figs.}
\Crefname{figure}{Figure}{Figures}
\crefname{equation}{Eq.}{Eqs.}
\Crefname{equation}{Equation}{Equations}
\crefname{appendix}{Appx.}{Appxs.}
\Crefname{appendix}{Appendix}{Appendices}

\makeatletter
\DeclareRobustCommand\onedot{\futurelet\@let@token\@onedot}
\def\@onedot{\ifx\@let@token.\else.\null\fi\xspace}
\def\eg{\emph{e.g}\onedot} 

\def\ie{\emph{i.e}\onedot}

\def\etal{\emph{et al}\onedot}
\makeatother

\newcommand{\affordanceModelName}{Afford-X}
\newcommand{\affordanceDatasetName}{COCO-Aff}
\newcommand{\affordanceDatasetNamelvis}{LVIS-Aff}
\acrodef{ai}[AI]{Artificial Intelligence}
\acrodef{llm}[LLM]{Large Language Model}
\acrodef{vlm}[VLM]{Visual Language Model}
\acrodef{mllm}[MLLM]{Multimodal Large Language Model}
\acrodef{mm}[MM]{Multimodal Model}
\acrodef{tamp}[TAMP]{Task-and-Motion Planning}
\acrodef{tod}[TOD]{Task-Oriented Detection}
\acrodef{detr}[DETR]{DEtection TRansformer}
\acrodef{mdetr}[MDETR]{Multimodal DEtection TRansformer}
\acrodef{gnn}[GNN]{Graph Neural Networks}
\acrodef{fast-rcnn}[Fast-RCNN]{Fast Region-Based Convolutional Neural Networks}
\acrodef{slam}[SLAM]{Simultaneous Localization And Mapping}
\acrodef{ioar}[IOAR]{Instance-Oriented Affordance Reasoning}
\acrodef{toar}[TOAR]{Task-Oriented Affordance Reasoning}
\acrodef{tom}[TOM]{Task-Oriented Manipulation}
\acrodef{cnn}[CNN]{Convolutional Neural Network}
\acrodef{kl}[KL]{Kullback-Leibler}
\acrodef{va}[VA]{Verb Attention}
\acrodef{bf}[BF]{Bi-Fusion}
\acrodef{sa}[SA]{Self-Attention}
\acrodef{ggnn}[GGNN]{Gated Graph Neural Networks}
\acrodef{cd}[CD]{Clustering Distillation}
\acrodef{ccr}[CCR]{Cluster Center Replacement}
\acrodef{sbtl}[SBTL]{Soft Binary Target Loss}
\acrodef{cl}[CL]{Cluster Loss}
\acrodef{fps}[FPS]{Frames Per Second}
\begin{document}

\title{\affordanceModelName{}: Generalizable and Slim Affordance Reasoning for Task-oriented Manipulation}

\author{%
    Xiaomeng~Zhu\textsuperscript{*}\,\orcidlink{0000-0002-0704-6314},~
    Yuyang~Li\textsuperscript{*}\,\orcidlink{0000-0002-5794-7997},~
    Leiyao~Cui\,\orcidlink{0009-0009-4925-6983},~
    Pengfei~Li\,\orcidlink{0009-0009-8114-6207},\\
    Huan-ang~Gao\,\orcidlink{0009-0004-6727-5778},~
    Yixin~Zhu~\textsuperscript{\Letter}\,\orcidlink{0000-0001-7024-1545},~
    and~Hao~Zhao~\textsuperscript{\Letter}\,\orcidlink{0000-0001-7903-581X}\\
\thanks{
Manuscript received February 23, 2025; revised June 16, 2026; accepted August 3, 2026.
This work is supported in part by the Brain Science and Brain-like Intelligence Technology---National Science and Technology Major Project (2025ZD0219400), the National Natural Science Foundation of China (62376009), the Beijing Nova program, the NVIDIA Academic Grant Program using Spark and Thor, the State Key Lab of General AI at Peking University, the PKU-BingJi Joint Laboratory for Artificial Intelligence, the Wuhan Major Scientific and Technological Special Program (2025060902020304), the Hubei Embodied Intelligence Foundation Model Research and Development Program, and the National Comprehensive Experimental Base for Governance of Intelligent Society, Wuhan East Lake High-Tech Development Zone.
\textit{(Corresponding authors: Yixin Zhu; Hao Zhao.)}}
\thanks{Xiaomeng Zhu, Yuyang Li, Leiyao Cui, and Yixin Zhu are with Peking University, Beijing 100871, China (email: xiaomeng.zhu@connect.ust.hk, y.li@stu.pku.edu.cn, cuileiyao24@mails.ucas.ac.cn, yixin.zhu@pku.edu.cn).}
\thanks{Xiaomeng Zhu is also with the Department of Computer Science and Engineering, Hong Kong University of Science and Technology, Hong Kong 999077, China.}
\thanks{Leiyao Cui is also with Shenyang Institute of Automation, Chinese Academy of Sciences, Shenyang 110016, China.}
\thanks{Pengfei Li and Hao Zhao are with the Institute for AI Industry Research, Tsinghua University, Beijing 100084, China (li-pf22@mails.tsinghua.edu.cn, zhaohao@air.tsinghua.edu.cn).} 
\thanks{Huan-ang Gao is with the Department of Computer Science, Tsinghua University, Beijing 100084, China (gha24@mails.tsinghua.edu.cn).}
\thanks{Yixin Zhu is also with Beijing Key Laboratory of Brain-Computer Interface and Mental Health Modulation, and XS Vision.}
\thanks{Implementable code is available at: \url{https://zhuxmmm.github.io/Afford-X}}
\thanks{\textsuperscript{*} Equal contribution.
}}

\maketitle
\begin{abstract}
Object affordance reasoning, the ability to infer object functionalities based on physical properties, is fundamental for task-oriented planning and activities in both humans and \ac{ai}. This capability, required for planning and executing daily activities in a task-oriented manner, relies on commonsense knowledge of object physics and functionalities, extending beyond simple object recognition.
Current computational models for affordance reasoning from perception lack generalizability, limiting their applicability in novel scenarios. Meanwhile, comprehensive \acp{llm} with emerging reasoning capabilities are challenging to deploy on local devices for task-oriented manipulations.
Here, we introduce \affordanceDatasetNamelvis{}, a large-scale dataset comprising 1,496 tasks and 119k images, designed to enhance the generalizability of affordance reasoning from perception.
Utilizing this dataset, we develop \affordanceModelName{}, an end-to-end trainable affordance reasoning model that incorporates Verb Attention and Bi-Fusion modules to improve multi-modal understanding. This model achieves up to a 12.1\% performance improvement over the best-reported results from non-\ac{llm} methods, while also demonstrating a 1.2\% enhancement compared to our previous conference paper. Additionally, it maintains a compact 187M parameter size and infers nearly 50 times faster than the GPT-4V API.
Our work demonstrates the potential for efficient, generalizable affordance reasoning models that can be deployed on local devices for task-oriented manipulations. {We showcase \affordanceModelName{}'s effectiveness in enabling task-oriented object grasping for robots across various tasks and environments, underscoring its efficiency and potential for manipulation tasks in physical environments.}
\end{abstract}

\section{Introduction}\label{sec:introduction}

\begin{figure*}[t!]
    \centering
    \includegraphics[width=\linewidth]{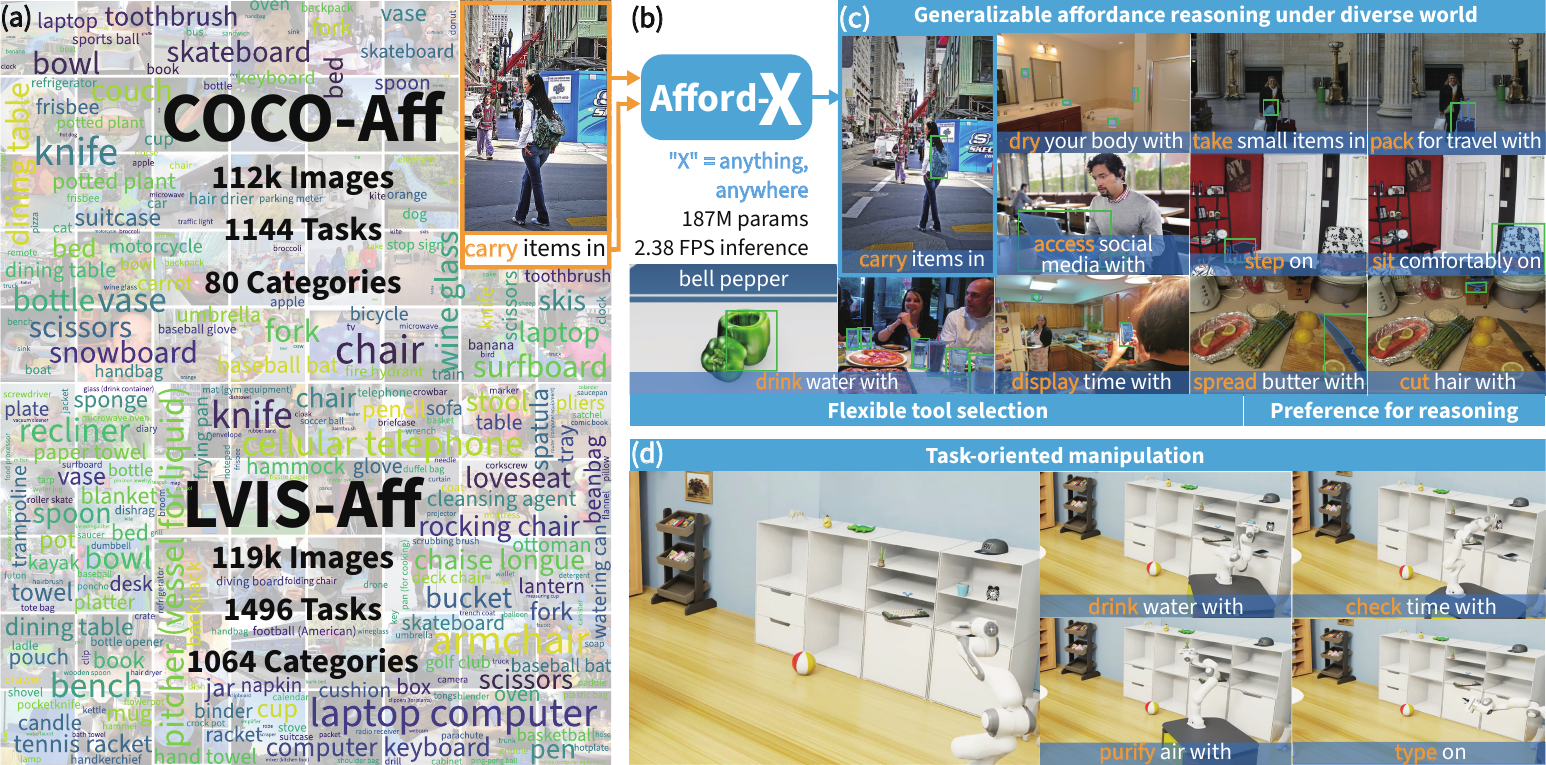}
    \caption{\textbf{Affordance reasoning for task-oriented manipulation.} \affordanceModelName{} provides efficient visual affordance reasoning through: (a) two comprehensive datasets---\affordanceDatasetName{} (112k images, 1,144 tasks, 80 categories) and \affordanceDatasetNamelvis{} (119k images, 1,496 tasks, 1,064 categories); (b) real-time processing (2.38 \ac{fps}) with a compact 187M-parameter architecture generating bounding boxes and object masks; (c) robust generalization demonstrated through task-specific object selection and multi-object identification at 0.7 confidence threshold; (d) integration with robotic systems for simulated task-oriented object grasping.}
    \label{fig:teaser}
\end{figure*}

\IEEEPARstart{E}{ffective} interaction with the world demands more than object recognition; it requires understanding how objects can be used. This concept, known as \textit{affordance reasoning}~\cite{gibson1979theory,zhu2020dark}, transcends the traditional ``what is where'' paradigm~\cite{marr2010vision} of object detection and classification systems~\cite{kamath2021mdetr,jia2021scaling,radford2021learning,li2022grounded,liu2023grounding}. Through affordance reasoning, agents infer potential functions from physical properties---a fundamental capability that enables both task-oriented manipulation and adaptive problem-solving in complex environments~\cite{qin2023robot,wu2024learning}.

This reasoning capability enables humans to naturally select appropriate tools for specific tasks~\cite{vaesen2012cognitive} and devise creative solutions in unfamiliar or resource-constrained environments~\cite{mccormack2011tool}. Consider, for example, repurposing a hollowed-out bell pepper as a water container when conventional containers are unavailable, as shown in \cref{fig:teaser}(c). Such adaptability, rooted in understanding object properties and their potential functions, exemplifies the flexibility required for effective interaction across diverse environments~\cite{zhu2015understanding}.

\begin{figure*}[t!]
    \centering
    \begin{subfigure}[b]{.499\linewidth}
        \centering
        \includegraphics[width=\linewidth]{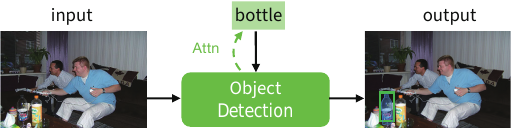}
        \caption{Object detection task: identifying the ``\textit{bottle}'' from visual input.}
        \label{fig:affordance_vs_detection_a}
    \end{subfigure}%
    \hfill %
    \begin{subfigure}[b]{.499\linewidth}
        \centering
        \includegraphics[width=\linewidth]{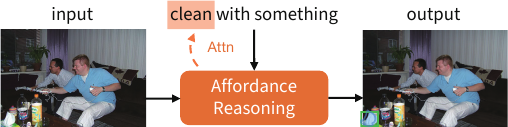}
        \caption{Affordance reasoning task: selecting a tool for ``\textit{cleaning}''.}
        \label{fig:affordance_vs_detection_b}
    \end{subfigure}%
    \\%
    \begin{subfigure}[b]{.499\linewidth}
        \centering
        \includegraphics[width=\linewidth]{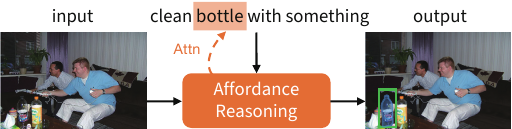}
        \caption{Traditional methods incorrectly focus on the ``\textit{bottle}'' when processing ``\textit{clean the bottle with something}''.}
        \label{fig:affordance_vs_detection_c}
    \end{subfigure}%
    \hfill %
    \begin{subfigure}[b]{.499\linewidth}
        \centering
        \includegraphics[width=\linewidth]{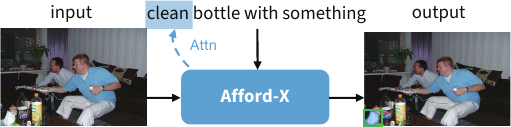}
        \caption{Our model correctly identifies the ``\textit{napkin}'' as the cleaning tool, despite the prominent bottle in the scene.}
        \label{fig:affordance_vs_detection_d}
    \end{subfigure}%
    \caption{\textbf{Comparison between object detection and affordance reasoning.}
    In our task-oriented setting, affordance reasoning aims to identify the object that best supports the intended action, rather than simply localizing the noun mentioned in the instruction.
    (a) Conventional object detection localizes the object specified by a noun query, \eg, ``\textit{bottle}''.
    (b) Affordance reasoning selects an object that can serve a functional role for a task, \eg, choosing a tool for ``\textit{clean with something}''.
    (c) Existing affordance methods may over-attend to the explicit noun in the instruction and incorrectly select the prominent ``\textit{bottle}'' for ``\textit{clean bottle with something}''.
    (d) \affordanceModelName{} emphasizes the task verb and correctly identifies the ``\textit{napkin}'' as the appropriate cleaning tool, despite the visually salient bottle in the scene.
    }
    \label{fig:affordance_vs_detection}
\end{figure*}

The importance of affordance reasoning extends beyond human cognition into \ac{ai} and robotics~\cite{hassanin2021visual,zhu2020dark}. In task-oriented manipulation~\cite{li2022toist,tang2023cotdet,qu2024rio,fang2020learning,wang2023task}, agents must process both task requirements (textual input) and environmental perception (visual input) to select and use appropriate objects for specific goals. This process requires reasoning about feature cues and matching them to novel task contexts~\cite{zhang2022understanding}, as illustrated in \cref{fig:affordance_vs_detection_a,fig:affordance_vs_detection_b}. Developing robust affordance reasoning capabilities could significantly enhance \ac{ai} systems' ability to operate flexibly in complex, real-world environments~\cite{allen2020rapid}.

Despite the significant benefits of affordance reasoning, its computational implementation faces several key challenges, particularly in designing frameworks suitable for local deployment and offline processing. Robotic platforms typically operate under strict computational constraints---whether using NVIDIA's Jetson Orin development board or even high-end RTX 4090 GPUs with 24GB memory~\cite{suder2023power}. These limitations prevent the deployment of large-scale pre-trained generative \acp{mllm}~\cite{achiam2023gpt} due to computing capabilities, power constraints, usage policies, and information security concerns~\cite{lera2017cybersecurity,mseer2023artificial}. While smaller pre-trained generative \acp{mllm} like SPHINX 1.1B~\cite{lin2023sphinx} can operate locally, their limited knowledge bases and reasoning capabilities prove insufficient for complex affordance reasoning. In contrast, pre-trained \acp{mm}~\cite{su2019vl,lu2019vilbert,li2020oscar,chen2020uniter,lu202012,kamath2021mdetr} achieve superior performance with fewer parameters and faster response times, likely because they leverage knowledge directly from the image feature space, capturing fine-grained visual details essential for affordance reasoning~\cite{qian2024affordancellm}.

Training these slim \acp{mm} presents additional challenges, particularly when pre-training techniques ignore the fundamental nature of affordance. Models trained directly on object detection datasets often develop biased understanding, over-emphasizing familiar nouns. This bias can lead to failures when task descriptions include prominently visible objects, as the model may misinterpret an affordance reasoning task as simple object detection. For example, given the task ``\textit{clean bottle with},'' a model might fixate on detecting ``\textit{bottle}'' while missing the critical action ``\textit{clean},'' failing to understand the task's true intent (see \cref{fig:affordance_vs_detection_c,fig:affordance_vs_detection_d}). Moreover, some pre-trained \acp{mm}, such as \ac{mdetr}~\cite{kamath2021mdetr}, rely on simple concatenation of visual and language features, potentially limiting their comprehension of multimodal data.

\begin{figure}[t!]
    \centering
    \includegraphics[width=\linewidth]{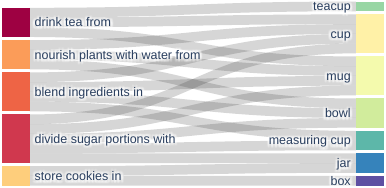}
    \caption{\textbf{Task-object mapping in affordance reasoning.} Affordance reasoning involves complex many-to-many relationships between tasks and objects. (a) We visualize task descriptions from \affordanceDatasetNamelvis{} on the left and their compatible object categories on the right. (b) The connections demonstrate how individual tasks can be accomplished with multiple objects. (c) The mapping reveals how single objects can serve multiple different tasks, highlighting the complexity of affordance relationships.}
    \label{fig:mapping_relation}
\end{figure}

Developing slim models with accurate and generalized affordance understanding requires both a large-scale corpus and diverse knowledge representation~\cite{wang2023large}. This diversity depends on three critical dimensions: the range of tasks, the quantity of images, and the variety of target object categories. Deficiencies in any of these dimensions can limit a robot's ability to make contextual decisions when encountering dynamic affordances, diverse layouts, and novel objects~\cite{qian2024affordancellm,liu2024moka,zhu2020dark}. However, expanding this knowledge base presents unique challenges due to the complex many-to-many mapping between affordance tasks and target objects---unlike the straightforward one-to-one relationships in object detection~\cite{lin2014coco,gupta2019lvis}. A single object can serve multiple tasks, and conversely, one task might be accomplished using various objects (see \cref{fig:mapping_relation})~\cite{xin2022cocotasks}. Despite previous efforts involving manual annotations~\cite{zhou2017scene,chuang2018adeaffordance,xin2022cocotasks} or \acp{llm}, creating truly diverse affordance reasoning datasets remains a significant challenge~\cite{qu2024rio}.

To address these challenges, we propose \affordanceModelName{}, a slim end-to-end multimodal reasoning framework built on a knowledge distillation architecture inspired by TOIST~\cite{li2022toist}. Our framework consists of paired teacher and student models sharing the same architecture. It processes text-based task descriptions and visual scene inputs to autonomously identify appropriate objects for given tasks, producing both detection bounding boxes and fine-grained segmentation masks (see \cref{fig:teaser}(b)). The training process occurs in two stages: first, the teacher model learns in an oracle manner using target object category labels in the text input; then, this knowledge transfers to the student model, which operates without such labels. This distillation approach proves particularly effective for scenarios with size constraints, as the student model's architecture can adapt to practical requirements while maintaining effective supervision from the teacher.

To enhance our framework's capabilities, we introduce two key modules. The \ac{va} module emphasizes action words in the input, ensuring accurate task understanding rather than mere object recognition. The \ac{bf} module, inspired by BLIP~\cite{li2022blip}, improves upon simple feature concatenation methods to better capture complex interactions between visual and textual information. Our experimental results (see \cref{fig:combined_metrics}) demonstrate that \affordanceModelName{}, despite its compact 187M parameters, outperforms even GPT-4V in affordance reasoning tasks while achieving real-time inference speeds of 2.38 \ac{fps}---making it practical for real-world deployment.

To strengthen \affordanceModelName{}'s generalization capabilities, we developed an automated pipeline that leverages \acp{llm} to convert object detection datasets into affordance reasoning datasets. This pipeline employs GPT-4 in dual roles: as a producer generating task-object pairs from object categories, and as a quality inspector filtering out errors and inconsistencies. Using this approach, we created two comprehensive datasets: \affordanceDatasetName{} from COCO2014~\cite{lin2014coco} and \affordanceDatasetNamelvis{} from LVIS~\cite{gupta2019lvis}. \affordanceDatasetName{} features 1,144 diverse tasks, 112k training images, and 80 object categories, while \affordanceDatasetNamelvis{} expands to 1,496 tasks, 119k images, and 1,064 object categories, offering broader coverage of both indoor and outdoor scenarios (see \cref{fig:teaser}(a)). Models trained on these datasets show significant improvements in generalization, with accuracy gains of 22.9\% and 24.7\% respectively on unseen tasks.

We validate these improvements through extensive testing of \affordanceModelName{}'s ability to support embodied agents in diverse simulated environments. Beyond evaluations on natural images from datasets like COCO~\cite{lin2014coco}, we utilize textured meshes from Objaverse~\cite{deitke2023objaverse} and OmniGibson~\cite{li2023behavior} to create scenes with diverse, randomly placed objects rendered using photorealistic ray-tracing, as shown in \cref{fig:teaser}(d). This approach enables assessment of our model's robustness to complex object geometry, appearance variations, and challenging environmental conditions including clustered objects, varying lighting, and visual distractions. We further demonstrate practical applicability through simulated object collection tasks, where \affordanceModelName{} integrates with standard grasp planners and motion planners on a mobile manipulator to perceive scenes, select appropriate objects, and execute retrieval actions.

This article significantly extends our previous TOIST work~\cite{li2022toist} through several key contributions:
\begin{itemize}[nolistsep,leftmargin=*]
    \item Development of knowledge distillation-based \affordanceModelName{} with innovative \ac{va} and \ac{bf} modules, enhancing action recognition and multimodal interpretation capabilities
    \item Creation of comprehensive affordance reasoning datasets: \affordanceDatasetName{} and its expanded version \affordanceDatasetNamelvis{}
    \item Extensive validation demonstrating significant performance improvements over our previous approach through both image datasets and simulated environments
\end{itemize}

The paper continues with a comprehensive review of related work (\cref{sec:related}), followed by detailed descriptions of our dataset construction (\cref{sec:dataset}), model architecture (\cref{sec:method}), and embodied affordance reasoning approach (\cref{sec:embodied}). We present experimental results (\cref{sec:experiment}), discuss task-oriented manipulation findings (\cref{sec:manipulation}), and conclude with key insights (\cref{sec:conclusion}).

\section{Related Work}\label{sec:related}

In this section, we review three key aspects of affordance research: learning-based affordance reasoning (\cref{sec:rw:affordance_reasoning}), datasets for training and evaluation (\cref{sec:rw:dataset}), and task-oriented manipulation approaches (\cref{sec:rw:tom}).

\subsection{Learning-based Affordance Reasoning}\label{sec:rw:affordance_reasoning}

The concept of affordance, introduced by American psychologist James J. Gibson~\cite{gibson1979theory}, proposes that environmental objects inherently offer action possibilities---for instance, a chair affords sitting. This fundamental notion, which emphasizes the relationship between physical properties and potential functions, has become central to how robots understand and interact with their environment~\cite{zhu2020dark}.

As affordance research entered the computer vision domain, early approaches focused on establishing direct mappings between visual features and potential actions, bypassing explicit object recognition~\cite{nguyen2016detecting}. These traditional methods combined handcrafted features---including shape, size, texture, color, and material---with Bayesian networks~\cite{friedman1997bayesian} or support vector machines~\cite{noble2006support} to encode relationships between geometric features and affordances~\cite{lopes2007affordance,montesano2008learning,uugur2010traversability}. However, these approaches struggled to generalize across diverse object appearances and environmental contexts~\cite{chen2023survey}.

The emergence of deep learning transformed affordance reasoning through its powerful feature extraction capabilities. These models leverage multi-layer networks and large-scale datasets to capture rich object features and learn affordance cues, achieving enhanced performance and robustness~\cite{do2018affordancenet}. Their success in identifying functional regions---such as graspable or supportable areas---has significantly advanced robotic task-oriented manipulation~\cite{nguyen2016detecting,levine2016end,gupta2017cognitive,do2018affordancenet}. However, challenges persist in capturing contextual dependencies and complex semantic relationships, particularly in cluttered scenes with multiple interacting objects.

The integration of semantic information with visual cues marked the next major advance. Modern models leverage category-level knowledge for object function inference, enabling more effective task-oriented object detection~\cite{sawatzky2019object,li2022toist}. This enhanced contextual understanding provides deeper insights into object-environment interactions~\cite{zhu2014reasoning,kamath2021mdetr,li2022toist,nguyen2023open,van2024open}. Transformer-based architectures, especially vision-language pre-training models~\cite{su2019vl,lu2019vilbert,li2020oscar,chen2020uniter,lu202012,kamath2021mdetr}, have further advanced the field through sophisticated cross-modal alignment techniques. These approaches demonstrate superior flexibility and generalizability compared to traditional \ac{cnn}-based methods, particularly in complex and dynamic scenarios.

Most recently, the rich commonsense knowledge embedded in \acp{llm} has opened new possibilities for affordance reasoning. CoTDet demonstrates this potential through structured task decomposition~\cite{wei2022chain}, while AffordanceLLM enhances open-world inference by combining visual perception with \ac{llm} capabilities~\cite{qian2024affordancellm}. {Along this line, recent works further exploit foundation models and 3D cues. OOAL learns open affordances from a single example with foundation models\cite{li2024one}, and both 3D-AffordanceLLM\cite{chu20253d} and GEAL~\cite{lu2025geal} push open-vocabulary affordance reasoning into 3D worlds.} However, deploying large-scale \acp{llm} like GPT-4 presents significant challenges for robotic platforms that require local, offline inference. While small-scale \acp{llm} offer an alternative, they struggle with limited knowledge bases and reasoning capabilities---particularly in sparse language spaces compared to dense image feature spaces~\cite{xu2022partafford}. To address these limitations, we propose \affordanceModelName{}, a \ac{mm}-based end-to-end framework inspired by TOIST~\cite{li2022toist}. Our approach achieves efficient inference with compact parameters, enabling broader deployment across manipulation platforms.

\subsection{Dataset for Affordance Reasoning}\label{sec:rw:dataset}

The proliferation of deep learning in affordance reasoning has highlighted datasets as a critical foundation for model development. A dataset's effectiveness depends on three key dimensions: the diversity of tasks, the quantity of images, and the range of object categories. \cref{tab:dataset} summarizes the major datasets in this field. The first significant milestone came from Myers \etal~\cite{myers2015affordance}, who introduced both a framework for joint affordance localization and recognition and the field's first pixel-level annotated dataset. While groundbreaking, this initial dataset focused primarily on surface features, overlooking the crucial role of human-object interactions in affordance reasoning. Chuang \etal~\cite{chuang2018adeaffordance} addressed this limitation with the ADE-Affordance dataset, built upon ADE20K~\cite{zhou2017scene}, incorporating both physical constraints and social norms to better align with real-world reasoning challenges.

\begin{table}[ht!]
    \centering
    \small
    \setlength{\tabcolsep}{3pt}
    \caption{\textbf{Comparison of affordance detection datasets.} We analyze the key characteristics of major datasets in affordance reasoning. (a) We evaluate datasets based on their number of images (\#Imgs), object categories (\#Cats), and affordance/task categories (\#Aff). (b) Each dataset builds upon different source datasets, providing varying foundations for affordance learning. (c) Our proposed datasets, \affordanceDatasetName{} and \affordanceDatasetNamelvis{}, achieve significant expansion across all three dimensions compared to existing work.}
    \begin{tabular*}{\linewidth}{@{\extracolsep{\fill}}lcccc}
        \toprule 
        \textbf{Dataset} & \textbf{\#Imgs} & \textbf{\#Cats} & \textbf{\#Aff} & \textbf{Source} \\\hline
        ADE-Aff~\cite{chuang2018adeaffordance} & 1000 & 150 & 7 & ADE20K~\cite{zhou2017scene} \\
        PAD~\cite{luo2021one} & 4002 & 72 & 31 & \textbackslash \\
        PADv2~\cite{lu2022phrase}& 30000 & 103 & 39 & \textbackslash \\
        PAD-L~\cite{lu2022phrase}& 4002 & 72 & 31 & \textbackslash \\
        COCO-Tasks~\cite{xin2022cocotasks} & 39724 & 49 & 14 & COCO2024~\cite{lin2014coco} \\
        RIO~\cite{qu2024rio}  & 40214 & 69 & $>$100 & COCO2024 \\
        \hline
        \affordanceDatasetName{} & 112k & 80 & 1144 & COCO2024 \\
        \affordanceDatasetNamelvis{} & 119k & 1064 & 1496 & LVIS~\cite{gupta2019lvis} \\
        \bottomrule
    \end{tabular*}
    \label{tab:dataset}
\end{table}

\begin{figure*}[t!]
    \centering
    \includegraphics[width=\linewidth]{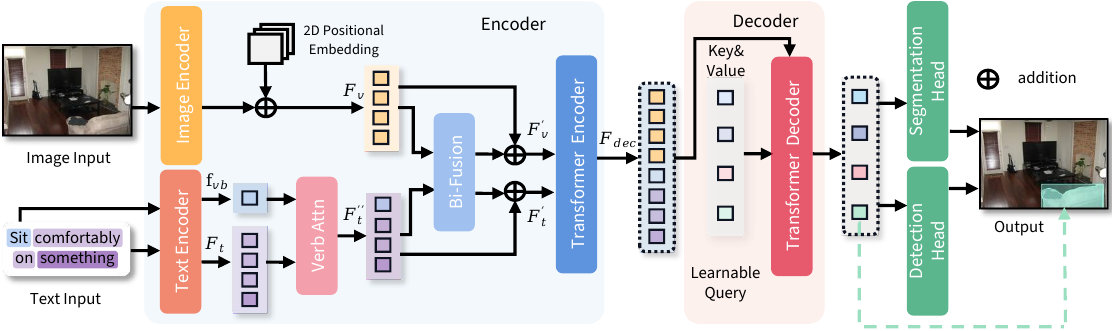}
    \caption{\textbf{Architecture of the \affordanceModelName{}.} Our framework processes visual and textual inputs through multiple specialized components. (a) The model begins by extracting visual features from images and textual features from prompts containing words like ``\textit{something}''. (b) The \textcolor{pinkblue}{Bi-Fusion module} performs bi-directional attention between visual and textual features to enhance multimodal understanding. (c) The \textcolor{pinkred}{Verb Attn module} strengthens the model's focus on action verbs to reduce category interference. (d) A Transformer encoder-decoder processes these enhanced features using learnable query vectors, producing parallel outputs for object detection and instance segmentation.}
    \label{fig:method_pipeline}
\end{figure*}

Recognizing that affordances fundamentally connect to human behavioral goals---reflecting Gibson's concept of animal-environment complementarity---researchers began developing goal-oriented datasets. Luo \etal pioneered this direction with PAD~\cite{luo2021one} and its successor PADv2~\cite{zhai2022one}, explicitly modeling the relationship between human goals and affordances while encompassing more complex scenarios. Lu \etal~\cite{lu2022phrase} further advanced this approach through PAD-L, which integrated natural language in affordance detection to enable object segmentation based on phrase-based affordance descriptions. However, their reliance on a limited affordance dictionary for paraphrasing constrained the capture of natural language complexity.

The field then shifted toward task-specific object selection. Sawatzky \etal~\cite{sawatzky2019object} introduced the COCO-Task dataset, derived from COCO~\cite{lin2014microsoft}, marking the first transformation of an object detection dataset into an affordance reasoning dataset. While innovative, its scope remained limited by predefined 14 tasks and rigid phrase representations. Qu \etal~\cite{qu2024rio} addressed these constraints with the RIO dataset, also built upon COCO2014, offering richer tasks and more diverse descriptions for broader scenario coverage. However, RIO's limited object categories prove insufficient for dynamic open-world environments, and despite leveraging \ac{llm} for task construction, it still requires extensive human annotation for task-object pair filtering.

To overcome these limitations, we propose an automated pipeline for converting object detection datasets into affordance reasoning datasets. Our approach employs \ac{llm} in dual roles---as both producer and inspector---significantly reducing the need for human annotation. Through this pipeline, we have created \affordanceDatasetName{} and \affordanceDatasetNamelvis{}, establishing new benchmarks in task coverage, image quantity, and object category diversity. These datasets provide models with a substantially richer knowledge base while maintaining data quality through automated verification, enabling more robust and generalizable affordance reasoning.

\subsection{Task-oriented Manipulation}\label{sec:rw:tom}

Ikeuchi and Hebert's seminal work~\cite{ikeuchi1996task} established that vision systems should adapt their architectures to specific tasks rather than pursuing a general-purpose approach. This task-oriented vision paradigm has become fundamental to robotic manipulation, enabling systems to handle diverse tasks posed by algorithms or human users. These tasks span a broad spectrum---from object manipulation with varying grasp types~\cite{ikeuchi1996task,miller2003automatic,vezzani2017grasping} and purposes~\cite{zhu2015understanding,zhu2020dark,wang2023task,song2015task} to complex environmental interactions such as door opening and water pouring~\cite{narayanan2015task,gong2023arnold,james2020rlbench}. At its core, task-oriented manipulation requires optimal system configuration through the selection of appropriate sensor signals~\cite{xia2022review}, task representations~\cite{miller2003automatic,vezzani2017grasping,zhu2015understanding}, processing modules, and manipulation policies~\cite{fang2020learning,wang2023task,agarwal2023dexterous,zhu2023toward}, conditioning on specific goals and target objects.

{Current task-oriented robotic vision and manipulation systems, however, typically operate under a significant constraint: they assume the suitable object for a task has already been specified and represented with proper visual representations (\eg, 3D mesh or point cloud)~\cite{nguyen2023open,li2023locate,zhang2025iaao,wu2025open}. This limitation reduces system autonomy by requiring human input for object selection. Our approach addresses this fundamental gap by integrating affordance reasoning capabilities within scene understanding~\cite{chuang2018learning}, enabling robots to identify appropriate objects for the given task within open-ended scenes with multiple candidate objects and obtain their 2.5D representations, \ie segmented colored point clouds, which are vital for many downstream manipulation tasks~\cite{nguyen2023open,li2023locate,zhang2025iaao,wu2025open}.} This integration represents a crucial step toward autonomous task-oriented manipulation systems that can adapt to diverse environments and task requirements.

\section{The \affordanceModelName{}}\label{sec:method}

This section presents our \affordanceModelName{} framework for affordance reasoning. We begin with a formal problem formulation (\cref{sec:method:formulation}), followed by the model architecture (\cref{sec:model_architecture}), noun-pronoun distillation strategy (\cref{sec:noun_pronoun_distillation}), and key architectural components (\cref{sec:method:modules}).

\subsection{Problem Formulation}\label{sec:method:formulation}

Given an RGB image $X_v \in \mathbb{R}^{3 \times H_0 \times W_0}$ and a task description $X_l$ (\eg, ``\textit{sit comfortably on}''), our goal is to detect and segment objects most suitable for the specified task. The model predicts bounding boxes $B_{\text{pred}} = \{b_1, \ldots, b_{n_{\text{pred}}}\}$, instance segmentation masks $M_{\text{pred}} = \{m_1, \ldots, m_{n_{\text{pred}}}\}$, and preference scores $S_{\text{pred}} = \{\hat{s}_1, \ldots, \hat{s}_{n_{\text{pred}}}\} \in [0,1]^{n_{pred}}$. Each bounding box $b_i \in [0,1]^4$ contains normalized center coordinates and dimensions, while preference scores indicate object suitability for the task. We denote the complete set of predictions as $O_{\text{pred}} = \langle B_{\text{pred}}, M_{\text{pred}}, S_{\text{pred}} \rangle$. Formally, we seek a function $f$ such that:
\begin{equation}
    f(X_v, X_l) = \langle B_{\text{pred}}, M_{\text{pred}}, S_{\text{pred}} \rangle.
\end{equation}

\subsection{Affordance Reasoning Model Architecture}\label{sec:model_architecture}

To identify suitable objects without explicit object category labels, we design the \affordanceModelName{} with parallel visual and textual pathways, enhanced by specialized modules for multimodal understanding (illustrated in \cref{fig:method_pipeline}). A pre-trained visual encoder processes the input image $X_v$ to extract visual features $F_v$, while a text encoder processes the task description $X_l$ to generate textual features $F_t$. Two key components enhance these representations: the \textbf{\ac{va} module} processes $F_t$ to produce enhanced text features $F_t'$ that emphasize task-specific actions, while the \textbf{\ac{bf} module} integrates $F_t'$ with $F_v$ to generate fused features $F_v'$ and $F_t''$ that capture fine-grained associations between vision and language.

The fused features pass through a transformer encoder-decoder architecture, where the encoder captures global relationships through self-attention mechanisms, and the decoder employs learnable query vectors to generate refined outputs. These outputs are projected through parallel heads to produce bounding boxes $B_{\text{pred}}$ and segmentation masks $M_{\text{pred}}$. The decoder also outputs logits $\mathbf{G}_{\rm{pred}} = [\hat{\mathbf{g}}_{1}, \ldots, \hat{\mathbf{g}}_{n_{\rm{pred}}}] \in \mathbb{R}^{{n_{\rm{pred}}} \times n_{\rm{max}}}$ for computing preference scores $S_{\text{pred}}$. {Here, $n_{\rm{pred}}$ is the number of decoder object queries, and $n_{\rm{max}}$ is the fixed padded text length that includes the special ``no-object'' token.} For each predicted object $i$, the preference score $\hat{s}_i \in S_{\text{pred}}$ is computed as:
\begin{equation}
    \hat{s}_i = 1 - \frac{\exp\left(\hat{g}_{n_{\text{max}}}^i\right)}{\sum_{j=1}^{n_{\text{max}}} \exp\left(\hat{g}_j^i\right)},
    \label{preference_score}
\end{equation}
{where $\hat{g}_j^i$ is the logit of object query $i$ at token position $j$ and $\hat{g}_{n_{\text{max}}}^i$ is the ``no-object'' logit. Thus $\hat{s}_i = 1 - P_i(\mathrm{no\text{-}object})$ measures how likely query $i$ is a valid task-relevant target.}

During training, we compute a bipartite matching between predicted and ground truth objects using the Hungarian algorithm~\cite{kuhn1955hungarian}. The matched predictions are supervised with localization losses (L1 loss and Generalized Intersection over Union loss~\cite{rezatofighi2019generalized}) and segmentation losses (Dice/F-1 loss~\cite{milletari2016v} and focal cross-entropy loss~\cite{lin2017focal}). We incorporate the soft-token prediction loss and contrastive alignment loss from  \ac{mdetr}~\cite{kamath2021mdetr}, adapting them to focus on the entire verb-pronoun description rather than individual tokens. The total loss for the \affordanceModelName{} is formulated as:
\begin{equation}
    \begin{aligned}
    \mathcal{L}_{\text{\affordanceModelName{}}} &= \lambda_1 \mathcal{L}_{\text{L1}} + \lambda_2 \mathcal{L}_{\text{GIoU}} + \lambda_3 \mathcal{L}_{\text{Dice}} \\
        &\quad + \lambda_4 \mathcal{L}_{\text{Cross}} + \lambda_5 \mathcal{L}_{\text{Token}} + \lambda_6 \mathcal{L}_{\text{Align}},
    \end{aligned}
    \label{loss_affordance}
\end{equation}
where $\lambda_1$ to $\lambda_6$ are weights for the respective loss components.

\subsection{Noun-Pronoun Distillation}\label{sec:noun_pronoun_distillation}

\begin{figure}[t!]
    \centering
    \includegraphics[width=\linewidth]{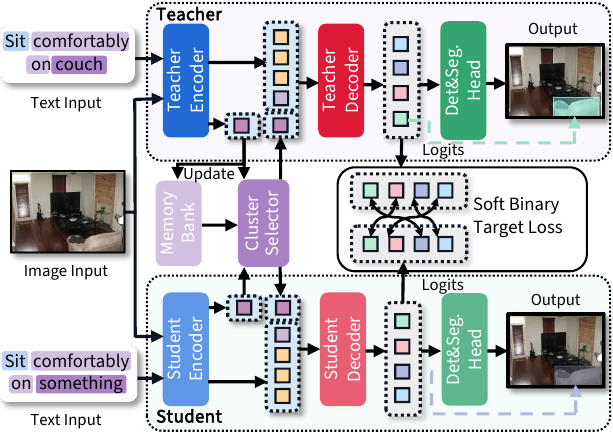}
    \caption{\textbf{Noun-pronoun distillation framework.} Our framework employs parallel teacher-student encoder-decoder architectures for affordance learning. The teacher processes specific noun-based descriptions (\eg, ``\textit{sit comfortably on couch}''), while the student handles generalized pronoun-based inputs (\eg, ``\textit{sit comfortably on something}''). Knowledge transfer occurs through two mechanisms: (i) a memory bank storing noun features that guides the student's cluster selector, and (ii) a soft binary target loss that aligns teacher-student logits. This design enables category-agnostic inference while maintaining category-informed understanding.}
    \label{fig:framework}
\end{figure}

To enable object inference without explicit category labels, we introduce a noun-pronoun distillation framework (illustrated in \cref{fig:framework}). Our approach uses a teacher model trained on task descriptions containing object labels (\eg, ``\textit{sit comfortably with a couch}'') and a student model that processes category-free descriptions (\eg, ``\textit{sit comfortably with something}''). Through clustering and preference distillation, the teacher transfers its object-centric knowledge to the student.

\paragraph*{Clustering Distillation}

We maintain a text feature memory bank that stores noun features from the teacher model, enabling the student to select appropriate noun prototypes for pronoun replacement. The process operates on enhanced text features $F_{\text{noun}}'$ and $F_{\text{pron}}'$, corresponding to noun and pronoun tokens after \ac{va} module processing.

The memory bank is structured as a $n_{\text{task}} \times n_{\text{mem}} \times d$ tensor, where $n_{\text{task}}$ represents the number of tasks, $n_{\text{mem}}$ is the per-task memory size, and $d$ is the feature dimension. For each task $j$, we maintain a queue $\mathbf{F}_{\text{mem}}^j = [F_{1}^j, F_{2}^j, \ldots, F_{n_{\text{mem}}}^j]$ of noun features. {During training, we add teacher noun features $F'_{\rm{noun}}$ and evict their nearest neighbors, a redundancy-aware update that fixes the memory size while preserving feature diversity.} We then apply K-means clustering to $\mathbf{F}_{\text{mem}}^j$ to obtain $K$ cluster centers $\mathbf{F}_{\text{c}}^j = \{ F_{c_1}^j, F_{c_2}^j, \ldots, F_{c_K}^j \}$.

Knowledge transfer occurs through a cluster selector in the student model, which uses nearest neighbor classification to select a prototype $F_{c_s}^j$ from $\mathbf{F}_{\text{c}}^j$ based on the pronoun feature $F_{\text{pron}}'$. This prototype replaces $F_{\text{pron}}'$ in the student's feature sequence. To ensure proper alignment, we define the cluster loss $\mathcal{L}_{\text{cluster}} = \|F_{\text{pron}}' - F_{c_s}^j \|_2$, which minimizes the Euclidean distance between the pronoun feature and selected cluster center.

\paragraph*{Preference Distillation}

We align teacher and student predictions through a soft binary target loss based on the \ac{kl} divergence. For each object query, we compute binary probabilities indicating positive (ground truth object) or negative matches: $\mathbf{p} = [p^{\rm{pos}}, p^{\rm{neg}}] \in \mathbb{R}^{1\times2}$. These probabilities are defined using the softmax function:
\begin{equation}
    p^{\text{pos}} = \dfrac{\sum_{j=1}^{n_{\text{max}} - 1} \exp(\hat{g}_j)}{\sum_{j=1}^{n_{\text{max}}} \exp(\hat{g}_j)}, \quad
    p^{\text{neg}} = \dfrac{\exp(\hat{g}_{n_{\text{max}}})}{\sum_{j=1}^{n_{\text{max}}} \exp(\hat{g}_j)},
\end{equation}

The probability sequences for teacher and student models, denoted as $\mathbf{P}_t = [\mathbf{p}_{t_1}, \ldots, \mathbf{p}_{t_{n_{\text{pred}}}}]$ and $\mathbf{P}_s = [\mathbf{p}_{s_1}, \ldots, \mathbf{p}_{s_{n_{\text{pred}}}}]$, are aligned through bipartite matching. Using the Hungarian algorithm~\cite{kuhn1955hungarian}, we find an optimal permutation $\sigma \in \mathfrak{S}_{n_{\text{pred}}}$ that minimizes the matching cost:
\begin{equation}
    \hat{\sigma} = \arg\min_{\sigma \in \mathfrak{S}{n_{\text{pred}}}} \sum_{i=1}^{n_{\text{pred}}} \mathcal{L}_{\text{match}}\left(y_{t_i}, y_{s_{\sigma(i)}}\right),
\end{equation}
where $y_{t_i} = (\hat{b}_{t_i}, \mathbf{p}_{t_i})$ combines the teacher's bounding box prediction $\hat{b}_{t_i}$ and probabilities, and $\mathcal{L}_{\text{match}}$ incorporates both box prediction losses and \ac{kl} divergence.

The soft binary target loss is then defined using the optimal assignment $\hat{\sigma}$:
\begin{equation}
    \mathcal{L}_{\text{binary}} = \sum_{i=1}^{n_{\text{pred}}} \mathcal{L}_\text{KL}\left(\mathbf{p}_{t_i}, \mathbf{p}_{s_{\hat{\sigma}(i)}}\right),
\end{equation}
where the KL divergence between teacher and student probabilities is:
\begin{equation}
    \resizebox{0.9\hsize}{!}{$\displaystyle
        \mathcal{L}_\text{KL}\left(\mathbf{p}_{t_i}, \mathbf{p}_{s_{\hat{\sigma}(i)}}\right) = p_{t_i}^{\text{pos}} \log \left( \dfrac{p_{t_i}^{\text{pos}}}{p_{s_{\hat{\sigma}(i)}}^{\text{pos}}} \right) + p_{t_i}^{\text{neg}} \log \left( \dfrac{p_{t_i}^{\text{neg}}}{p_{s_{\hat{\sigma}(i)}}^{\text{neg}}} \right).%
    $}%
\end{equation}

Minimizing $\mathcal{L}_{\text{binary}}$ aligns the student's binary query probabilities with the teacher's. Since the preference score $\hat{s}_i$ (\cref{preference_score}) follows a similar formulation to $p^{\text{pos}}$, this effectively transfers preference knowledge from teacher to student.

\paragraph*{Overall Training Loss}

We combine all components into a final training objective for the \affordanceModelName{} with noun-pronoun distillation:
\begin{equation}
    \mathcal{L}_{\text{\affordanceModelName{}}-\text{NP}} = \mathcal{L}_{\text{\affordanceModelName{}}}^t + \mathcal{L}_{\text{\affordanceModelName{}}}^s + \lambda_7 \mathcal{L}_{\text{cluster}}^s + \lambda_8 \mathcal{L}_{\text{binary}}^s,
\end{equation}
where $\mathcal{L}_{\text{\affordanceModelName{}}}^t$ and $\mathcal{L}_{\text{\affordanceModelName{}}}^s$ represent the teacher and student model losses, and $\lambda_7$, $\lambda_8$ weight the distillation components. The distillation losses $\mathcal{L}_{\text{cluster}}^s$ and $\mathcal{L}_{\text{binary}}^s$ apply only to the student model. During inference, we employ only the student model with the fixed memory bank, maintaining category-agnostic object identification.

We provide additional methodological details, including \emph{noun feature} representations in \cref{appendix:noun_features} and \emph{loss function} derivations in \cref{appendix:loss}. The effectiveness of our proposed modules is validated through extensive experiments detailed in \cref{sec:exp:ablation,appendix:ablation_study}. Furthermore, in \cref{sec:exp:comparison}, we demonstrate that the integration of these modules achieves state-of-the-art performance across multiple benchmarks.

\begin{figure*}[t!]
    \centering
    \includegraphics[width=\linewidth]{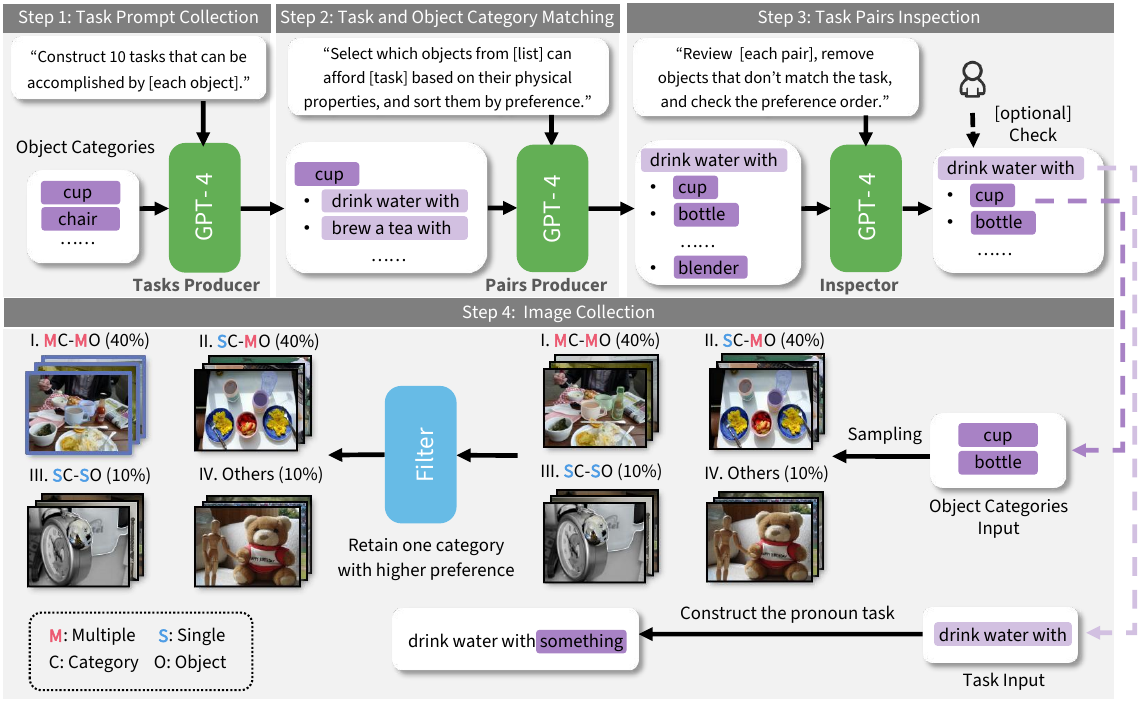}
    \caption{\textbf{Pipeline for affordance dataset construction.} Our automated pipeline transforms detection datasets into affordance knowledge bases through four steps: (i) task generation---creating diverse tasks per object category, (ii) pair matching---associating tasks with suitable objects and establishing affordance rankings, (iii) quality inspection---validating task-object pairs and ranking accuracy, and (iv) image sampling---following composition rules (40\% MCMO, 40\% SCMO, 10\% SCSO, 10\% no targets) while retaining highest-priority objects per task. This systematic approach ensures comprehensive coverage of affordance relationships while maintaining data quality.}
    \label{fig:dataset_construction}
\end{figure*}

\subsection{Verb Attention module and Bi-Fusion module}\label{sec:method:modules}

To enhance multimodal understanding, we introduce two specialized components: the \ac{bf} module for cross-modal feature integration and the \ac{va} module for action-focused reasoning.

\paragraph*{\ac{bf} module}

We design this module to overcome limitations of  \ac{mdetr}'s feature concatenation approach by implementing bi-directional cross-modal attention between visual and textual features. This direct interaction captures fine-grained vision-language associations through parallel bi-directional attention:
\begin{equation}
    \begin{aligned}
        F_v' &= F_v + \gamma_v \cdot \mathrm{Attn}_{v \rightarrow t}\left(\mathrm{LN}(F_v), \mathrm{LN}(F_t)\right), \\
        F_t' &= F_t + \gamma_t \cdot \mathrm{Attn}_{t \rightarrow v}\left(\mathrm{LN}(F_t), \mathrm{LN}(F_v)\right),
    \end{aligned}
\end{equation}
where $\mathrm{Attn}_{v \rightarrow t}$ and $\mathrm{Attn}_{t \rightarrow v}$ implement multi-head attention between modalities, $\mathrm{LN}$ performs layer normalization, and learnable parameters $\gamma_v$, $\gamma_t$ control cross-modal influence.

\paragraph*{\ac{va} module}

This module enhances action-related information processing while reducing interference from dominant object categories. Leveraging the standardized verb-object-preposition format of task descriptions, it applies cross-attention between verb features and the complete textual prompt. Given the first verb's feature $F_{{vb}} \in \mathbb{R}^{C_t}$ and full text features $F_t \in \mathbb{R}^{L \times C_t}$ from the text encoder (where $L$ is sequence length and $C_t$ is feature dimension), the module computes:
\begin{equation}
    F_t'' = F_t + \mathrm{CrossAttn}\left(\mathrm{LN}(F_t), \mathrm{LN}(F_{{vb}})\right),
\end{equation}
where $\mathrm{CrossAttn}$ performs cross-attention to produce enhanced text features $F_t''$ with dimensions matching $F_t$, effectively amplifying action-related information in the final representation.

\subsection{Summary}

The combination of noun-pronoun distillation, bi-directional feature fusion, and verb-focused attention enables \affordanceModelName{} to effectively reason about object affordances without relying on explicit category labels. The teacher-student framework transfers object-centric knowledge while maintaining category-agnostic inference, the \ac{bf} module ensures comprehensive multimodal understanding, and the \ac{va} module emphasizes action-specific features critical for affordance reasoning. Together, these components form a robust architecture that bridges the gap between category-specific training and category-agnostic deployment while maintaining high performance in affordance detection tasks.

\section{Dataset Construction}\label{sec:dataset}

We present a scalable approach for creating large-scale affordance knowledge bases through automated conversion of object detection datasets. Our pipeline transforms standard detection annotations into rich affordance-task pairs, yielding two comprehensive datasets: \affordanceDatasetName{} from COCO and \affordanceDatasetNamelvis{} from LVIS. We detail our conversion methodology in \cref{sec:dataset_collection} and analyze dataset characteristics in \cref{sec:dataset_statistics}.

\subsection{Dataset Collection}\label{sec:dataset_collection}

A primary challenge in developing \affordanceModelName{} is establishing comprehensive affordance knowledge without depending on \ac{llm} inference. While manual dataset construction is possible, it becomes impractical due to the complex many-to-many relationships between tasks and objects: single objects can serve multiple purposes, tasks can utilize various objects, and objects have different levels of suitability for each task. This intricate mapping makes manual knowledge base construction both time-intensive and potentially inconsistent.

We address this challenge through an automated pipeline that leverages \ac{llm} capabilities for dataset construction. Our approach prioritizes three critical factors: task diversity (enabling broad affordance reasoning), image quantity (supporting physical property learning and scene generalization), and object category variety (covering diverse usage scenarios). While existing detection datasets provide rich visual and categorical resources, we needed a systematic method to generate diverse tasks and establish meaningful task-object relationships.

Our pipeline (illustrated in \cref{fig:dataset_construction}) employs GPT-4 for both task-object pair generation and quality inspection. After filtering uncommon object categories (\eg, animals, musical instruments, food items) that rarely serve as tools, we proceed through four systematic steps:

\textbf{Step 1: Task Prompt Collection.} A GPT-4-based task producer generates 10 diverse tasks per object category, building an initial task pool. This step captures various potential uses for each object, ensuring comprehensive affordance coverage while maintaining natural and practical tasks.

\textbf{Step 2: Task and Object Category Matching.} A GPT-4-based pair producer matches tasks with relevant object categories, incorporating commonsense preference rankings. For instance, in ``\textit{drink water with},'' cups receive higher rankings than bottles, reflecting intuitive usage preferences. This ranking system captures nuanced distinctions in object suitability for specific tasks.

\textbf{Step 3: Task Pairs Inspection.} A GPT-4-based inspector performs multi-level quality control: filtering tasks against predefined criteria, verifying object-task match rankings, and removing inappropriate pairs (\eg, excluding blenders from ``\textit{drink water with}''). An optional manual review reduces task redundancy to optimize training efficiency, though this primarily serves computational rather than quality purposes.

\textbf{Step 4: Image Collection.} Following COCO-Tasks~\cite{xin2022cocotasks}, we organize images into four configurations: MCMO (multiple categories, multiple objects), SCMO (single category, multiple objects), SCSO (single category, single object), and Others (random images without target categories). We retain highest-priority objects per task and append ``\textit{something}'' to prompts, balancing task specificity with visual diversity.

This automated pipeline enables efficient construction of large-scale affordance datasets while maintaining data quality. By leveraging \ac{llm} capabilities for generation and inspection, we create comprehensive knowledge bases that capture complex task-object relationships. Human validation and preference-aware target-selection examples are provided in~\cref{appendix:data:human_validation,appendix:data:preference_examples}.

\begin{figure*}[t!]
    \centering
    \includegraphics[width=\linewidth]{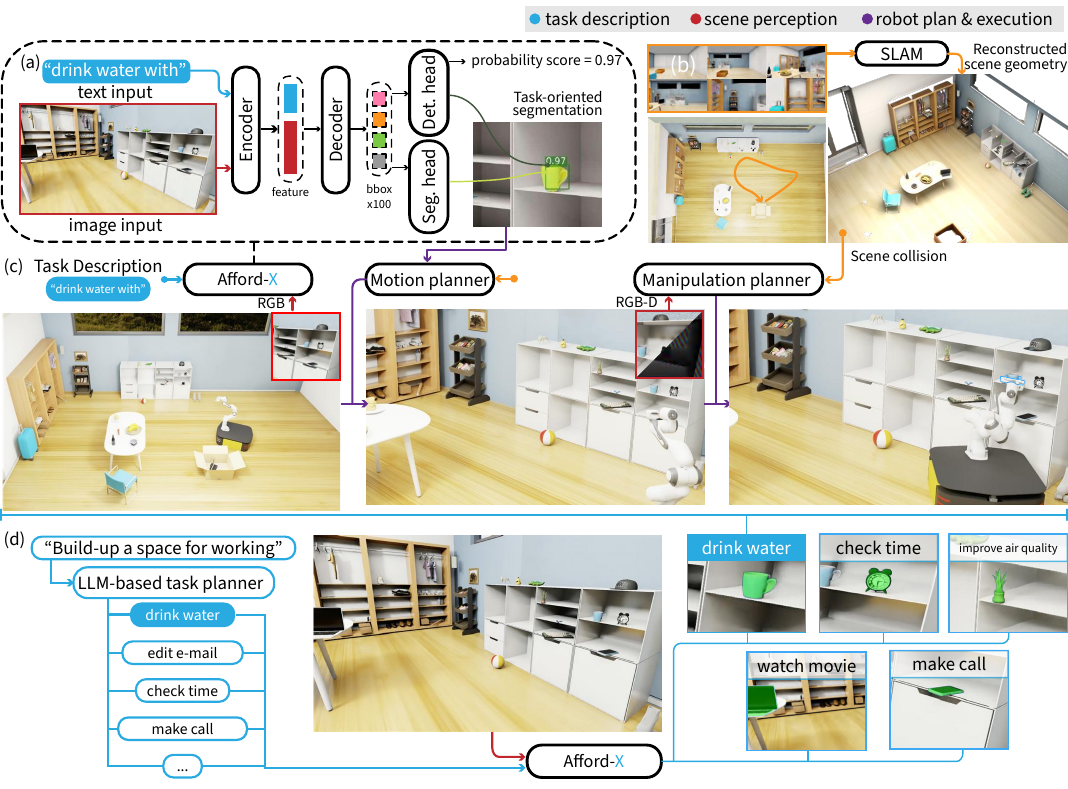}
    \caption{\textbf{System infrastructure for object selection and grasping as a showcase of task-oriented manipulation.} Our robot executes the task ``drink water with'' through a multi-stage process: (a) \affordanceModelName{} performs initial affordance reasoning on RGB input, generating instance segmentation masks for suitable objects. (b) Optionally, the robot navigates the scene to reconstruct a point cloud with SLAM for collision-free motion planning in completely unseen scenes. (c) The robot executes task-oriented object grasping by locating an object via \affordanceModelName{}, approaching for detailed perception, and performing grasping. (d) For complex tasks (\eg, ``build up a space for working''), the system decomposes the high-level goal into sequential sub-tasks, applying procedure (c) to each. See more examples in \cref{appendix:tom}.}
    \label{fig:infra_compressed}
\end{figure*}

\subsection{Dataset Statistics}\label{sec:dataset_statistics}

Our construction pipeline produced two comprehensive affordance datasets: \affordanceDatasetName{} and \affordanceDatasetNamelvis{}. \affordanceDatasetName{}, derived from COCO2014~\cite{lin2014coco}, contains 112k images spanning 1130 task prompts and 80 object categories. We partitioned the dataset into training (600 images/task) and test (150 images/task) sets, sampling from COCO2014's training and validation sets respectively for fair evaluation.

\affordanceDatasetNamelvis{} represents a significant expansion using LVIS~\cite{gupta2019lvis}, encompassing 119k images, 1494 task prompts, and 1064 object categories. We followed the same sampling strategy, drawing training (600 images/task) and test (150 images/task) sets from COCO2017's training and validation sets respectively, maintaining strict separation for reliable evaluation.

Both datasets exhibit long-tail distributions in task-object relationships, reflecting natural variations in object functionality. Objects with limited functional diversity generated fewer unique tasks during Step 1 due to task overlap, leading to subsequent filtering. Despite this natural skew, the datasets provide comprehensive coverage of daily-life affordance scenarios, establishing robust knowledge bases for affordance reasoning. More details for the specifications and characteristics of both constructed datasets is presented in \cref{appendix:data}

\section{Embodied Affordance Reasoning}\label{sec:embodied}

{As discussed in~\cref{sec:rw:tom}, traditional task-oriented robotics often relies on human supervision or predefined rules for object selection prior to specific manipulation, limiting autonomy in open-world scenarios. Afford-X focuses on selecting the suitable task-relevant object in an open-ended scene and obtaining its visual representation for downstream manipulation modules. This step is especially important when diverse objects and contexts are present, where the target object cannot be assumed to be specified in advance.}

To demonstrate this, we evaluate \affordanceModelName{}'s practical applications by integrating it into embodied agents within simulated environments. {Our evaluation examines two key capabilities: (i) scene-level affordance reasoning, targeted on analyzing object affordance in complex 3D environments, and (ii) task-oriented object grasping, as a minimal yet representative showcase of manipulation based on the outputs of affordance reasoning.} We leverage Isaac Sim's photorealistic rendering to provide both high-fidelity rendering and precise ground-truth annotations, enabling systematic assessment of how our model bridges perception and action in embodied contexts.

\subsection{Affordance Reasoning in the Scene}\label{sec:inscene}

We developed a comprehensive evaluation within NVIDIA Isaac Sim to rigorously evaluate \affordanceModelName{}'s affordance reasoning in 3D environments. This evaluation assesses the model's understanding of object-task relationships across diverse environmental configurations and viewing conditions.

Our test environments employ a multi-source scene composition approach. We begin with base scenes from the Evermotion dataset and OmniGibson~\cite{li2023behavior}, providing realistic room layouts and furniture arrangements. We augment these with randomly placed textured meshes from Objaverse~\cite{deitke2023objaverse}, following two principles: (i) including both task-suitable and unsuitable objects to test discrimination capability, and (ii) randomized object placement for complexity.

We evaluate \affordanceModelName{}'s ability in selecting appropriate object for the desired task in the simulated 3D scenes, and compare it with \ac{llm}-integrated pipelines. The evaluation features an RGB-D camera for environmental perception. Isaac Sim's physics engine ensures authentic environmental interactions, while its ray-tracing renderer generates photorealistic images with lighting and materials. The simulator provides ground-truth annotations including object bounding boxes and segmentation masks for quantitative evaluation. During testing, \affordanceModelName{} processes RGB images alongside task instructions, generating segmentation masks and for identifying task-relevant objects. This enables the robot to identify suitable objects for task execution. These procedures enables systematic evaluation across varying scenes, viewpoints, and task contexts while maintaining experimental reproducibility. The results are reported and analyzed in \cref{sec:exp:sructure}. Further towards the open-ended world, \cref{appendix:tom} reports a more in-depth evaluation on \affordanceModelName{}'s capabilities in diverse 3D scenes, and \cref{appendix:tom:occ_and_size} reports analysis on object size and occlusion.

\begin{table*}[t!]
    \centering
    \small
    \caption{\textbf{Comparison of \affordanceModelName{} with state-of-the-art methods.} Comprehensive evaluation across COCO-Tasks, \affordanceDatasetName{}, and \affordanceDatasetNamelvis{} datasets demonstrates the effectiveness of our approach, with \affordanceModelName{} consistently achieving superior   performance in both affordance understanding and instance segmentation tasks. The performance gains stem from our proposed \ac{va} and \ac{bf} modules, which enhance the baseline architecture. Results marked with $\dagger$ are from original papers, with $\ddagger$ are from our previous conference work, and with \textbf{bold} and \underline{underlined} values indicating best and second-best performance, respectively.}
    \begin{tabular*}{\linewidth}{@{\extracolsep{\fill}}clcccccc}
        \toprule
        \multirow{2}{*}{\textbf{Index}} & 
        \multirow{2}{*}{\textbf{Method}} & \multicolumn{2}{c}{\textbf{COCO-Tasks}} & \multicolumn{2}{c}{\textbf{\affordanceDatasetName{}}} & \multicolumn{2}{c}{\textbf{\affordanceDatasetNamelvis{}}} \\ 
        \cmidrule(lr){3-4} \cmidrule(lr){5-6} \cmidrule(lr){7-8}
        & & $\rm{mAP}^{\rm{box}}$ & $\rm{mAP}^{\rm{mask}}$ & $\rm{mAP}^{\rm{box}}$ & $\rm{mAP}^{\rm{mask}}$ & $\rm{mAP}^{\rm{box}}$ & $\rm{mAP}^{\rm{mask}}$ \\ 
        \midrule
        (a) & Fast R-CNN~\cite{ren2016faster} + GGNN$^\dagger$ & 32.6  & -  & - & -  & - & -  \\
        (b) & YOLO + GGNN~\cite{sawatzky2019object}$^\dagger$           & 33.2 & -  & -  & -  & - & -   \\
        (c) & \ac{mdetr} (w/o pretraining) + GGNN$^\ddagger$ & 9.6  & 8.6  & - & -  & - & -  \\
        (d) & \ac{mdetr} + GGNN$^\ddagger$ & 36.8 & 30.3  & -  & -  & - & -   \\
        (e) & ViTDet (ViT-B)~\cite{li2022exploring} + GGNN           & 22.5 & 29.5 & 27.4 & 22.8 & 6.6 & 5.9  \\
        (f) & ViTDet (ViT-L) + GGNN           & 32.1 & 24.6 & 29.7 & 24.7 & 8.0 & 7.1  \\
        (g) & ViTDet (ViT-H) + GGNN           & 33.8 & 25.9 & 31.5  & 26.1  & 8.5  & 7.4  \\\midrule
        (h) & \ac{mdetr} \cite{kamath2021mdetr}                         & 41.3$^\ddagger$ & 35.2$^\ddagger$ & 44.7  & 41.0  & 25.1  & 22.7 \\
        (i) &\ac{mdetr} (w/ VA \& BF)                         & 43.2 & 36.9 & \underline{45.2}  & \underline{41.4} & \underline{26.8}   & \underline{24.2} \\ \midrule
        (j)  & TOIST~\cite{li2022toist}          & \underline{44.1}$^\ddagger$ & \underline{39.0}$^\ddagger$ & 44.9 & 41.3 & 26.2 & 23.4 \\ 
        (k) & \affordanceModelName{} (w/ VA \& BF)           & \textbf{45.3} & \textbf{39.2}  & \textbf{45.8} & \textbf{42.5} & \textbf{27.7} & \textbf{24.8}  \\  
        \bottomrule
    \end{tabular*}
    \label{table:main_result}
\end{table*}

\subsection{{Task-oriented Object Grasping}}\label{sec:tom}

{We use object grasping as a minimal yet representative task to evaluate whether the predicted object-level affordance mask can support downstream manipulation.} Our system architecture (\cref{fig:infra_compressed}) combines \affordanceModelName{} with established manipulation modules: GraspNet~\cite{fang2020graspnet} as the grasp planner and cuRobo~\cite{sundaralingam2023curobo} as the collision-aware motion planner. The manipulation follows the four-phase procedure:
(i) \textbf{Affordance reasoning:} \affordanceModelName{} processes the scene's RGB frame with a task description, identifying suitable target objects through segmentation masks.
(ii) \textbf{Viewpoint optimization:} The robot positions itself at a prescribed viewing distance $d_\mathrm{view}$ from the identified target for detailed perception, where the target camera location is computed using the depth image additionally acquired in step (i); 
{An affordance mask is inferred from each perspective and used to segment the RGB-D image into an object point cloud from the masked region and a scene point cloud from the remaining region. 
(iii) \textbf{Grasp planning:} The grasp planner~\cite{fang2020graspnet} generates object-level grasp proposals (each consists of a gripper pose, a width, and a quality score) using the segmented object point cloud as the grasping target. 
(iv) \textbf{Grasp execution:} A motion planner generates trajectories for the grasp proposals using the scene point cloud as collider and executes the planned grasp motion with the highest grasp score on the robot.}
We operate in simulated scenes and objects from OmniGibson~\cite{li2023behavior} (see \cref{fig:sim_scenes}). During grasp execution, we use a fixed joint to attach the object to the gripper when the gripper is closing if the object is close enough to the gripper, which is a standardized protocol to isolate affordance reasoning from unstable contact simulation.

For complex tasks requiring multiple steps (\eg, ``\textit{build up a space for working}'', \cref{fig:infra_compressed}(d)), we employ an \ac{llm} to decompose the high-level goal into atomic sub-tasks. Each sub-task follows the same three-phase protocol sequentially, enabling structured completion of complex manipulations through affordance-guided action sequences. This integration of affordance reasoning with manipulation planning advances autonomous task-oriented robotics, enabling robots to independently identify and utilize appropriate objects based on task requirements. \cref{appendix:tom:long_horizon_tasks} provides more details and results.

\section{Experiment}\label{sec:experiment}

To validate the effectiveness of \affordanceModelName{}, we conduct comprehensive experiments across three key dimensions: (i) performance comparison and ablation studies (\cref{sec:exp:comparison,sec:exp:ablation}), (ii) dataset analysis (\cref{sec:data_analysis}), and (iii) real-world applicability (\cref{sec:exp:sructure,sec:manipulation}). The latter includes comparative analysis against \ac{llm}-based approaches and validation in task-oriented manipulation scenarios.

\subsection{Implementation Details}\label{sec:exp:details}

\paragraph*{Model Architecture}

We implement \affordanceModelName{} using RoBERTa-base~\cite{liu2019roberta} for text encoding and ResNet-101~\cite{he2016deep} as the \ac{cnn} backbone. To leverage existing vision-language understanding capabilities, we initialize our model with pre-trained weights from \ac{mdetr}~\cite{kamath2021mdetr}. We evaluate this architecture across three progressively more challenging datasets: COCO-Tasks, \affordanceDatasetName{}, and \affordanceDatasetNamelvis{}.

\paragraph*{Training Protocol}

Our training follows a carefully designed multi-stage approach. In the initial verb-pronoun and verb-noun stages, we employ the Adam optimizer with a batch size of 36, applying a uniform learning rate of $10^{-5}$ across the text encoder, backbone network, and \ac{bf} module. The subsequent distillation stage requires more precise parameter updates, leading to the reduced batch size 18 and text encoder learning rate $5 \times 10^{-6}$ while maintaining other learning rates.

\paragraph*{Data Augmentation}

To enhance model robustness to real-world variations, we implement a comprehensive data augmentation pipeline. This includes dynamic image resizing that randomly scales the shortest side between 480-800 pixels while capping the longest side at 1333 pixels. We further augment the training data through random cropping (probability 0.5), generating diverse viewports between 384-1333 pixels.

\paragraph*{Evaluation Metrics}

We evaluate model performance using the AP@0.5 metric, which assesses localization accuracy and ranking effectiveness through predicted preference scores $S_{\text{pred}}$. For comprehensive evaluation across different tasks, we report the mean Average Precision (mAP@0.5), calculated by averaging AP@0.5 scores across all experimental tasks.

\subsection{Performance Analysis}\label{sec:exp:comparison}

\begin{figure*}[t!]
    \centering
    \setlength{\tabcolsep}{2pt}
    \begin{tabular}{
        >{\centering\arraybackslash}m{0.5em} c c c}
    \parbox[c]{0.5em}{
        \vspace{-0.9cm}
        \small{\textbf{\rotatebox{90}{Prediction}}}\\[0.3cm]
        \small{\textbf{\rotatebox{90}{Ground Truth}}}
    }
    &
    \begin{subfigure}[t]{0.32\linewidth}
        \centering
        \begin{tabular}{cc}
            \includegraphics[
                trim=50 20 0 20, clip,
                width=1.1in, height=0.825in
            ]{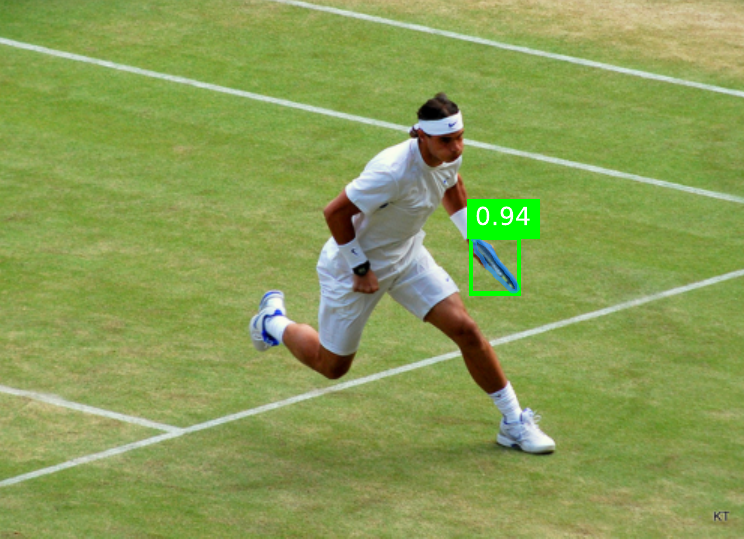} 
            &
            \includegraphics[
                trim=0 35 0 70, clip,
                width=1.1in, height=0.825in
            ]{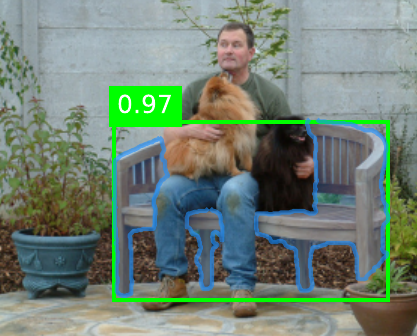} \\
            \includegraphics[
                trim=50 20 0 20, clip,
                width=1.1in, height=0.825in
            ]{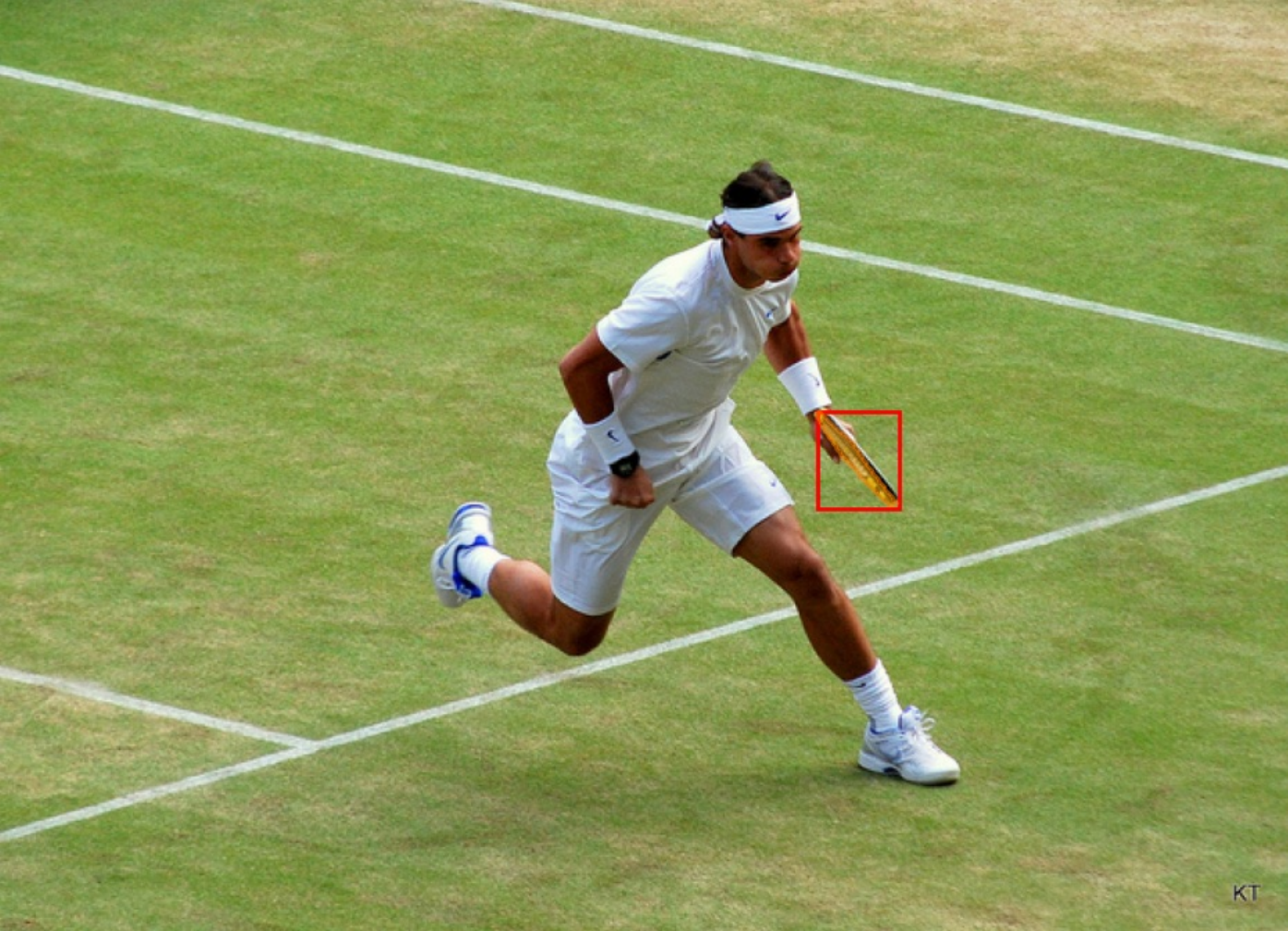} 
            &
            \includegraphics[
                trim=0 85 0 180, clip,
                width=1.1in, height=0.825in
            ]{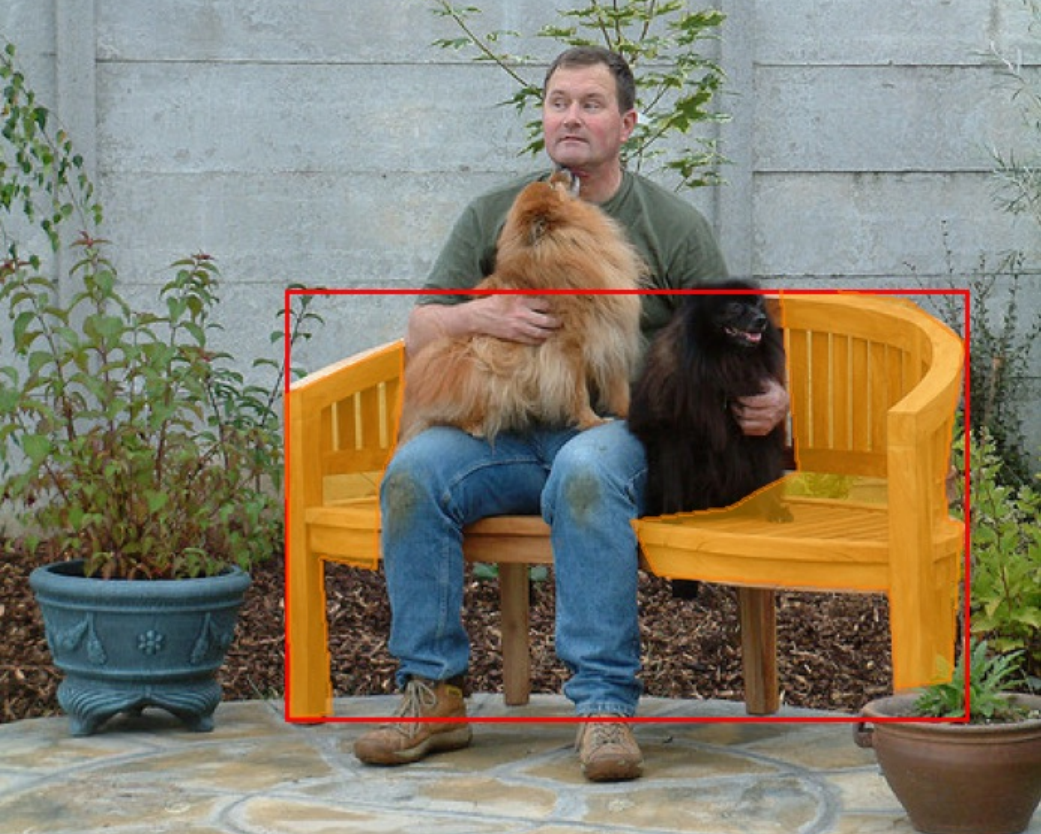}\\
            {\small dig hole with} & {\small step on}
        \end{tabular}
        \caption{COCO-Tasks}
        \label{fig:vis_coco_tasks}
    \end{subfigure}
    &
    \begin{subfigure}[t]{0.32\linewidth}
        \centering
        \begin{tabular}{cc}
            \includegraphics[
                width=1.1in, height=0.825in
            ]{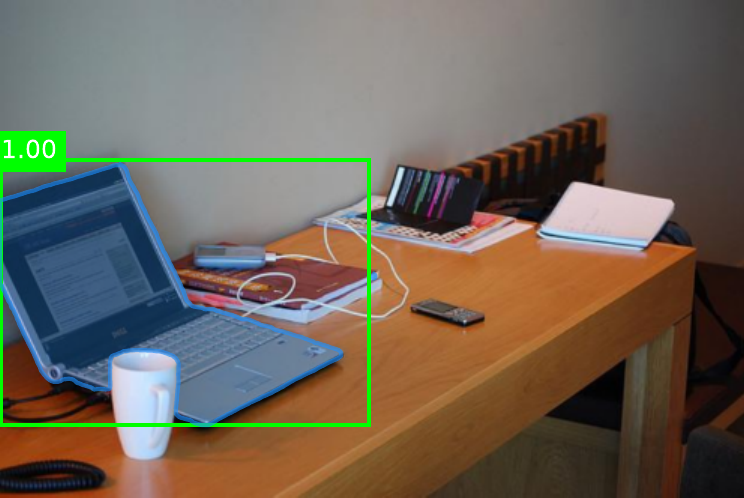} 
            &
            \includegraphics[
                trim=0 0 0 10, clip,
                width=1.1in, height=0.825in
            ]{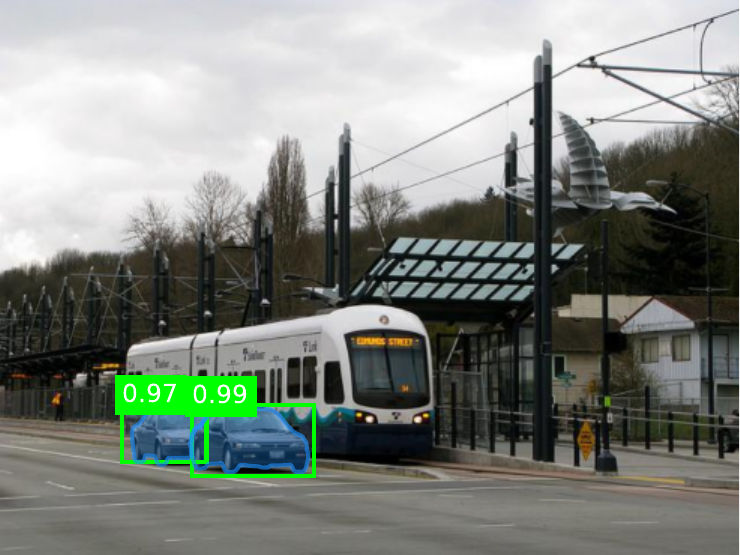} \\
            \includegraphics[
                width=1.1in, height=0.825in
            ]{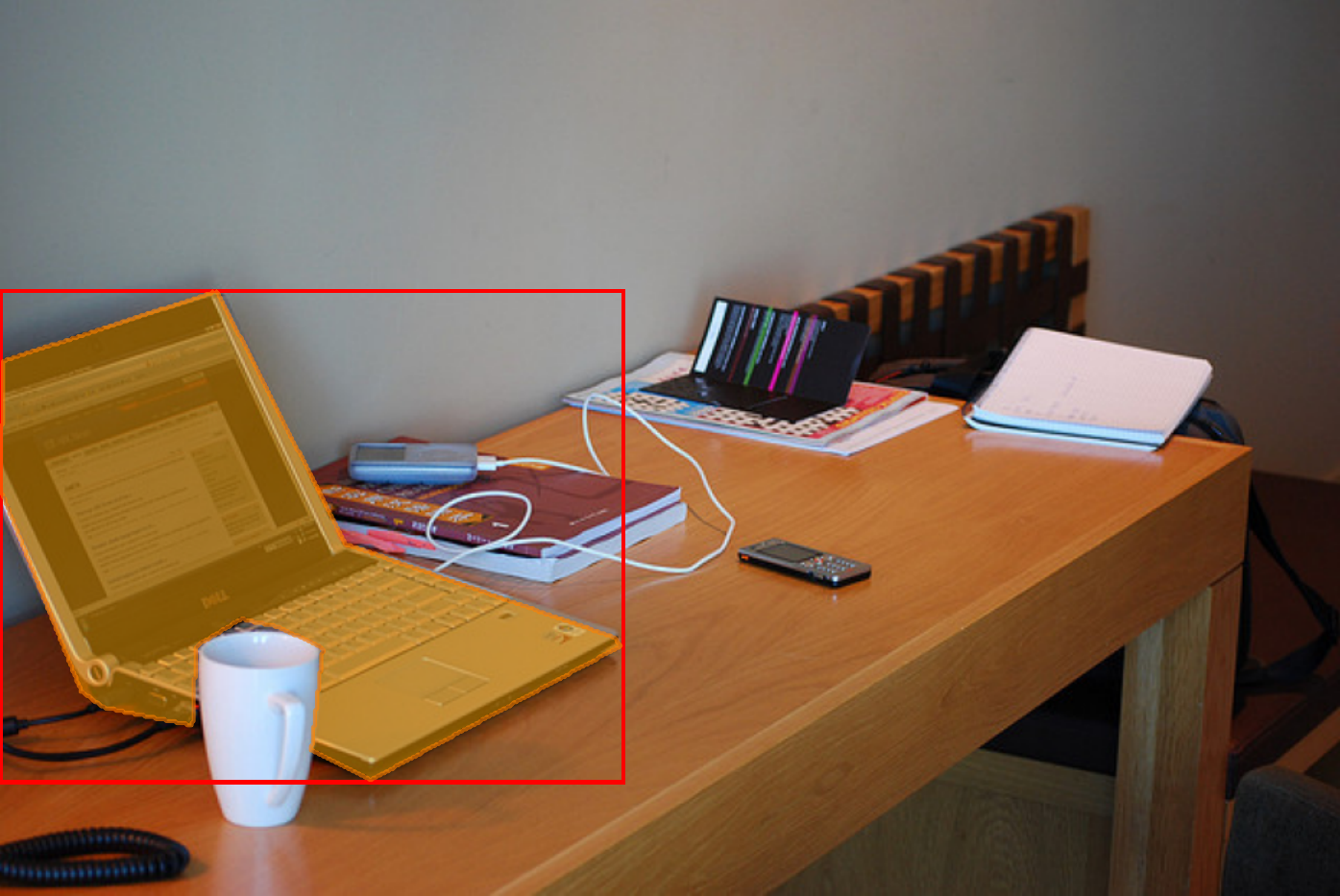} 
            &
            \includegraphics[
                trim=0 0 50 10, clip,
                width=1.1in, height=0.825in
            ]{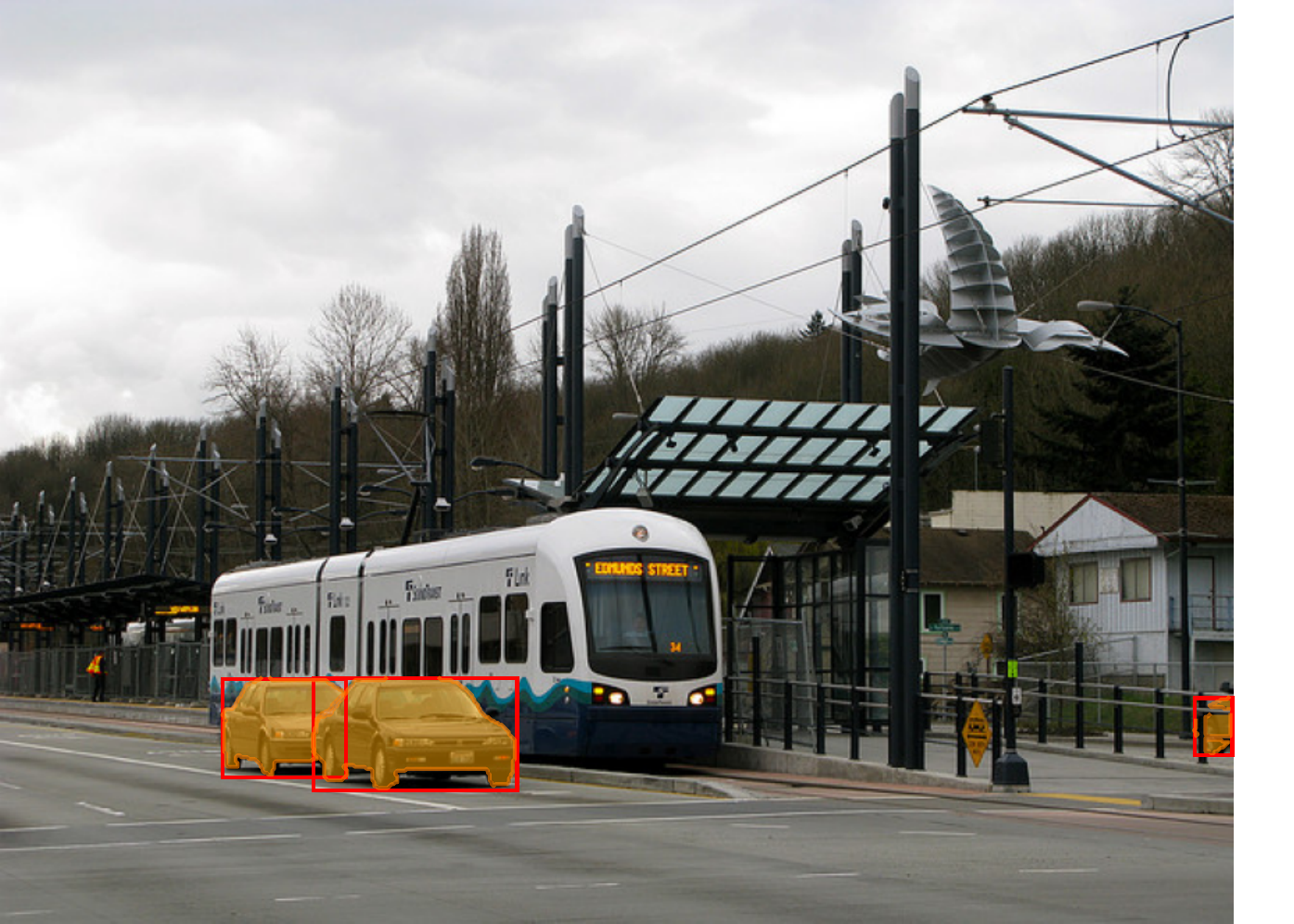}\\
            {\small{edit photos with}} & {\footnotesize commute to work with}
        \end{tabular}
        \caption{COCO-Aff}
        \label{fig:vis_coco_aff}
    \end{subfigure}
    &
    \begin{subfigure}[t]{0.32\linewidth}
        \centering
        \begin{tabular}{cc}
            \includegraphics[
                width=1.1in, height=0.825in
            ]{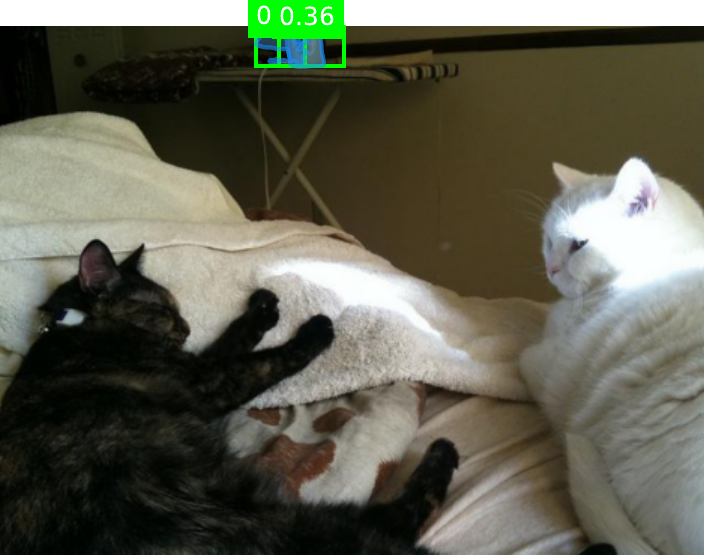} 
            &
            \includegraphics[
                trim=0 110 0 0, clip,
                width=1.1in, height=0.825in
            ]{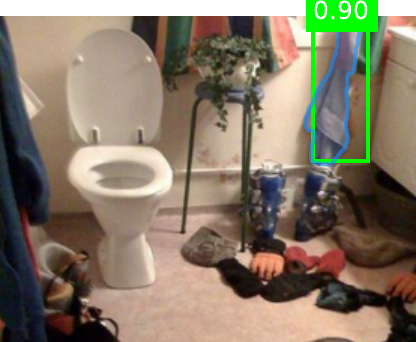} \\
            \includegraphics[
                width=1.1in, height=0.825in
            ]{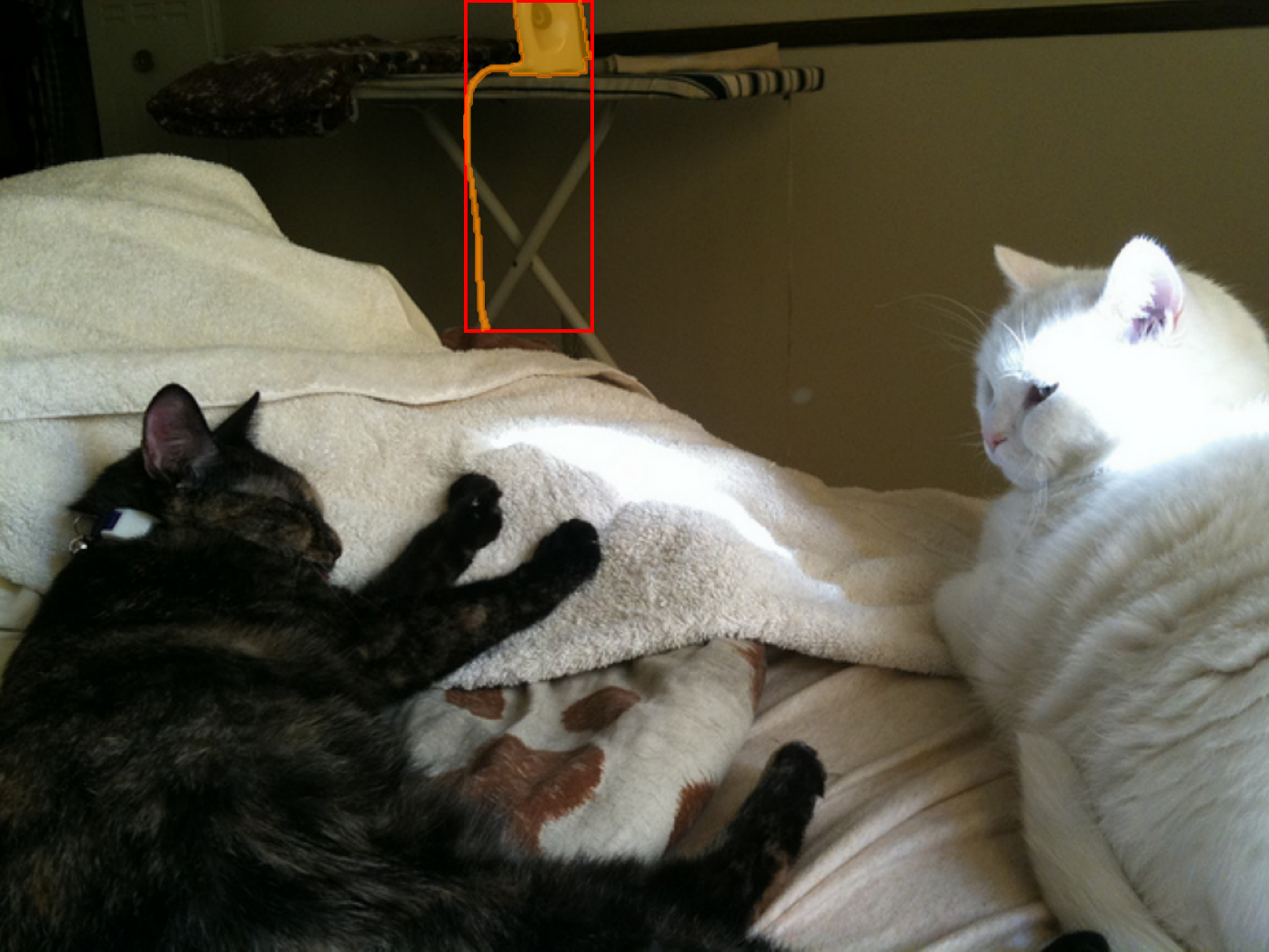} 
            &
            \includegraphics[
                trim=0 300 0 0, clip,
                width=1.1in, height=0.825in
            ]{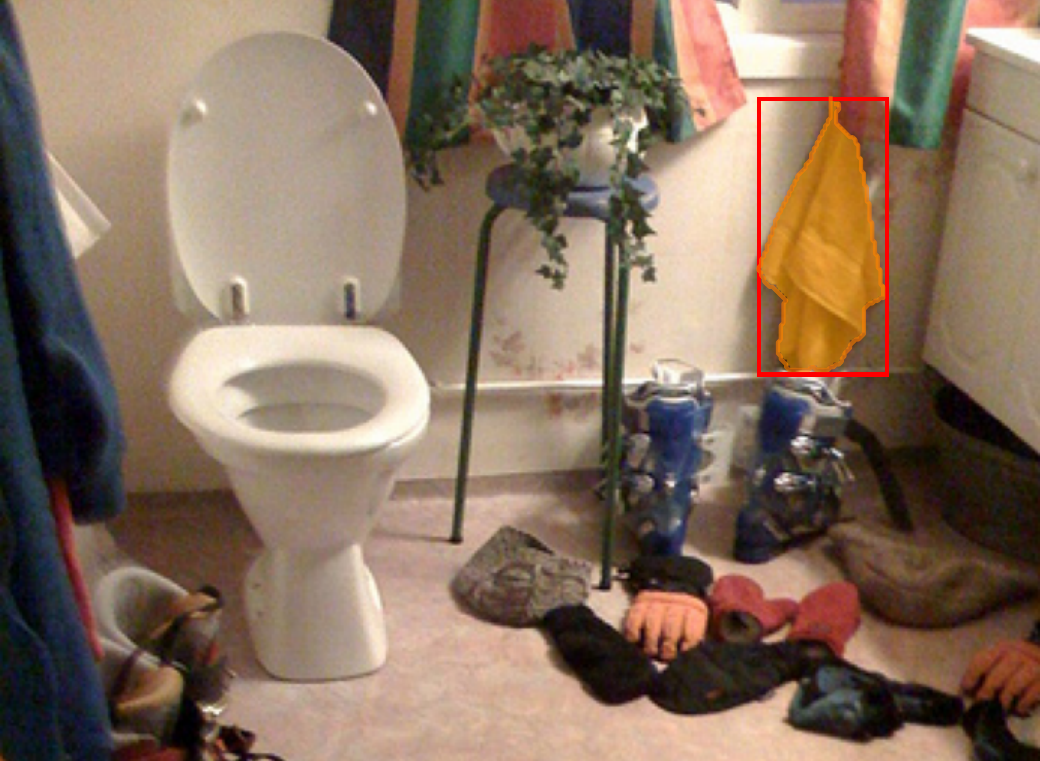}\\
            {\small iron clothes with} & {\small clean windows with}
        \end{tabular}
        \caption{LVIS-Aff}
        \label{fig:vis_lvis_aff}
    \end{subfigure}
    \end{tabular}
    \caption{\textbf{Qualitative results of \affordanceModelName{} on three datasets.} The first row of each section shows model predictions, with corresponding ground truth in the second row. Results demonstrate performance across (a) COCO-Tasks, showing sports and office scenes; (b) \affordanceDatasetName{}, depicting indoor and outdoor environments; and (c) \affordanceDatasetNamelvis{}, illustrating diverse interaction scenarios. Green boxes indicate model predictions while orange boxes show ground truth annotations.}
    \label{fig:main_vis}
\end{figure*}

Our comprehensive experimental results are presented in \cref{table:main_result}. We evaluate our approach against existing methods based on pre-trained \acp{mm}, excluding \ac{llm}-augmented baselines for fair comparison. Prior approaches fall into two categories: two-stage methods combining object detection with \ac{ggnn}~\cite{sawatzky2019object} (rows (a)-(g)), and one-stage methods built on \ac{mdetr} (rows (h)-(k)).

In the two-stage category, we evaluated three object detection frameworks combined with \ac{ggnn}: CNN-based Fast R-CNN~\cite{ren2016faster}, YOLO (version 11), and transformer-based approaches (\ac{mdetr}, ViTDet~\cite{li2022exploring}). Despite using COCO-pretrained weights, these methods showed limited performance. Even on the relatively simple COCO-Tasks dataset, Fast R-CNN and YOLO implementations achieved only 32-33\% $\rm mAP^{box}$. While ViTDet offered modest improvements, scaling from ViT-B to ViT-H backbone failed to overcome fundamental performance limitations.

The one-stage category, represented by \ac{mdetr}-based approaches, demonstrated notably stronger results. The baseline \ac{mdetr} implementation (row (h)) achieved 41.3\% $\rm mAP^{box}$ and 35.2\% $\rm mAP^{mask}$ on COCO-Tasks, suggesting that vision-language aligned features better capture object semantics across diverse scenarios. Importantly, this one-stage approach significantly outperformed its two-stage counterpart (row (d)), despite sharing the same \ac{mdetr} detection backbone. This performance gap highlights the advantages of end-to-end training for affordance reasoning tasks.

Building on these insights, we enhanced the baseline \ac{mdetr} with our proposed \ac{va} and \ac{bf} modules. This enhancement yielded substantial improvements across all datasets: COCO-Tasks performance increased to 43.2\% $\rm mAP^{box}$ and 36.9\% $\rm mAP^{mask}$, with similar gains on \affordanceDatasetName{} and \affordanceDatasetNamelvis{}. Our full \affordanceModelName{} framework, which integrates these modules with TOIST, further pushes performance boundaries, achieving 45.3\% $\rm mAP^{box}$ and 39.2\% $\rm mAP^{mask}$ on COCO-Tasks. Notably, \affordanceModelName{} maintains its leading position on the more challenging \affordanceDatasetNamelvis{} dataset, which features long-tail distributions and diverse affordance relationships, reaching 27.7\% $\rm mAP^{box}$ and 24.8\% $\rm mAP^{mask}$.

Qualitative results in \cref{fig:main_vis} reveal both strengths and limitations across all three datasets. While \affordanceModelName{} successfully identifies major affordance regions, it sometimes struggles with fine-grained functional elements crucial for complete affordance understanding. For instance, in the ``iron clothes with'' example, the model misses the iron's power cord, while the ``clean windows with'' case shows imprecise bounding box localization extending beyond the cleaning implement. A more detailed failure analysis is provided in~\cref{appendix:failure_analysis}.

\subsection{Ablation Study}\label{sec:exp:ablation}

We conduct comprehensive ablation studies to evaluate three critical components of \affordanceModelName{}: (i) distillation architecture, (ii) \ac{va} and \ac{bf} modules, and (iii) cluster number K. Using the TOIST framework as our baseline, we systematically analyze each component's contribution through extensive experiments.

\paragraph*{Knowledge Transfer through Distillation}

To address categorical bias in the COCO-Tasks dataset, we introduce a novel two-stage Noun-Pronoun Distillation framework. Rather than directly minimizing the distance between pronoun-based ($l_{\rm{pron}}^{\rm{tr}}$) and noun-based ($l_{\rm{noun}}^{\rm{tr}}$) representations, our framework strategically transfers knowledge through an intermediate feature space ($l_{c_s}^j$). This is achieved by first training a teacher model with explicit noun categories and then distilling this knowledge into a baseline model operating on verb-pronoun instructions. As shown in \cref{table:direct_distill}, this approach improves the verb-pronoun $\rm mAP^{box}$ from 41.3\% to 44.1\% compared to {\ac{mdetr} re-trained on COCO-Tasks with verb-pronoun inputs and without noun-pronoun distillation}. Notably, this sophisticated distillation strategy outperforms direct distillation between $l_{\rm{pron}}^{\rm{tr}}$ and $l_{\rm{noun}}^{\rm{tr}}$, confirming the effectiveness of our architectural design.

\begin{table}[ht!]
    \centering
    \small
    \caption{\textbf{Different distillation methods on COCO-Tasks.} Values in parentheses show improvements over the baseline.}
    \begin{tabular*}{\linewidth}{@{\extracolsep{\fill}}lll}
    \toprule
    \textbf{Method} & \multicolumn{1}{c}{$\rm{mAP}^{box}$}            & \multicolumn{1}{c}{$\rm{mAP}^{mask}$} \\ \midrule
    \ac{mdetr} & 41.3 & 35.2 \\
    distill from $l_{c_s}^j$ to $l_{\rm{pron}}^{\rm{tr}}$ &\textbf{44.1 (\textcolor{ForestGreen}{+2.8})} & \textbf{39.0 (\textcolor{ForestGreen}{+3.8})}  \\
    distill from $l_{\rm{noun}}^{\rm{tr}}$ to $l_{\rm{pron}}^{\rm{tr}}$ & { 41.9 (\textcolor{ForestGreen}{+0.6})}    & { 36.0 (\textcolor{ForestGreen}{+0.8})} \\
    \bottomrule
    \end{tabular*}
    \label{table:direct_distill}
\end{table}

To further refine feature grouping and preference scoring, we propose a \ac{cd} approach comprising three complementary components: \ac{ccr} for anchoring pronoun tokens, \ac{cl} to promote refined feature grouping, and \ac{sbtl} to improve preference modeling. As demonstrated in \cref{table:abl_loss}, while individual components provide modest improvements, their combination yields substantial gains of +2.8\% $\rm mAP^{box}$ and +3.8\% $\rm mAP^{mask}$ over the baseline. When integrated into the complete \affordanceModelName{} framework, these enhancements achieve state-of-the-art performance of 45.3\% $\rm mAP^{box}$ on verb-pronoun tasks.

\begin{table}[ht!]
    \centering
    \small
    \caption{\textbf{Ablation study of clustering distillation components on COCO-Tasks.} Each row shows performance impact when specific components are enabled (\checkmark) or disabled (\texttimes). \ac{ccr}: \acl{ccr}, \ac{cl}: \acl{cl}, \ac{sbtl}: \acl{sbtl}.}
    \begin{tabular*}{\linewidth}{@{\extracolsep{\fill}}cccccc}
    \toprule
    \multirow{2}{*}{\textbf{Index}} & \multicolumn{3}{c}{\textbf{Method Components}} & \multicolumn{2}{c}{\textbf{Performance}} \\
    \cmidrule(lr){2-4} \cmidrule(lr){5-6}
    & \textbf{CCR} & \textbf{CL} & \textbf{SBTL} & \textbf{mAP\textsuperscript{box}} & \textbf{mAP\textsuperscript{mask}} \\
    \midrule
    (a) & \texttimes & \texttimes & \texttimes & 41.3 & 35.2 \\
    (b) & \texttimes & \texttimes & \checkmark & 43.4 {\small(\textcolor{ForestGreen}{+2.1})} & 38.0 {\small(\textcolor{ForestGreen}{+2.8})} \\
    (c) & \texttimes & \checkmark & \texttimes & 42.0 {\small(\textcolor{ForestGreen}{+0.7})} & 37.1 {\small(\textcolor{ForestGreen}{+1.9})} \\
    (d) & \texttimes & \checkmark & \checkmark & 43.8 {\small(\textcolor{ForestGreen}{+2.5})} & 38.6 {\small(\textcolor{ForestGreen}{+3.4})} \\
    (e) & \checkmark & \texttimes & \texttimes & 42.0 {\small(\textcolor{ForestGreen}{+0.7})} & 37.0 {\small(\textcolor{ForestGreen}{+1.8})} \\
    (f) & \checkmark & \texttimes & \checkmark & 42.3 {\small(\textcolor{ForestGreen}{+1.0})} & 37.3 {\small(\textcolor{ForestGreen}{+2.1})} \\
    (g) & \checkmark & \checkmark & \texttimes & 42.3 {\small(\textcolor{ForestGreen}{+1.0})} & 37.5 {\small(\textcolor{ForestGreen}{+2.3})} \\
    (h) & \checkmark & \checkmark & \checkmark & \textbf{44.1} {\textbf{\small(\textcolor{ForestGreen}{+2.8})}} & \textbf{39.0} {\textbf{\small(\textcolor{ForestGreen}{+3.8})}} \\
    \bottomrule
    \end{tabular*}
    \label{table:abl_loss}
\end{table}

\paragraph*{Analysis of \ac{va} and \ac{bf} Modules}

\cref{tab:comparison_results} presents ablation results for the \ac{va} and \ac{bf} modules across three input settings: verb-pronoun, verb-noun, and distill. We evaluate their effectiveness by sequentially integrating them into our baseline.

\begin{table}[ht!]
    \centering
    \small
    \caption{\textbf{Ablation study of \affordanceModelName{} components.} Analysis showing performance impact of \acs{va} and \acs{bf} modules across verb-pronoun, verb-noun, and distillation settings. Values in parentheses indicate improvements over baseline configuration.}
    \begin{tabular*}{\linewidth}{@{\extracolsep{\fill}}cccccc}
    \toprule
    \multirow{2}{*}{\textbf{Index}} & \multicolumn{2}{c}{\textbf{Method}} & \multicolumn{3}{c}{\textbf{Performance}} \\
    \cmidrule(lr){2-3}\cmidrule(lr){4-6}
    & \textbf{VA} & \textbf{BF} & \textbf{verb-pronoun} & \textbf{verb-noun} & \textbf{distill} \\
    \midrule
    (a) & \texttimes & \texttimes & 41.3 & 53.2 & 44.1 \\
    (b) & \checkmark & \texttimes & 43.2 {\small(\textcolor{ForestGreen}{+1.9})} & 53.8 {\small(\textcolor{ForestGreen}{+0.6})} & 44.8 {\small(\textcolor{ForestGreen}{+0.7})} \\
    (c) & \texttimes & \checkmark & 43.0 {\small(\textcolor{ForestGreen}{+1.7})} & 53.9 {\small(\textcolor{ForestGreen}{+0.7})} & 44.6 {\small(\textcolor{ForestGreen}{+0.5})} \\
    (d) & \checkmark & \checkmark & \textbf{43.4} {\textbf{\small(\textcolor{ForestGreen}{+2.1})}} & \textbf{54.8} {\textbf{\small(\textcolor{ForestGreen}{+1.6})}} & \textbf{45.3} {\textbf{\small(\textcolor{ForestGreen}{+1.2})}} \\
    \bottomrule
    \end{tabular*}
    \label{tab:comparison_results}
\end{table}

\begin{figure}[t!]
    \centering
    \begin{subfigure}[t]{0.25\linewidth}
        \centering
        \caption*{Image}
        \includegraphics[width=\linewidth]{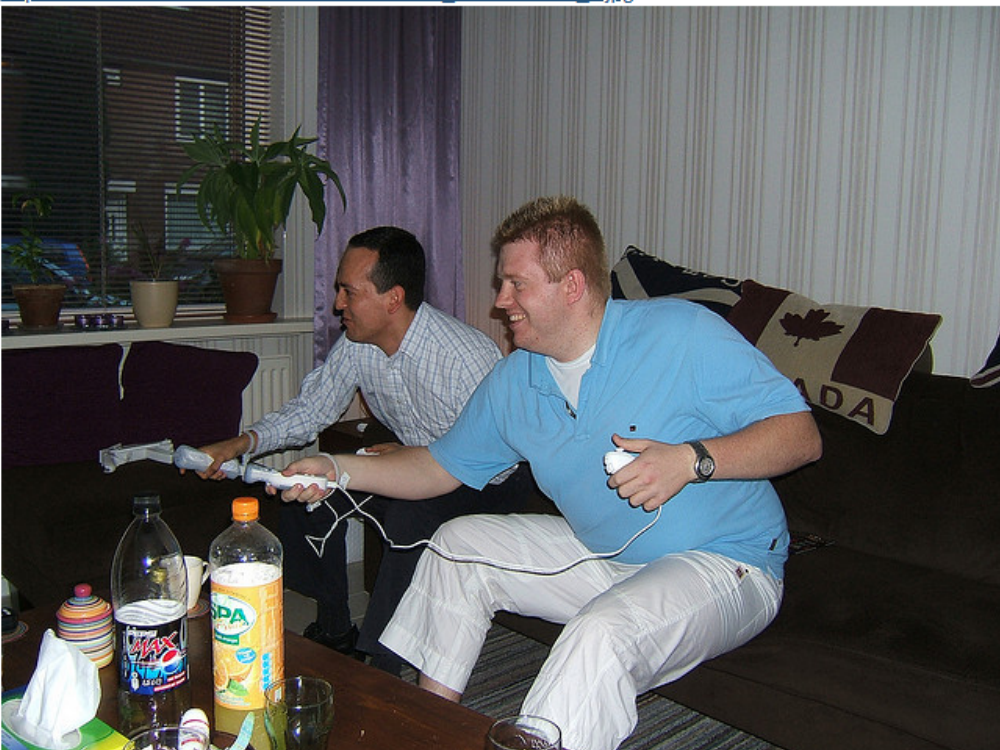}
        \caption*{Prompt:}
    \end{subfigure}%
    \begin{subfigure}[t]{0.25\linewidth}
        \centering
        \caption*{w/o Verb Attn}
        \includegraphics[width=\linewidth]{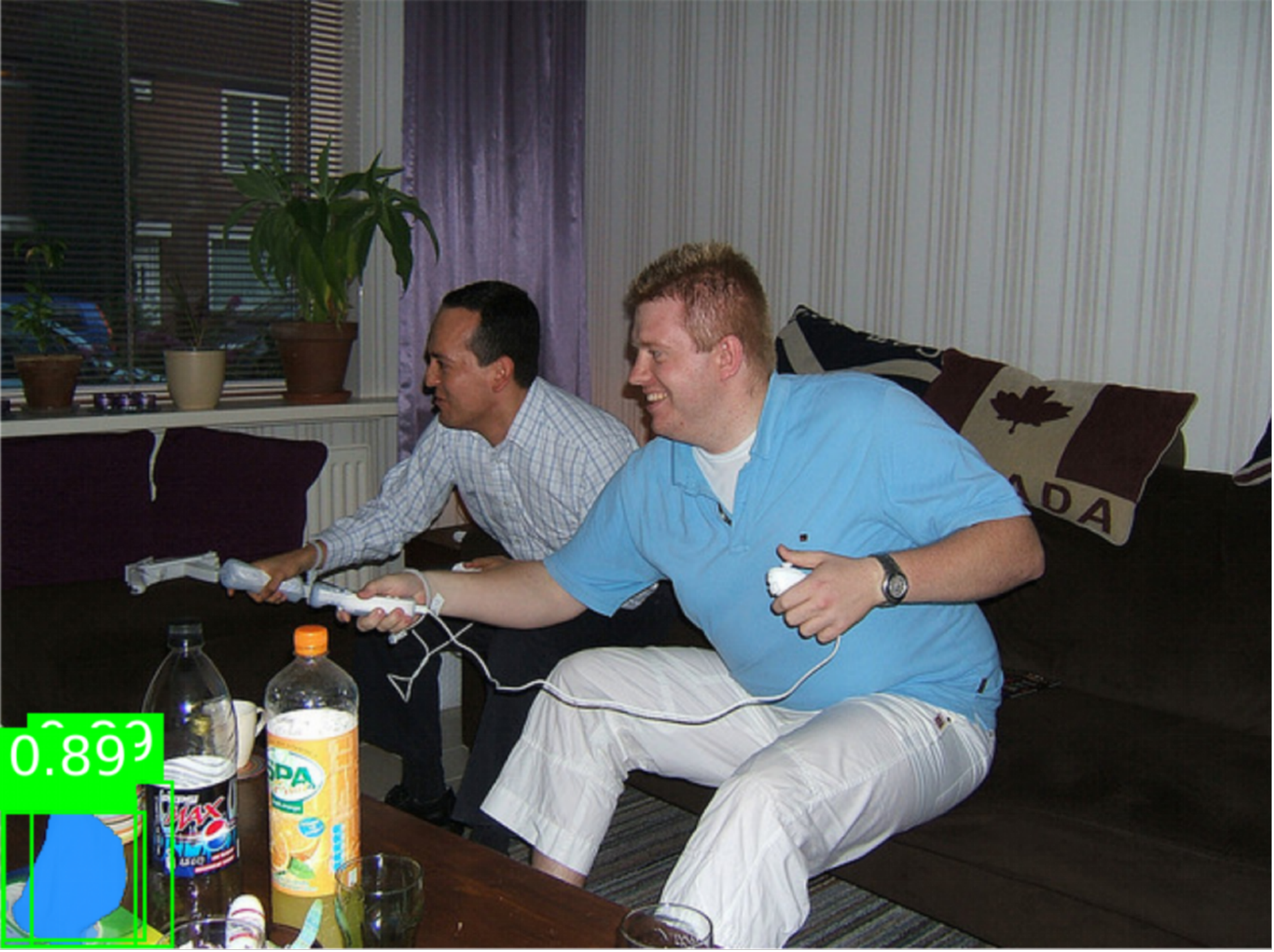}
        \caption*{clean with}
    \end{subfigure}%
    \begin{subfigure}[t]{0.25\linewidth}
        \centering
        \caption*{w/o Verb Attn}
        \includegraphics[width=\linewidth]{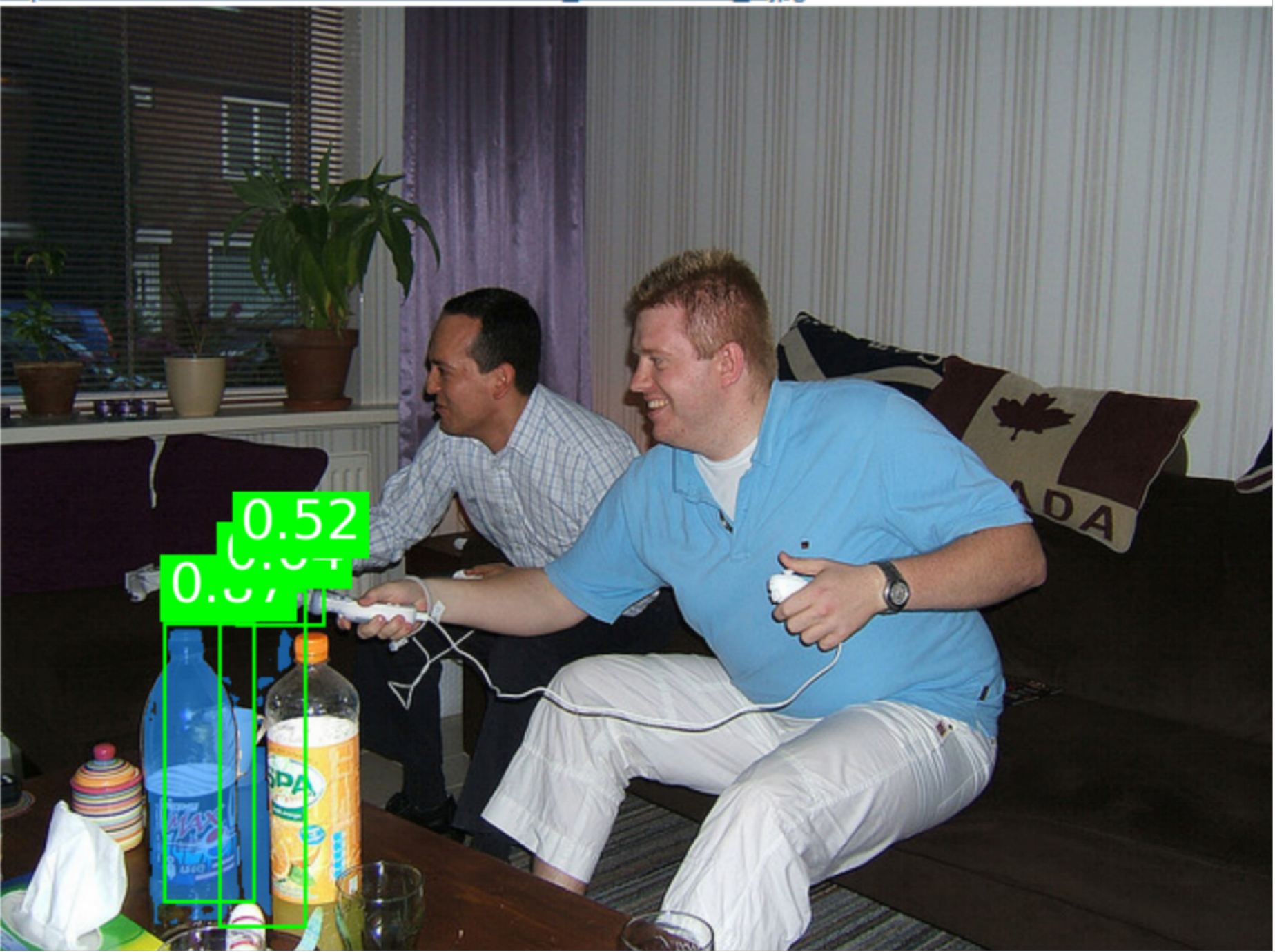}
        \caption*{clean \textcolor{gblue}{bottle} with}
    \end{subfigure}%
    \begin{subfigure}[t]{0.25\linewidth}
        \centering
        \caption*{w/ Verb Attn}
        \includegraphics[width=\linewidth]{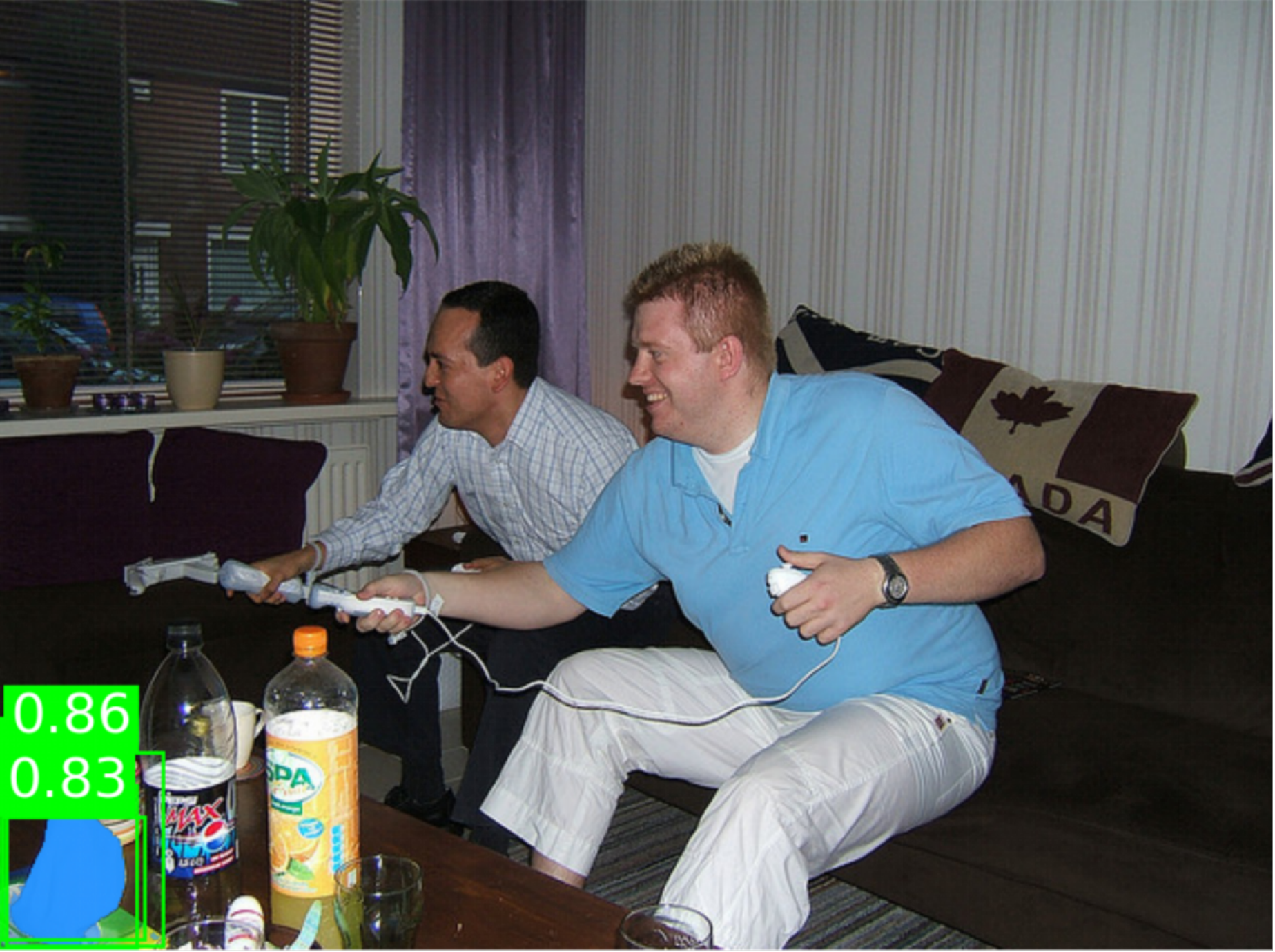}
        \caption*{clean \textcolor{gblue}{bottle} with}
    \end{subfigure}%
    \\%
    \begin{subfigure}[t]{0.25\linewidth}
        \centering
        \caption*{Image}
        \includegraphics[width=\linewidth]{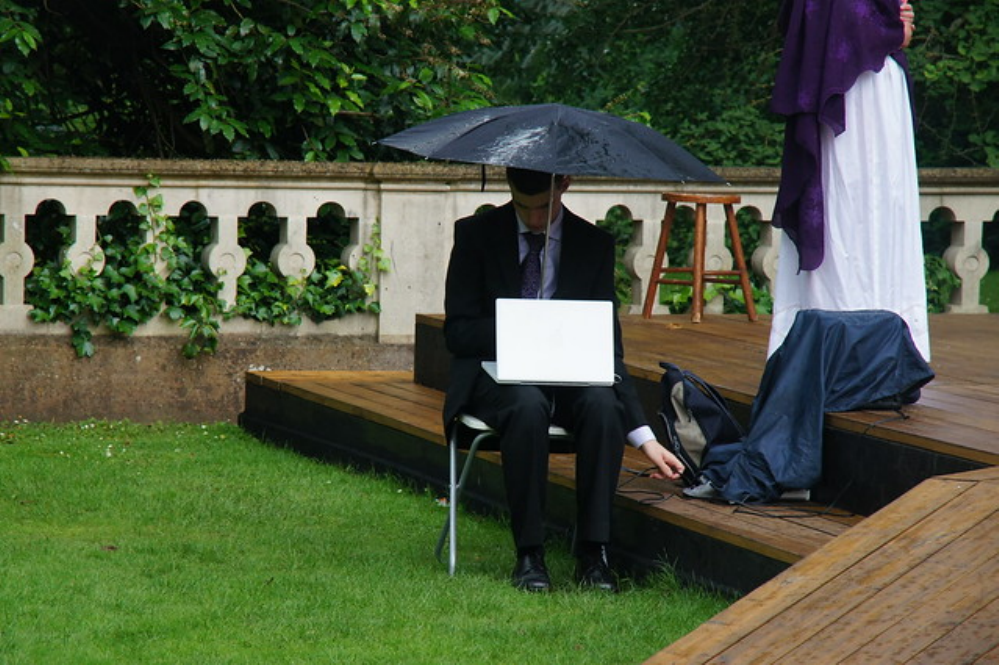}
        \caption*{Prompt:}
    \end{subfigure}%
    \begin{subfigure}[t]{0.25\linewidth}
        \centering
        \caption*{w/o Verb Attn}
        \includegraphics[width=\linewidth]{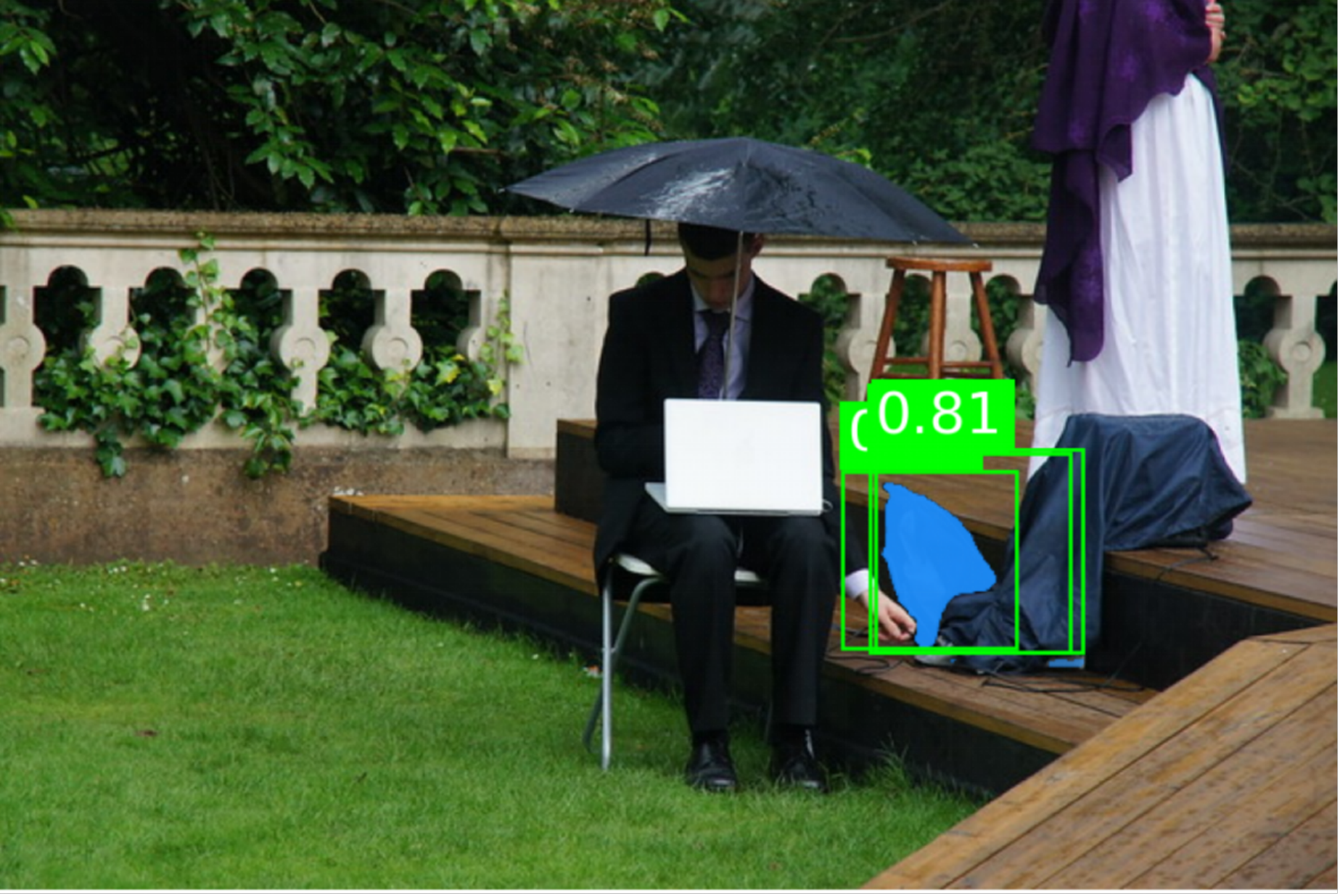}
        \caption*{take items in}
    \end{subfigure}%
    \begin{subfigure}[t]{0.25\linewidth}
        \centering
        \caption*{w/o Verb Attn}
        \includegraphics[width=\linewidth]{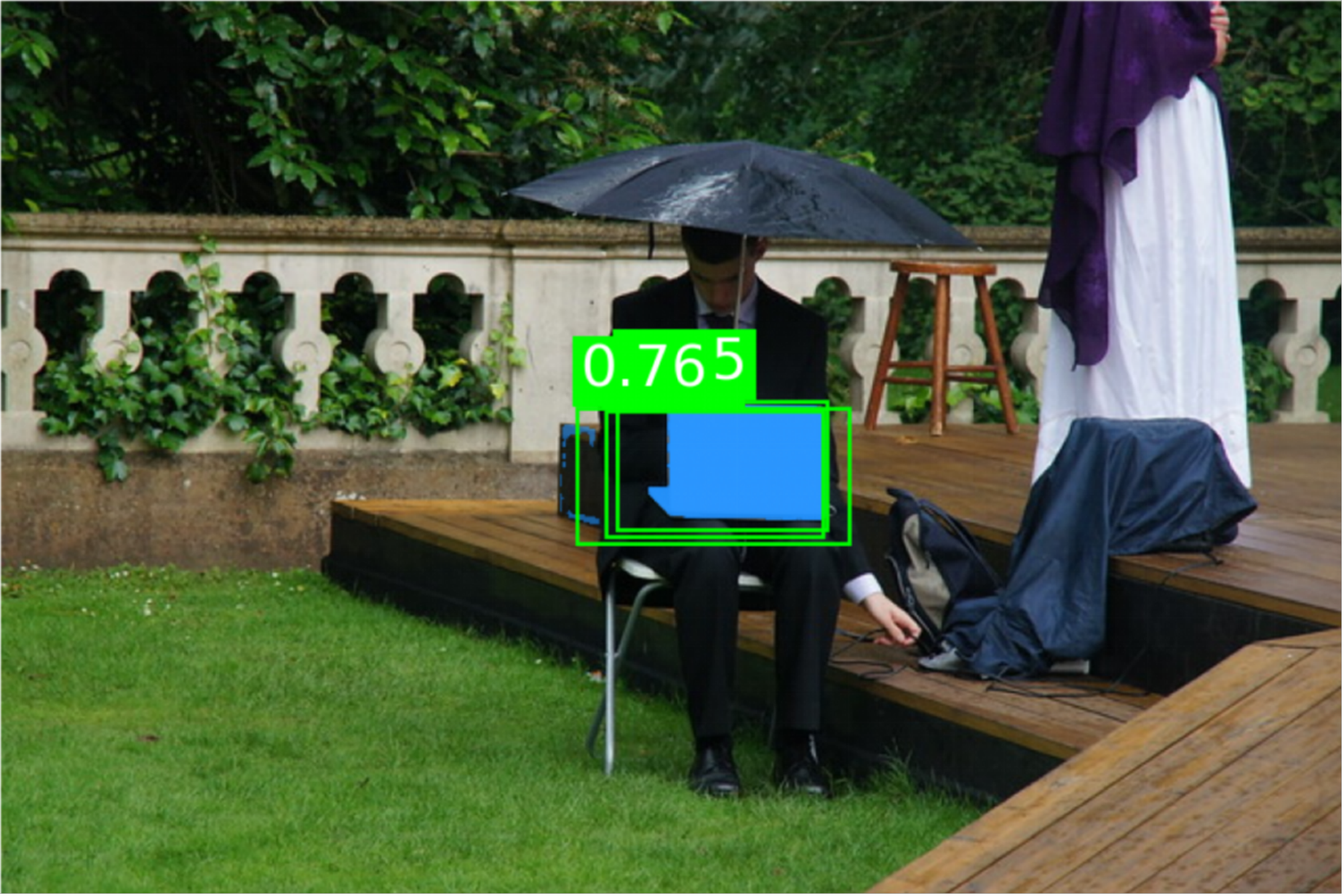}
        \caption*{take \textcolor{gblue}{computer} in}
    \end{subfigure}%
    \begin{subfigure}[t]{0.25\linewidth}
        \centering
        \caption*{w/ Verb Attn}
        \includegraphics[width=\linewidth]{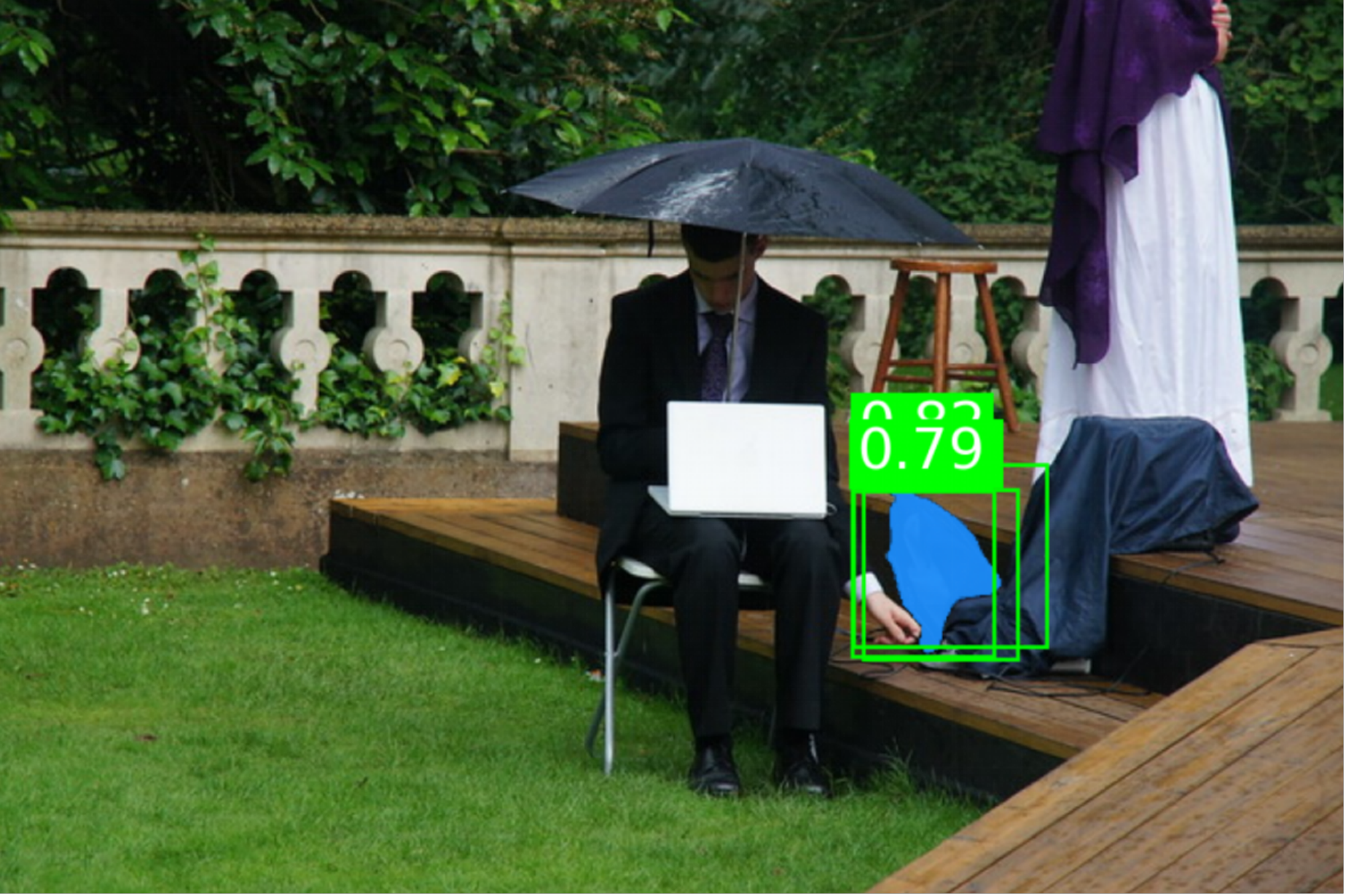}
        \caption*{take \textcolor{gblue}{computer} in}
    \end{subfigure}%
    \caption{\textbf{Impact of \ac{va} module on affordance reasoning performance.} The \ac{va} module reduces failures in affordance reasoning when text prompts include prominent objects. Column 1 shows the original images. Columns 2 and 3 present results without the module for prompts without and with object labels, respectively. Column 4 displays results with the module for prompts with object labels.}
    \label{fig:verb_attn}
\end{figure}

Individual module integration shows consistent improvements across all input configurations. The \ac{va} module (row (b)) yields gains of +1.9\%, +0.6\%, and +0.7\% across the three input types, demonstrating its effectiveness in emphasizing action-related cues while reducing noun-based interference. Similarly, the \ac{bf} module (row (c)) improves performance by +1.7\% and +0.7\% for pretrained student and teacher models respectively, with an additional +0.5\% gain during distillation, confirming that bidirectional cross-modal attention enhances visual-language alignment for affordance reasoning. The combination of both modules (row (d)) demonstrates synergistic benefits, achieving +2.1\% improvement for verb-pronoun input and +1.6\% for verb-noun input, surpassing individual module gains. While the joint implementation shows a more modest +1.2\% improvement under the distill setting, likely due to architectural constraints, the results validate the complementary nature of both modules in enhancing affordance reasoning.

Qualitative analysis in \cref{fig:verb_attn} illustrates the \ac{va} module's impact on attention mechanisms. The baseline model correctly identifies target objects for tasks like ``clean with'' (towel) and ``take item in'' (backpack) but becomes susceptible to interference from distracting nouns (bottle and computer). With the \ac{va} module, the model maintains accurate affordance reasoning while effectively filtering such distractions, providing visual confirmation of our quantitative findings and demonstrating enhanced functional reasoning capabilities. Additional qualitative visualizations and free-form task-description evaluations are provided in~\cref{apppendix:qualitative_results,appendix:free_form_robustness}.

\begin{figure}[b!]
    \centering
    \includegraphics[width=0.9\linewidth]{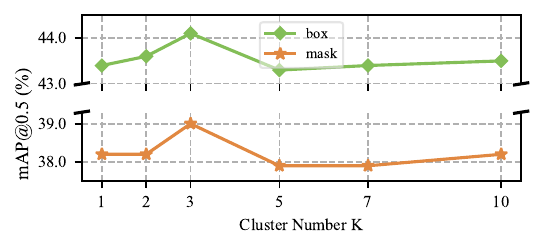}
    \caption{\textbf{Effects of clustering number.} A moderate number of clusters balances prototype diversity and stability.}
    \label{fig:abl_cluster}
\end{figure}

\paragraph*{Analysis of Cluster Number K}

\begin{figure*}[t!]
    \centering
    \begin{subfigure}{\linewidth}
        \centering
        \includegraphics[width=\linewidth]{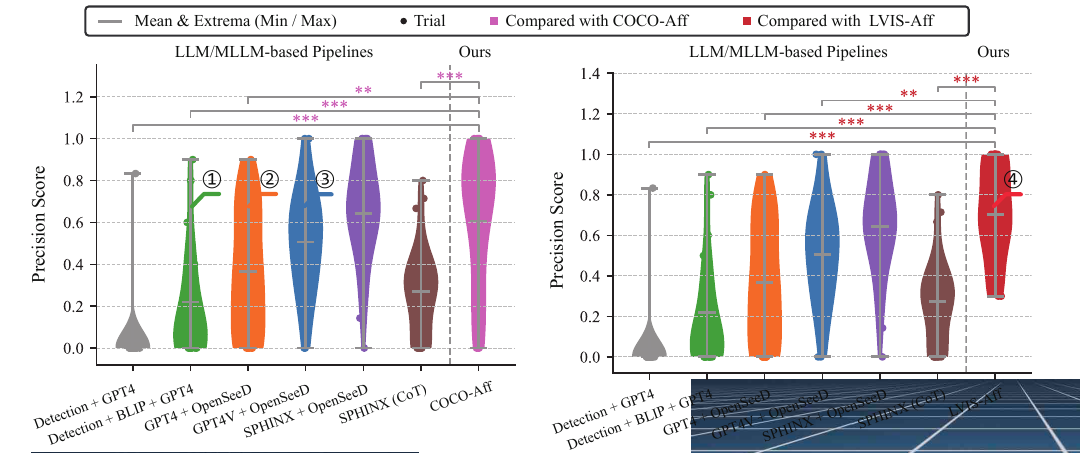}
        \caption{Precision comparison of different methods}
        \label{fig:metrics_a}
    \end{subfigure}
    \\
    \begin{subfigure}{\linewidth}
        \centering
        \includegraphics[width=\linewidth]{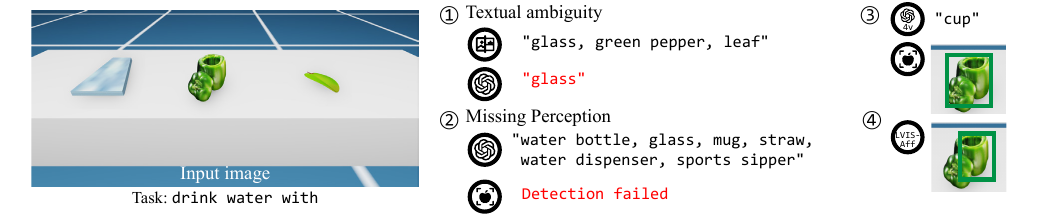}
        \caption{Analysis of method performance in affordance reasoning}
        \label{fig:metrics_b}
    \end{subfigure}
    \caption{\textbf{Comparison between \affordanceModelName{} and \ac{llm}/\ac{mllm}-based methods.} \affordanceModelName{} demonstrates superior performance in affordance reasoning through two key analyses. First, quantitative precision comparisons reveal significant advantages over \ac{llm}/\ac{mllm}-based methods, as shown in (a) the violin plot where statistical significance is denoted by stars. Second, (b) qualitative analysis highlights \affordanceModelName{}'s ability to overcome two fundamental limitations of these pipelines: textual ambiguity and limited visual grounding. While \acp{llm} can suffer from semantic uncertainty and \acp{mllm} may still struggle with precise task-relevant grounding, \affordanceModelName{} directly identifies and selects objects based on their functional affordances.}
    \label{fig:combined_metrics}
\end{figure*}

We systematically investigate optimal cluster numbers ($\rm{K}$) in the distillation framework; see also \cref{fig:abl_cluster}. We evaluate $\rm{K}$ values from 1 to 10, noting that higher values introduce excessive clustering complexity. Results show that all tested configurations improve upon the baseline model's performance (41.3\% $\rm{mAP^{box}}$, 35.2\% $\rm{mAP^{mask}}$). $\rm{K}=3$ emerges as the optimal choice, suggesting that a moderate number of clusters effectively balances feature aggregation and information preservation during knowledge distillation from noun features to the student model. {Specifically, a moderate $K$ balances prototype diversity and stability, whereas a larger $K$ over-fragments noun features and destabilizes prototypes for pronoun-based inference.}

\subsection{Dataset Analysis}\label{sec:data_analysis}

To examine how large-scale visual and linguistic information improves affordance reasoning, we analyze two key factors: \emph{task-scale} and \emph{category-scale}.

\paragraph*{Task-scale Analysis}

We evaluate models trained on three datasets of increasing scale: COCO-Tasks (14 tasks), \affordanceDatasetName{} (1,144 tasks), and \affordanceDatasetNamelvis{} (1,496 tasks). As shown in \cref{tab:task_scale}, while the COCO-Tasks-trained model achieves strong performance on seen tasks ($\rm{mAP}^{\rm{box}}=45.3\%$), it fails to generalize to unseen tasks ($\rm{mAP}^{\rm{box}}=1.6\%$). In contrast, training on \affordanceDatasetName{} substantially improves generalization to unseen tasks ($\rm{mAP}^{\rm{box}}=24.5\%$), with further gains achieved by \affordanceDatasetNamelvis{} ($\rm{mAP}^{\rm{box}}=26.3\%$), demonstrating the benefits of expanded task and object coverage.

\begin{table}[ht!]
    \centering
    \small
    \caption{\textbf{Task-scale analysis.} Comparison of model performance on 14 \textit{seen tasks} from COCO-Tasks and 80 \textit{unseen tasks} from LVIS.}
    \label{tab:task_scale}
    \begin{tabular*}{\linewidth}{@{\extracolsep{\fill}}lcccccc}
        \toprule
        \multirow{2}{*}{\textbf{Training Data.}} & \multicolumn{2}{c}{\textbf{Seen-Tasks (14)}} & \multicolumn{2}{c}{\textbf{Unseen-Tasks (80)}} \\
        \cmidrule(lr){2-3} \cmidrule(lr){4-5} \cmidrule(lr){6-7}
         & $\rm{mAP}^{\rm{box}}$ & $\rm{mAP}^{\rm{mask}}$ & $\rm{mAP}^{\rm{box}}$ & $\rm{mAP}^{\rm{mask}}$ &  \\
        \midrule
        COCO-Tasks & 45.3 & 39.2 & 1.6 & 1.3  \\
        \affordanceDatasetName{} & 43.9 & 38.3 & 24.5 & 24.1\\
        \affordanceDatasetNamelvis{} & --- & --- & 26.3 & 25.6\\
        \bottomrule
    \end{tabular*}
\end{table}

\paragraph*{Category-scale Analysis}

We further evaluate generalization using 40 unseen tasks: 20 with common categories (present in both COCO and LVIS) and 20 with novel categories (LVIS-exclusive). As shown in \cref{tab:object_categories}, the \affordanceDatasetName{}-trained model performs well on common category tasks (35.0\% $\rm{mAP}^{\rm{box}}$) but struggles with novel categories (4.9\% $\rm{mAP}^{\rm{box}}$). Models trained on \affordanceDatasetNamelvis{} show improved performance across both task types (37.2\% and 8.3\% $\rm{mAP}^{\rm{box}}$ respectively), demonstrating that expanded category coverage enhances both affordance reasoning and generalization.

\begin{table}[ht!]
    \centering
    \small
    \caption{\textbf{Category-scale analysis.} Performance comparison between \affordanceDatasetName{} and \affordanceDatasetNamelvis{} trained models on unseen tasks involving common categories (shared between COCO/LVIS) and novel categories (LVIS-exclusive).}
    \label{tab:object_categories}
    \begin{tabular*}{\linewidth}{@{\extracolsep{\fill}}lcccc}
        \toprule
        \multirow{2}{*}{\textbf{Training Data.}} & \multicolumn{2}{c}{\textbf{Common categories}} & \multicolumn{2}{c}{\textbf{Novel categories}} \\
        \cmidrule(lr){2-3} \cmidrule(lr){4-5}
         & $\rm{mAP}^{\rm{box}}$ & $\rm{mAP}^{\rm{mask}}$ & $\rm{mAP}^{\rm{box}}$ & $\rm{mAP}^{\rm{mask}}$ \\
        \midrule
        COCO-Aff & 35.0 & 33.9 & 4.9 & 4.8 \\
        LVIS-Aff & 37.2 & 35.3 & 8.3 & 8.2 \\
        \bottomrule
    \end{tabular*}
\end{table}

\begin{figure*}[b!]
    \centering
    \includegraphics[width=\linewidth]{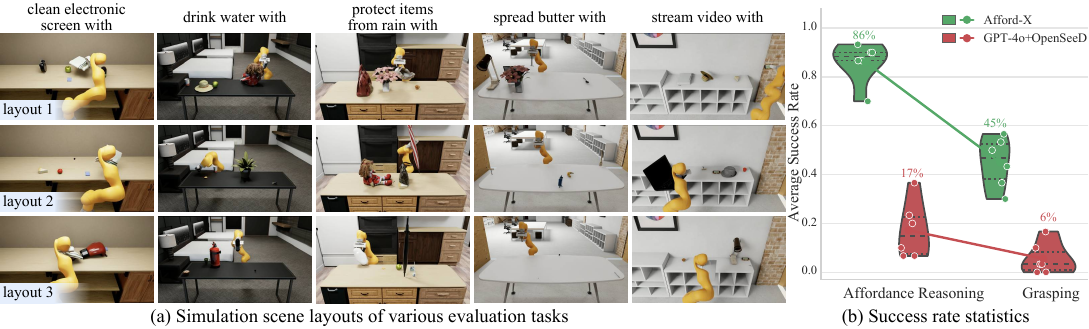}
    \caption{{\textbf{Task-oriented object grasping across different layouts in simulated environments.} (a) As a showcase for task-oriented manipulation, the robotic system integrated with \affordanceModelName{} successfully identifies the appropriate object across various spatial and object configurations in simulated environments, and executes grasping as a showcase of task-oriented manipulation. (b) Success rates of each step throughout the manipulation trial demonstrate Afford-X's capability in supporting task-oriented manipulation in simulated environments.}}
    \label{fig:sim_scenes_layout}
\end{figure*}

\subsection{Affordance Reasoning in Simulated 3D Scenes}\label{sec:exp:sructure}

We evaluate \affordanceModelName{}'s potential as an alternative to \ac{llm}/\ac{mllm}-based pipelines in robotic deployments through extensive testing in simulated 3D environments with textured object meshes, as described in \cref{sec:inscene}. Our evaluation compares baseline methods trained on \affordanceDatasetName{} and \affordanceDatasetNamelvis{} against three categories of \ac{llm}/\ac{mllm}-based pipelines: (i) Detect before Reasoning ({rows} (a) and (b)), where object detection precedes \ac{llm} processing; (ii) Reason before Detection (rows (c)-(e)), where an \ac{llm} or \ac{mllm} identifies {task-relevant candidate nouns or object categories before localization with OpenSeeD}; and (iii) Simultaneous Perception and Reasoning ({row (f)}), where an \ac{mllm} jointly processes the task and image for direct bounding box output. Additional comparisons with \ac{mllm}-based pipelines are provided in~\cref{appendix:mllm_comparison}.

Our experimental framework encompasses 50 tasks (30 seen, 20 unseen) in simulated environments, with each task evaluated across 10 images containing three candidate objects---one capable of affording the task and two non-affording objects randomly selected from the candidate pool. Success criteria require bounding box predictions to achieve mIoU above 0.5, as shown in \cref{fig:combined_metrics}, where the ``Detection'' methods utilize RAM++~\cite{zhang2024recognize} and Grounding Dino~\cite{liu2023grounding} for object detection based on given labels. To ensure practical relevance, all evaluations were conducted on a single 24GB RTX 3090 GPU, comparing API-based services with memory-constrained ($\leq$24GB) SPHINX tiny models while accounting for API communication latency in GPT-4 implementations. The \ac{fps} and parameter sizes for each baseline are detailed in \cref{tab:method_comparison}, where $>$ indicates API usage, excluding these model parameters from statistics. {Notably, our model requires only 3.2 GB locally, far below SPHINX (10–11.5 GB), leaving ample headroom for concurrent robotic modules.}

\begin{table}[ht!]
    \centering
    \footnotesize
    \caption{\textbf{Computational efficiency comparison.} Analysis of \acs{fps}, parameters, and GPU memory across \ac{llm}/\ac{mllm}-based pipelines and our proposed approaches, measured on a standard NVIDIA 3090 GPU (24GB). {SPHINX parameters include the visual encoder, and GPU Mem. denotes local deployment memory.}}
    \setlength{\tabcolsep}{2pt}
    \begin{tabular*}{\linewidth}{@{\extracolsep{\fill}}c>{\raggedright\arraybackslash}p{0.42\linewidth}ccc@{}}
        \toprule
        \textbf{Index} & \textbf{Method}  & \textbf{\acs{fps}} & \textbf{Parameters} & {\textbf{GPU Mem.}} \\
        \midrule
        (a) & Detection + GPT-4             & 1.18  & $>$369M & {---} \\
        (b) & Detection + BLIP~\cite{li2022blip} + GPT-4      & 0.27  & $>$498M & --- \\
        (c) & GPT-4 + OpenSeeD~\cite{zhang2023simple}         & 0.11  & $>$116M & {$>1.3$\,GB} \\
        (d) & GPT-4V + OpenSeeD        & 0.04  & $>$116M & {$>1.3$\,GB} \\
        (e) & SPHINX ~\cite{lin2023sphinx} + OpenSeeD       & 0.32  & {3.2B} & {11.5\,GB} \\
        (f) & SPHINX (CoT)            & 0.34  & {3.1B} & {10.0\,GB} \\
        \midrule
        (g) & \affordanceDatasetName{}                & \textbf{2.38}  & \textbf{187M} & {\textbf{3.2}\,GB} \\
        (h) & \affordanceDatasetNamelvis{}             & \textbf{2.38}  & \textbf{187M} & {\textbf{3.2}\,GB} \\
        \bottomrule
    \end{tabular*}
    \vspace{2pt}
    \scriptsize
    \label{tab:method_comparison}
\end{table}

The experimental results revealed significant performance variations across pipeline architectures. \Ac{llm}-based label reasoning pipelines demonstrated notably poor performance (2.33\% recall, 3.44\% mAP), primarily due to semantic ambiguity in object labeling---exemplified by cases where distinct tools like hammers and chisels share generic labels, compromising reasoning accuracy. While image captioning integration showed improvements, limited caption semantic density continued to impede accurate object identification.

\begin{figure*}[t!]
    \centering
    \begin{subfigure}{0.65\linewidth}
        \centering
        \includegraphics[width=5.02in, height=2in]{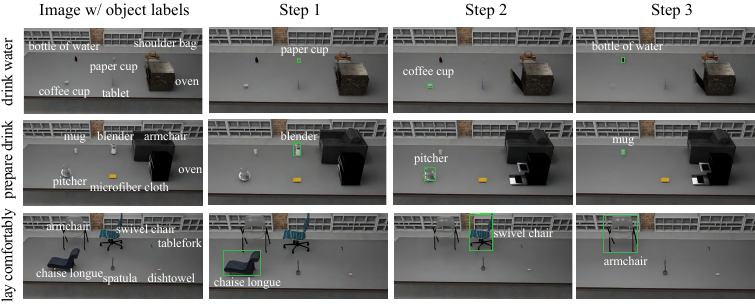}
        \caption{\affordanceModelName{}'s object selection preferences.}
        \label{fig:sim_object_selection}
    \end{subfigure}
    \hfill
    \begin{subfigure}{0.3\linewidth}
        \centering
        \includegraphics[height=2in]{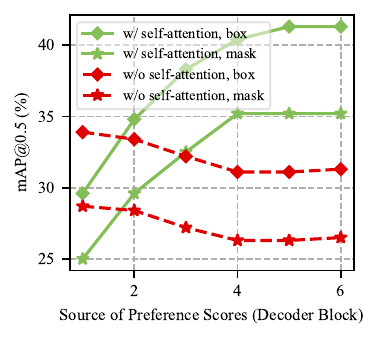}
        \caption{Impact of self-attention.}
        \label{fig:abl_preference}
    \end{subfigure}
    \caption{\textbf{Sequential object preference and self-attention analysis.} (a) We demonstrate \affordanceModelName{}'s affordance-based selection process across three sequential steps. The model exhibits hierarchical preferences as objects are progressively removed from the environment. Each experimental row presents a different task scenario where available objects are labeled to show possible choices. (b) The performance curves reveal the significant impact of self-attention layers on preference modeling. Our analysis shows that models with self-attention consistently achieve higher accuracy across decoder blocks compared to variants without this mechanism.}
    \label{fig:combined_visualization}
\end{figure*}

Reason-before-detection pipelines showed improved results, with GPT-4V + OpenSeeD achieving 49.61\% recall and 50.63\% mAP. However, this performance remained slightly below the SPHINX model, potentially due to GPT-4V's inherent value-based constraints on safety and societal norms, particularly evident in scenarios like ``hold water'' tasks where unconventional but viable solutions like bamboo tubes might be overlooked. End-to-end \ac{mllm} systems, while promising, faced practical limitations---the memory-constrained SPHINX tiny model required supplementary detection algorithms or chain-of-thought prompting for bounding box output, achieving 39.76\% mAP at 0.49 \ac{fps}, indicating limited practical utility.

In contrast, \affordanceModelName{} demonstrated superior performance through direct bounding box output, achieving 60.67\% mAP on \affordanceDatasetName{} and 67.26\% mAP on \affordanceDatasetNamelvis{} while maintaining 2.38 \ac{fps}. These results emphasize the importance of specialized task-oriented datasets for training efficient end-to-end models. Qualitative analysis through four representative case studies revealed the limitations of existing approaches---detection models struggled with semantic ambiguity (particularly in ``glass'' interpretation between structural glass and drinking vessels), while \acp{llm} without visual perception showed a tendency to infer non-existent objects. Both \ac{mllm}-based pipelines and our \affordanceModelName{} exhibited advanced affordance understanding, successfully identifying unconventional affordances, such as recognizing modified green peppers as viable containers.

This comprehensive evaluation demonstrates that domain-specific training can outperform generic pre-trained \acp{mllm} for affordance reasoning tasks, suggesting a promising direction for practical robotic applications.

\subsection{{Task-oriented Object Grasping}}\label{sec:manipulation}

We have conducted comprehensive evaluations through a series of systematic experiments in simulated environments, designed to assess \affordanceModelName{}'s adaptability in diverse indoor contexts, utilizing controlled virtual environments to ensure precise manipulation of experimental variables, including object placement, scene complexity, and task parameters. 

\paragraph*{Robot Evaluation in Indoor Scenarios}

Our primary evaluation focuses on the robot's performance across distinct indoor settings---dining room, living room, and office---each rendered in multiple stylistic variants with varying object configurations. We tasked the robot with performing affordance-based grasping (\eg, ``drink water with'') that align with typical user interactions in residential and workplace settings. \cref{fig:sim_scenes_layout,fig:sim_scenes} illustrates the results in a table-top manipulation setting using a fixed KUKA IIWA-Panda robot arm. The manipulation is performed within various scenes featuring substantial variation in object placement, room layouts, and environmental complexity, providing a comprehensive test of our model's adaptability.

{\cref{fig:sim_scenes_layout}(b) reports the cumulative success rates of the manipulation pipeline over 150 trials (6 tasks $\times$ 5 scenes $\times$ 5 random object layouts), where we compare Afford-X against the powerful baseline ``GPT-4o + OpenSeeD''. The success rates are evaluated for two key steps: affordance reasoning (step i in~\cref{sec:tom}), and grasping (steps ii-iv in~\cref{sec:tom}). For Afford-X, affordance reasoning succeeds in 86\% of the trials, showing that Afford-X can reliably select and segment task-relevant objects in 3D simulated environments. The grasp execution succeeds in 45\% of the trials (52.3\% of trials with successful affordance reasoning), where most failures occur during grasp planning and execution (\eg, grasping thin and flat objects like spoon and fork) that are challenging for a generic grasp pose planner. The baseline shows 17\% and 6\% success rates in the two steps. These results demonstrate that Afford-X effectively provides object-level grounding for task-oriented object grasping, and reveals the valuable future study of integrating finer part-level affordance reasoning and task-specific manipulation strategies for more advanced task-oriented manipulation.} Further, \cref{appendix:tom:long_horizon_tasks} extends this manipulation to identifying and collecting appropriate objects for a complex task in a larger interior scene with more objects in sight. After decomposing the complex task into multiple steps with an LLM-based task planner, a mobile base manipulator grasps and collects the objects identified by \affordanceModelName{}.

\paragraph*{Preference on Multiple Options}

In real-world scenarios where multiple options are available for the same task, \affordanceModelName{} demonstrates sophisticated preference modeling aligned with human utility patterns. For instance, when prompted with ``\textit{prepare drinks with},'' the model exhibits hierarchical understanding of utility, prioritizing a blender, followed by a pitcher, and finally a mug. To systematically evaluate this capability, we designed an experimental setup featuring six candidate objects placed on a large table in an interior scene, including three positive object categories for specific tasks (\eg, ``\textit{prepare drinks with}'' or ``\textit{lay comfortably on}''). Through an iterative elimination procedure, the robot selects the object with the highest score, which is subsequently removed from the environment for the next query. This process, illustrated in \cref{fig:sim_object_selection}, thoroughly tests the model's ability to dynamically adapt its preferences. Similarly, for the task ``\textit{lay comfortably on},'' the model demonstrates nuanced understanding by favoring a lounge over a chair, indicating awareness of comfort-oriented furniture hierarchies.

The model's sophisticated preference modeling capability is fundamentally enabled by self-attention (SA) layers in the decoder. To validate this mechanism, we conducted a comparative analysis between two model variants—one incorporating SA layers and one without—while maintaining identical parameter budgets to isolate the SA component's contribution. Using bounding boxes and masks generated by the final decoder block, we computed mean Average Precision (mAP) values based on preference scores from each intermediate decoder layer. The SA-equipped model demonstrates progressive improvement through deeper decoder layers, achieving final scores of 41.3\% ($\rm{mAP^{box}}$) and 35.2\% ($\rm{mAP^{mask}}$) from initial values of 29.6\% and 25.0\%, respectively. In contrast, the non-SA version shows limited improvement, reaching maximum values of only 33.9\% ($\rm{mAP^{box}}$) and 28.7\% ($\rm{mAP^{mask}}$). These results directly implicate SA layers as crucial components in capturing and refining pairwise preferences—critical for strategic decision-making during progressive selection.

For comprehensive results and detailed analyses beyond the scope of this section, readers are directed to \cref{appendix:tom}.

\section{Conclusion}\label{sec:conclusion}

We present \affordanceModelName{}, a novel framework addressing a fundamental challenge in robotics: achieving sophisticated affordance reasoning while maintaining computational efficiency for local deployment. Our approach integrates a noun-pronoun distillation pipeline with specialized \ac{va} and \ac{bf} modules, complemented by an automated pipeline for constructing comprehensive affordance reasoning datasets—\affordanceDatasetName{} and \affordanceDatasetNamelvis{}. Through extensive experimentation, we demonstrate our framework's enhanced generalization capabilities while maintaining real-time processing speeds, validated through robotic manipulation experiments in simulated environments.

Our current framework faces two primary limitations. First, geometric features alone prove insufficient for certain affordance reasoning scenarios—for instance, distinguishing between drinking cups and toothbrush holders despite identical geometric features, where usage context and hygiene considerations necessitate different affordances. Second, the lack of 3D information constrains system performance in complex spatial environments, particularly in assessing object relationships and occlusion scenarios. Future work could address them through integration of contextual knowledge bases and incorporation of depth information to enhance spatial understanding.

\setstretch{0.975}
\bibliography{reference_header_shorter,reference}

@article{zhu2020dark,
  title={Dark, beyond deep: A paradigm shift to cognitive ai with humanlike common sense},
  author={Zhu, Yixin and Gao, Tao and Fan, Lifeng and Huang, Siyuan and Edmonds, Mark and Liu, Hangxin and Gao, Feng and Zhang, Chi and Qi, Siyuan and Wu, Ying Nian and others},
  journal={Engineering},
  volume={6},
  number={3},
  pages={310--345},
  year={2020},
  publisher={Elsevier}
}

@inproceedings{sundaralingam2023curobo,
  title={Curobo: Parallelized collision-free robot motion generation},
  author={Sundaralingam, Balakumar and Hari, Siva Kumar Sastry and Fishman, Adam and Garrett, Caelan and Van Wyk, Karl and Blukis, Valts and Millane, Alexander and Oleynikova, Helen and Handa, Ankur and Ramos, Fabio and others},
  booktitle=ICRA,
  year={2023},
  organization={IEEE}
}

@inproceedings{zhang2025iaao,
  title={Iaao: Interactive affordance learning for articulated objects in 3d environments},
  author={Zhang, Can and Lee, Gim Hee},
  booktitle=CVPR,
  year={2025}
}

@inproceedings{wu2025open,
  title={Open-Vocabulary 3D Affordance Understanding via Functional Text Enhancement and Multilevel Representation Alignment},
  author={Wu, Lin and Wei, Wei and Yu, Peizhuo and Lan, Jianglin},
  booktitle={Proceedings of the 33rd ACM International Conference on Multimedia},
  year={2025}
}

@inproceedings{radford2021learning,
  title={Learning transferable visual models from natural language supervision},
  author={Radford, Alec and Kim, Jong Wook and Hallacy, Chris and Ramesh, Aditya and Goh, Gabriel and Agarwal, Sandhini and Sastry, Girish and Askell, Amanda and Mishkin, Pamela and Clark, Jack and others},
  booktitle=ICML,
  year={2021}
}

@inproceedings{li2022toist,
  title={Toist: Task oriented instance segmentation transformer with noun-pronoun distillation},
  author={Li, Pengfei and Tian, Beiwen and Shi, Yongliang and Chen, Xiaoxue and Zhao, Hao and Zhou, Guyue and Zhang, Ya-Qin},
  booktitle=NIPS,
  year={2022}
}

@inproceedings{lin2014microsoft,
  title={Microsoft coco: Common objects in context},
  author={Lin, Tsung-Yi and Maire, Michael and Belongie, Serge and Hays, James and Perona, Pietro and Ramanan, Deva and Doll{\'a}r, Piotr and Zitnick, C Lawrence},
  booktitle=ECCV,
  year={2014}
}

@article{zhai2022one,
  title={One-shot object affordance detection in the wild},
  author={Zhai, Wei and Luo, Hongchen and Zhang, Jing and Cao, Yang and Tao, Dacheng},
  journal=IJCV,
  volume={130},
  number={10},
  pages={2472--2500},
  year={2022},
  publisher={Springer}
}

@inproceedings{zhu2015understanding,
  title={Understanding tools: Task-oriented object modeling, learning and recognition},
  author={Zhu, Yixin and Zhao, Yibiao and Chun Zhu, Song},
  booktitle=CVPR,
  year={2015}
}

@article{fang2020learning,
  title={Learning task-oriented grasping for tool manipulation from simulated self-supervision},
  author={Fang, Kuan and Zhu, Yuke and Garg, Animesh and Kurenkov, Andrey and Mehta, Viraj and Fei-Fei, Li and Savarese, Silvio},
  journal=IJRR,
  volume={39},
  number={2-3},
  pages={202--216},
  year={2020},
  publisher={SAGE Publications Sage UK: London, England}
}

@inproceedings{xu2022partafford,
  title={PartAfford: Part-level Affordance Discovery from 3D Objects},
  author={Xu, Chao and Chen, Yixin and Wang, He and Zhu, Song-Chun and Zhu, Yixin and Huang, Siyuan},
  booktitle={ECCV VOLI Workshop},
  year={2022}
}

@inproceedings{do2018affordancenet,
  title={Affordancenet: An end-to-end deep learning approach for object affordance detection},
  author={Do, Thanh-Toan and Nguyen, Anh and Reid, Ian},
  booktitle=ICRA,
  year={2018}
}

@inproceedings{tang2023cotdet,
  title={Cotdet: Affordance knowledge prompting for task driven object detection},
  author={Tang, Jiajin and Zheng, Ge and Yu, Jingyi and Yang, Sibei},
  booktitle=ICCV,
  year={2023}
}

@inproceedings{qian2024affordancellm,
  title={Affordancellm: Grounding affordance from vision language models},
  author={Qian, Shengyi and Chen, Weifeng and Bai, Min and Zhou, Xiong and Tu, Zhuowen and Li, Li Erran},
  booktitle=CVPR,
  year={2024}
}

@inproceedings{wei2022chain,
  title={Chain-of-thought prompting elicits reasoning in large language models},
  author={Wei, Jason and Wang, Xuezhi and Schuurmans, Dale and Bosma, Maarten and Xia, Fei and Chi, Ed and Le, Quoc V and Zhou, Denny and others},
  booktitle=NIPS,
  year={2022}
}

@incollection{ikeuchi1996task,
  title={Task-oriented vision},
  author={Ikeuchi, Katsushi and Hebert, Martial},
  booktitle={Exploratory vision: the active eye},
  pages={257--277},
  year={1996},
  publisher={Springer}
}

@article{gibson1979theory,
  title={The theory of affordances. The ecological approach to visual perception},
  author={Gibson, James J},
  journal={The people, place and, space reader},
  pages={56--60},
  year={1979},
  publisher={Routledge New York, NY, USA}
}

@inproceedings{su2019vl,
  title={Vl-bert: Pre-training of generic visual-linguistic representations},
  author={Su, Weijie and Zhu, Xizhou and Cao, Yue and Li, Bin and Lu, Lewei and Wei, Furu and Dai, Jifeng},
  booktitle=ICLR,
  year={2019}
}

@inproceedings{lu2019vilbert,
  title={Vilbert: Pretraining task-agnostic visiolinguistic representations for vision-and-language tasks},
  author={Lu, Jiasen and Batra, Dhruv and Parikh, Devi and Lee, Stefan},
  booktitle= NIPS,
  year={2019}
}

@inproceedings{chen2020uniter,
  title={Uniter: Universal image-text representation learning},
  author={Chen, Yen-Chun and Li, Linjie and Yu, Licheng and El Kholy, Ahmed and Ahmed, Faisal and Gan, Zhe and Cheng, Yu and Liu, Jingjing},
  booktitle=ECCV,
  year={2020}
}

@inproceedings{lu202012,
  title={12-in-1: Multi-task vision and language representation learning},
  author={Lu, Jiasen and Goswami, Vedanuj and Rohrbach, Marcus and Parikh, Devi and Lee, Stefan},
  booktitle=CVPR,
  year={2020}
}

@inproceedings{kamath2021mdetr,
  title={Mdetr-modulated detection for end-to-end multi-modal understanding},
  author={Kamath, Aishwarya and Singh, Mannat and LeCun, Yann and Synnaeve, Gabriel and Misra, Ishan and Carion, Nicolas},
  booktitle=ICCV,
  year={2021}
}

@article{wang2023task,
  title={Task-Oriented Robot Cognitive Manipulation Planning Using Affordance Segmentation and Logic Reasoning},
  author={Wang, Zhongli and Tian, Guohui},
  journal={IEEE Transactions on Neural Networks and Learning Systems},
  year={2023},
  publisher={IEEE}
}

@inproceedings{zhu2014reasoning,
  title={Reasoning about object affordances in a knowledge base representation},
  author={Zhu, Yuke and Fathi, Alireza and Fei-Fei, Li},
  booktitle=ECCV,
  year={2014}
}

@inproceedings{chuang2018adeaffordance,
  title={Learning to Act Properly: Predicting and Explaining Affordances from Images},
  author={Ching-Yao Chuang and Jiaman Li and A. Torralba and S. Fidler},
  booktitle=CVPR,
  year={2018},
}

@inproceedings{luo2021one,
  title={One-shot affordance detection},
  author={Luo, Hongchen and Zhai, Wei and Zhang, Jing and Cao, Yang and Tao, Dacheng},
  booktitle=IJCAI,
  year={2021}
}

@inproceedings{xin2022cocotasks,
  title={A Visual Affordance Reasoning Network Based on Graph Attention},
  author={Jianjia Xin and Lichun Wang and Shaofan Wang and Dehui Kong and Jinghua Li and Baocai Yin},
  booktitle={International Conference on Digital Home (ICDH)},
  year={2022}
}

@inproceedings{gupta2019lvis,
  title={LVIS: A Dataset for Large Vocabulary Instance Segmentation},
  author={Gupta, Agrim and Dollar, Piotr and Girshick, Ross and He, Kaiming and Dollar, Piotr and Girshick, Ross},
  booktitle=CVPR,
  year={2019}
}

@inproceedings{li2023locate,
  title={Locate: Localize and transfer object parts for weakly supervised affordance grounding},
  author={Li, Gen and Jampani, Varun and Sun, Deqing and Sevilla-Lara, Laura},
  booktitle=CVPR,
  year={2023}
}

@inproceedings{lin2014coco,
  title={Microsoft COCO: Common Objects in Context},
  author={Lin, Tsung-Yi and Maire, Michael and Belongie, Serge and Hays, James and Perona, Pietro and Ramanan, Deva and Doll{\'a}r, Piotr and Zitnick, C Lawrence},
  booktitle=ECCV,
  year={2014}
}

@inproceedings{narayanan2015task,
  title={Task-oriented planning for manipulating articulated mechanisms under model uncertainty},
  author={Narayanan, Venkatraman and Likhachev, Maxim},
  booktitle=ICRA,
  year={2015}
}

@article{song2015task,
  title={Task-based robot grasp planning using probabilistic inference},
  author={Song, Dan and Ek, Carl Henrik and Huebner, Kai and Kragic, Danica},
  journal=TRO,
  volume={31},
  number={3},
  pages={546--561},
  year={2015},
  publisher={IEEE}
}

@article{zhu2023toward,
  title={Toward human-like grasp: Functional grasp by dexterous robotic hand via object-hand semantic representation},
  author={Zhu, Tianqiang and Wu, Rina and Hang, Jinglue and Lin, Xiangbo and Sun, Yi},
  journal=PAMI,
  volume={45},
  number={10},
  pages={12521--12534},
  year={2023},
  publisher={IEEE}
}

@inproceedings{miller2003automatic,
  title={Automatic grasp planning using shape primitives},
  author={Miller, Andrew T and Knoop, Steffen and Christensen, Henrik I and Allen, Peter K},
  booktitle=ICRA,
  year={2003}
}

@article{xia2022review,
  title={A review on sensory perception for dexterous robotic manipulation},
  author={Xia, Ziwei and Deng, Zhen and Fang, Bin and Yang, Yiyong and Sun, Fuchun},
  journal=IJRR,
  volume={19},
  number={2},
  pages={17298806221095974},
  year={2022},
  publisher={SAGE Publications Sage UK: London, England}
}

@inproceedings{vezzani2017grasping,
  title={A grasping approach based on superquadric models},
  author={Vezzani, Giulia and Pattacini, Ugo and Natale, Lorenzo},
  booktitle=ICRA,
  year={2017}
}

@inproceedings{agarwal2023dexterous,
  title={Dexterous functional grasping},
  author={Agarwal, Ananye and Uppal, Shagun and Shaw, Kenneth and Pathak, Deepak},
  booktitle=CoRL,
  year={2023}
}

@inproceedings{gong2023arnold,
  title={ARNOLD: A benchmark for language-grounded task learning with continuous states in realistic 3D scenes},
  author={Gong, Ran and Huang, Jiangyong and Zhao, Yizhou and Geng, Haoran and Gao, Xiaofeng and Wu, Qingyang and Ai, Wensi and Zhou, Ziheng and Terzopoulos, Demetri and Zhu, Song-Chun and others},
  booktitle=ICCV,
  year={2023}
}

@article{james2020rlbench,
  title={Rlbench: The robot learning benchmark \& learning environment},
  author={James, Stephen and Ma, Zicong and Arrojo, David Rovick and Davison, Andrew J},
  journal=RA-L,
  volume={5},
  number={2},
  pages={3019--3026},
  year={2020},
  publisher={IEEE}
}

@inproceedings{sawatzky2019object,
  title={What object should i use?-task driven object detection},
  author={Sawatzky, Johann and Souri, Yaser and Grund, Christian and Gall, Jurgen},
  booktitle=CVPR,
  year={2019}
}

@inproceedings{zhou2017scene,
  title={Scene parsing through ade20k dataset},
  author={Zhou, Bolei and Zhao, Hang and Puig, Xavier and Fidler, Sanja and Barriuso, Adela and Torralba, Antonio},
  booktitle=CVPR,
  year={2017}
}

@article{lu2022phrase,
  title={Phrase-based affordance detection via cyclic bilateral interaction},
  author={Lu, Liangsheng and Zhai, Wei and Luo, Hongchen and Kang, Yu and Cao, Yang},
  journal={IEEE Transactions on Artificial Intelligence},
  volume={4},
  number={5},
  pages={1186--1198},
  year={2022},
  publisher={IEEE}
}

@inproceedings{fang2020graspnet,
  title={Graspnet-1billion: A large-scale benchmark for general object grasping},
  author={Fang, Hao-Shu and Wang, Chenxi and Gou, Minghao and Lu, Cewu},
  booktitle=CVPR,
  year={2020}
}

@inproceedings{myers2015affordance,
  title={Affordance detection of tool parts from geometric features},
  author={Myers, Austin and Teo, Ching L and Ferm{\"u}ller, Cornelia and Aloimonos, Yiannis},
  booktitle=ICRA,
  year={2015}
}

@article{montesano2008learning,
  title={Learning object affordances: from sensory--motor coordination to imitation},
  author={Montesano, Luis and Lopes, Manuel and Bernardino, Alexandre and Santos-Victor, Jos{\'e}},
  journal=TRO,
  volume={24},
  number={1},
  pages={15--26},
  year={2008},
  publisher={IEEE}
}

@article{hassanin2021visual,
  title={Visual affordance and function understanding: A survey},
  author={Hassanin, Mohammed and Khan, Salman and Tahtali, Murat},
  journal={ACM Computing Surveys (CSUR)},
  volume={54},
  number={3},
  pages={1--35},
  year={2021},
  publisher={ACM New York, NY, USA}
}

@inproceedings{qu2024rio,
  title={RIO: A benchmark for reasoning intention-oriented objects in open environments},
  author={Qu, Mengxue and Wu, Yu and Liu, Wu and Liang, Xiaodan and Song, Jingkuan and Zhao, Yao and Wei, Yunchao},
  booktitle=NIPS,
  year={2024}
}

@inproceedings{li2022blip,
  title={Blip: Bootstrapping language-image pre-training for unified vision-language understanding and generation},
  author={Li, Junnan and Li, Dongxu and Xiong, Caiming and Hoi, Steven},
  booktitle=ICML,
  year={2022}
}

@inproceedings{zhang2023simple,
  title={A simple framework for open-vocabulary segmentation and detection},
  author={Zhang, Hao and Li, Feng and Zou, Xueyan and Liu, Shilong and Li, Chunyuan and Yang, Jianwei and Zhang, Lei},
  booktitle=ICCV,
  year={2023}
}

@inproceedings{lin2023sphinx,
  title = {SPHINX: A Mixer of Weights, Visual Embeddings and Image Scales for Multi-modal Large Language Models},
  author = {Lin, Ziyi and Liu, Dongyang and Zhang, Renrui and Gao, Peng and Qiu, Longtian and Xiao, Han and Qiu, Han and Shao, Wenqi and Chen, Keqin and Han, Jiaming and Huang, Siyuan and Zhang, Yichi and He, Xuming and Qiao, Yu and Li, Hongsheng},
  booktitle=ECCV,
  year={2024}
}

@inproceedings{zhang2024recognize,
  title={Recognize anything: A strong image tagging model},
  author={Zhang, Youcai and Huang, Xinyu and Ma, Jinyu and Li, Zhaoyang and Luo, Zhaochuan and Xie, Yanchun and Qin, Yuzhuo and Luo, Tong and Li, Yaqian and Liu, Shilong and others},
  booktitle=CVPR,
  year={2024}
}

@article{liu2023grounding,
  title={Grounding dino: Marrying dino with grounded pre-training for open-set object detection},
  author={Liu, Shilong and Zeng, Zhaoyang and Ren, Tianhe and Li, Feng and Zhang, Hao and Yang, Jie and Li, Chunyuan and Yang, Jianwei and Su, Hang and Zhu, Jun and others},
  journal={arXiv preprint arXiv:2303.05499},
  year={2023}
}

@article{liu2024moka,
      title={MOKA: Open-World Robotic Manipulation through Mark-Based Visual Prompting},
      author={Kuan Fang and Fangchen Liu and Pieter Abbeel and Sergey Levine},
      journal=RSS,
      year={2024}
  }

@inproceedings{deitke2023objaverse,
  title={Objaverse: A universe of annotated 3d objects},
  author={Deitke, Matt and Schwenk, Dustin and Salvador, Jordi and Weihs, Luca and Michel, Oscar and VanderBilt, Eli and Schmidt, Ludwig and Ehsani, Kiana and Kembhavi, Aniruddha and Farhadi, Ali},
  booktitle=CVPR,
  year={2023}
}

@inproceedings{li2023behavior,
  title={Behavior-1k: A benchmark for embodied ai with 1,000 everyday activities and realistic simulation},
  author={Li, Chengshu and Zhang, Ruohan and Wong, Josiah and Gokmen, Cem and Srivastava, Sanjana and Mart{\'\i}n-Mart{\'\i}n, Roberto and Wang, Chen and Levine, Gabrael and Lingelbach, Michael and Sun, Jiankai and others},
  booktitle=CoRL,
  year={2023}
}

@inproceedings{li2022grounded,
  title={Grounded language-image pre-training},
  author={Li, Liunian Harold and Zhang, Pengchuan and Zhang, Haotian and Yang, Jianwei and Li, Chunyuan and Zhong, Yiwu and Wang, Lijuan and Yuan, Lu and Zhang, Lei and Hwang, Jenq-Neng and others},
  booktitle=CVPR,
  year={2022}
}

@inproceedings{jia2021scaling,
  title={Scaling up visual and vision-language representation learning with noisy text supervision},
  author={Jia, Chao and Yang, Yinfei and Xia, Ye and Chen, Yi-Ting and Parekh, Zarana and Pham, Hieu and Le, Quoc and Sung, Yun-Hsuan and Li, Zhen and Duerig, Tom},
  booktitle=ICML,
  year={2021}
}

@book{marr2010vision,
  title={Vision: A computational investigation into the human representation and processing of visual information},
  author={Marr, David},
  year={2010},
  publisher={MIT press}
}

@article{zhang2022understanding,
  title={Understanding Physical Effects for Effective Tool-use},
  author={Zhang, Zeyu and Jiao, Ziyuan and Wang, Weiqi and Zhu, Yixin and Zhu, Song-Chun and Liu, Hangxin},
  journal=RA-L,
  volume={7},
  number={4},
  pages={9469--9476},
  year={2022},
  publisher={IEEE}
}

@article{qin2023robot,
  title={Robot tool use: A survey},
  author={Qin, Meiying and Brawer, Jake and Scassellati, Brian},
  journal={Frontiers in Robotics and AI},
  volume={9},
  pages={1009488},
  year={2023},
  publisher={Frontiers Media SA}
}

@article{allen2020rapid,
  title={Rapid trial-and-error learning with simulation supports flexible tool use and physical reasoning},
  author={Allen, Kelsey R and Smith, Kevin A and Tenenbaum, Joshua B},
  journal=PNAS,
  volume={117},
  number={47},
  pages={29302--29310},
  year={2020},
  publisher={National Acad Sciences}
}

@article{levine2016end,
  title={End-to-end training of deep visuomotor policies},
  author={Levine, Sergey and Finn, Chelsea and Darrell, Trevor and Abbeel, Pieter},
  journal=JMLR,
  volume={17},
  number={39},
  pages={1--40},
  year={2016}
}

@inproceedings{nguyen2016detecting,
  title={Detecting object affordances with convolutional neural networks},
  author={Nguyen, Anh and Kanoulas, Dimitrios and Caldwell, Darwin G and Tsagarakis, Nikos G},
  booktitle=IROS,
  year={2016}
}

@inproceedings{gupta2017cognitive,
  title={Cognitive mapping and planning for visual navigation},
  author={Gupta, Saurabh and Davidson, James and Levine, Sergey and Sukthankar, Rahul and Malik, Jitendra},
  booktitle=CVPR,
  year={2017}
}

@inproceedings{li2020oscar,
  title={Oscar: Object-semantics aligned pre-training for vision-language tasks},
  author={Li, Xiujun and Yin, Xi and Li, Chunyuan and Zhang, Pengchuan and Hu, Xiaowei and Zhang, Lei and Wang, Lijuan and Hu, Houdong and Dong, Li and Wei, Furu and others},
  booktitle=ECCV,
  year={2020}
}

@article{vaesen2012cognitive,
  title={The cognitive bases of human tool use},
  author={Vaesen, Krist},
  journal={Behavioral and Brain Sciences},
  volume={35},
  number={4},
  pages={203--218},
  year={2012},
  publisher={Cambridge University Press}
}

@book{mccormack2011tool,
  title={Tool use and causal cognition},
  author={McCormack, Teresa and Hoerl, Christoph and Butterfill, Stephen},
  year={2011},
  publisher={Oxford University Press}
}

@article{suder2023power,
  title={Power Requirements Evaluation of Embedded Devices for Real-Time Video Line Detection},
  author={Suder, Jakub and Podbucki, Kacper and Marciniak, Tomasz},
  journal={Energies},
  volume={16},
  number={18},
  pages={6677},
  year={2023},
  publisher={MDPI}
}

@article{achiam2023gpt,
  title={Gpt-4 technical report},
  author={Achiam, Josh and Adler, Steven and Agarwal, Sandhini and Ahmad, Lama and Akkaya, Ilge and Aleman, Florencia Leoni and Almeida, Diogo and Altenschmidt, Janko and Altman, Sam and Anadkat, Shyamal and others},
  journal={arXiv preprint arXiv:2303.08774},
  year={2023}
}

@incollection{lera2017cybersecurity,
  title={Cybersecurity of robotics and autonomous systems: Privacy and safety},
  author={Lera, Francisco J Rodr{\'\i}guez and Llamas, Camino Fern{\'a}ndez and Guerrero, {\'A}ngel Manuel and Olivera, Vicente Matell{\'a}n},
  booktitle={Robotics-legal, ethical and socioeconomic impacts},
  year={2017},
  publisher={InTech London, UK}
}

@incollection{mseer2023artificial,
  title={Artificial Intelligence and Security Challenges},
  author={Mseer, Ismail Noori and Ahmed, Syed Muqtar},
  booktitle={From Industry 4.0 to Industry 5.0: Mapping the Transitions},
  pages={49--55},
  year={2023},
  publisher={Springer}
}

@article{wang2023large,
  title={Large-scale multi-modal pre-trained models: A comprehensive survey},
  author={Wang, Xiao and Chen, Guangyao and Qian, Guangwu and Gao, Pengcheng and Wei, Xiao-Yong and Wang, Yaowei and Tian, Yonghong and Gao, Wen},
  journal={Machine Intelligence Research},
  volume={20},
  number={4},
  pages={447--482},
  year={2023},
  publisher={Springer}
}

@inproceedings{wu2024learning,
  title={Learning environment-aware affordance for 3d articulated object manipulation under occlusions},
  author={Wu, Ruihai and Cheng, Kai and Zhao, Yan and Ning, Chuanruo and Zhan, Guanqi and Dong, Hao},
  booktitle=NIPS,
  year={2024}
}

@article{friedman1997bayesian,
  title={Bayesian network classifiers},
  author={Friedman, Nir and Geiger, Dan and Goldszmidt, Moises},
  journal={Machine Learning},
  volume={29},
  pages={131--163},
  year={1997},
  publisher={Springer}
}

@article{noble2006support,
  title={What is a support vector machine?},
  author={Noble, William S},
  journal={Nature Biotechnology},
  volume={24},
  number={12},
  pages={1565--1567},
  year={2006},
  publisher={Nature Publishing Group UK London}
}

@inproceedings{lopes2007affordance,
  title={Affordance-based imitation learning in robots},
  author={Lopes, Manuel and Melo, Francisco S and Montesano, Luis},
  booktitle=IROS,
  year={2007}
}

@article{uugur2010traversability,
  title={Traversability: A case study for learning and perceiving affordances in robots},
  author={U{\u{g}}ur, Emre and {\c{S}}ahin, Erol},
  journal={Adaptive Behavior},
  volume={18},
  number={3-4},
  pages={258--284},
  year={2010},
  publisher={SAGE Publications Sage UK: London, England}
}

@article{chen2023survey,
  title={A survey of visual affordance recognition based on deep learning},
  author={Chen, Dongpan and Kong, Dehui and Li, Jinghua and Wang, Shaofan and Yin, Baocai},
  journal={IEEE Transactions on Big Data},
  year={2023},
  publisher={IEEE}
}

@inproceedings{nguyen2023open,
  title={Open-vocabulary affordance detection in 3d point clouds},
  author={Nguyen, Toan and Vu, Minh Nhat and Vuong, An and Nguyen, Dzung and Vo, Thieu and Le, Ngan and Nguyen, Anh},
  booktitle=IROS,
  year={2023}
}

@inproceedings{van2024open,
  title={Open-vocabulary affordance detection using knowledge distillation and text-point correlation},
  author={Van Vo, Tuan and Vu, Minh Nhat and Huang, Baoru and Nguyen, Toan and Le, Ngan and Vo, Thieu and Nguyen, Anh},
  booktitle=ICRA,
  year={2024}
}

@article{kuhn1955hungarian,
  title={The Hungarian method for the assignment problem},
  author={Kuhn, Harold W},
  journal={Naval Research Logistics Quarterly},
  volume={2},
  number={1-2},
  pages={83--97},
  year={1955},
  publisher={Wiley Online Library}
}

@inproceedings{rezatofighi2019generalized,
  title={Generalized intersection over union: A metric and a loss for bounding box regression},
  author={Rezatofighi, Hamid and Tsoi, Nathan and Gwak, JunYoung and Sadeghian, Amir and Reid, Ian and Savarese, Silvio},
  booktitle=CVPR,
  year={2019}
}

@inproceedings{milletari2016v,
  title={V-net: Fully convolutional neural networks for volumetric medical image segmentation},
  author={Milletari, Fausto and Navab, Nassir and Ahmadi, Seyed-Ahmad},
  booktitle=ThreeDV,
  year={2016}
}

@inproceedings{lin2017focal,
  title={Focal loss for dense object detection},
  author={Lin, Tsung-Yi and Goyal, Priya and Girshick, Ross and He, Kaiming and Doll{\'a}r, Piotr},
  booktitle=ICCV,
  year={2017}
}

@inproceedings{chuang2018learning,
  title={Learning to act properly: Predicting and explaining affordances from images},
  author={Chuang, Ching-Yao and Li, Jiaman and Torralba, Antonio and Fidler, Sanja},
  booktitle=CVPR,
  year={2018}
}

@article{liu2019roberta,
  title={Roberta: A robustly optimized bert pretraining approach},
  author={Liu, Yinhan},
  journal={arXiv preprint arXiv:1907.11692},
  volume={364},
  year={2019}
}

@inproceedings{he2016deep,
  title={Deep residual learning for image recognition},
  author={He, Kaiming and Zhang, Xiangyu and Ren, Shaoqing and Sun, Jian},
  booktitle=CVPR,
  year={2016}
}

@article{ren2016faster,
  title={Faster R-CNN: Towards real-time object detection with region proposal networks},
  author={Ren, Shaoqing and He, Kaiming and Girshick, Ross and Sun, Jian},
  journal=PAMI,
  volume={39},
  number={6},
  pages={1137--1149},
  year={2016},
  publisher={IEEE}
}

@inproceedings{li2022exploring,
  title={Exploring plain vision transformer backbones for object detection},
  author={Li, Yanghao and Mao, Hanzi and Girshick, Ross and He, Kaiming},
  booktitle=ECCV,
  year={2022}
}

@inproceedings{lu2025geal,
  title={Geal: Generalizable 3d affordance learning with cross-modal consistency},
  author={Lu, Dongyue and Kong, Lingdong and Huang, Tianxin and Lee, Gim Hee},
  booktitle=CVPR,
  year={2025}
}

@inproceedings{chu20253d,
  title={3d-affordancellm: Harnessing large language models for open-vocabulary affordance detection in 3d worlds},
  author={Chu, Hengshuo and Deng, Xiang and Lv, Qi and Chen, Xiaoyang and Li, Yinchuan and Hao, Jianye and Nie, Liqiang},
  booktitle=ICLR,
  year={2025}
}

@inproceedings{li2024one,
  title={One-shot open affordance learning with foundation models},
  author={Li, Gen and Sun, Deqing and Sevilla-Lara, Laura and Jampani, Varun},
  booktitle=CVPR,
  year={2024}
}

@string {PNAS = "PNAS"}

@string {PAMI = "TPAMI"}

@string {IJCV = "IJCV"}

@string {CVPR = "CVPR"}

@string {ICCV = "ICCV"}

@string {ECCV = "ECCV"}

@string {JMLR = "JMLR"}

@string {NIPS = "NeurIPS"}

@string {ICML = "ICML"}

@string {ICLR = "ICLR"}

@string {IJCAI = "IJCAI"}

@string {IJRR = "IJRR"}

@string {TRO = "T-RO"}

@string {RA-L = "RA-L"}

@string {IROS = "IROS"}

@string {ICRA = "ICRA"}

@string {RSS = "RSS"}

@string {CoRL = "CoRL"}

@string {ThreeDV = "3DV"}

@string {CHI = "CHI"}
\bibliographystyle{ieeetr}
\setstretch{1}

\vspace{-8pt}
\vskip -3\baselineskip plus -1fil
\begin{IEEEbiography}[{\raisebox{2.2\baselineskip}{\includegraphics[width=1in,height=1.25in,clip,keepaspectratio]{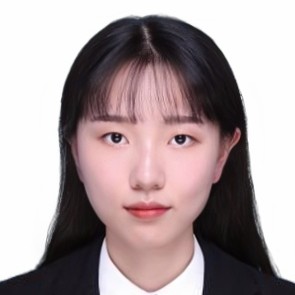}}}]{Xiaomeng Zhu}
is a Ph.D. student in the Department of Computer Science and Engineering at the Hong Kong University of Science and Technology. She received her B.Eng. degree from the University of Electronic Science and Technology of China and her M.S. degree from the Institute of Automation, Chinese Academy of Sciences. Her research interests include robotic perception, human intent understanding, and human-robot collaboration.
\end{IEEEbiography}

\vskip -3\baselineskip plus -1fil
\begin{IEEEbiography}[{\raisebox{2.2\baselineskip}{\includegraphics[width=1in,height=1.25in,clip,keepaspectratio]{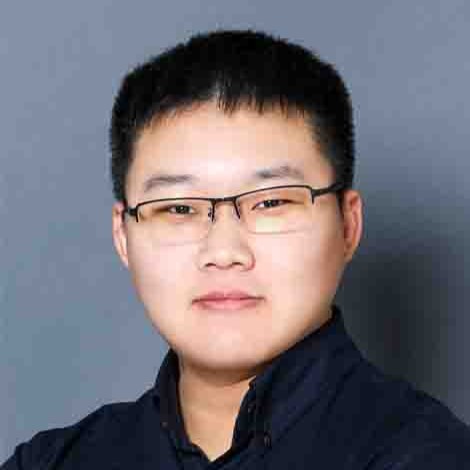}}}]{Yuyang Li} is a Ph.D. candidate at the Institute for AI, Peking University. He received his bachelor's degree in engineering from the Department of Automation at Tsinghua University in 2024. He aspires to advance the development of embodied intelligence with versatile dexterity based on multi-modal perception.
\end{IEEEbiography}

\vskip -3\baselineskip plus -1fil
\begin{IEEEbiography}[{\raisebox{2.2\baselineskip}{\includegraphics[width=1in,height=1.25in,clip,keepaspectratio]{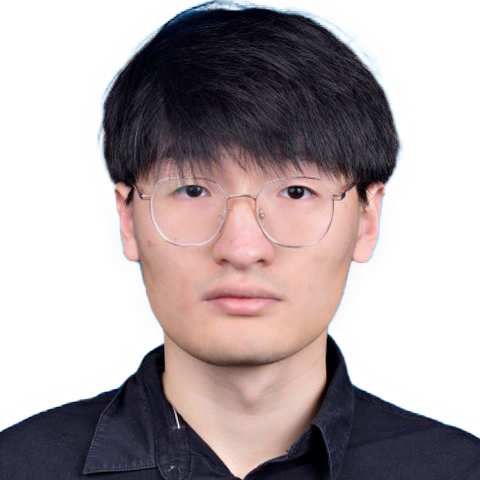}}}]{Leiyao Cui} received his B.Eng. degree from the school of Information and Electronics at the Beijing Institute of Technology in 2024. Currently, he is a Ph.D. student at the Shenyang Institute of Automation, Chinese Academy of Sciences. He is also a research intern at the Institute for AI, Peking University. His research interest currently focuses on robot perception.
\end{IEEEbiography}

\vskip -3\baselineskip plus -1fil
\begin{IEEEbiography}[{\raisebox{2.2\baselineskip}{\includegraphics[width=1in,height=1.25in,clip,keepaspectratio]{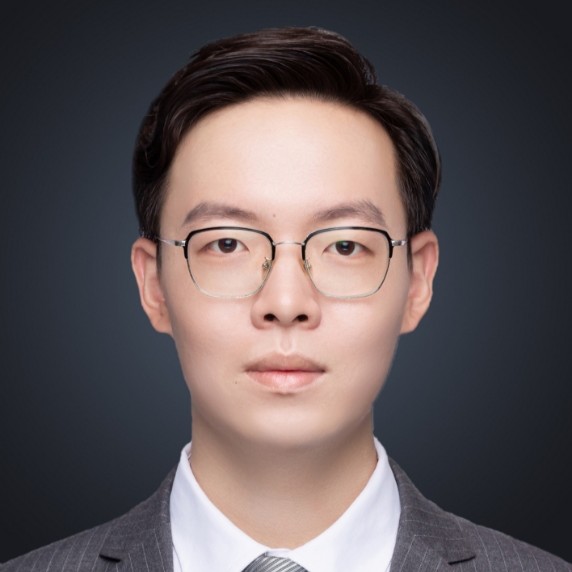}}}]{Pengfei Li} is a fourth-year Ph.D. student at the Institute for AI Industry Research, Tsinghua University. He received the B.E. degree in Computer Science and Technology from University of Chinese Academy of Sciences, Beijing, China, in 2022. His research interests include autonomous driving and robotics.
\end{IEEEbiography}

\vskip -3\baselineskip plus -1fil
\begin{IEEEbiography}[{\raisebox{2.2\baselineskip}{\includegraphics[width=1in,height=1.25in,clip,keepaspectratio]{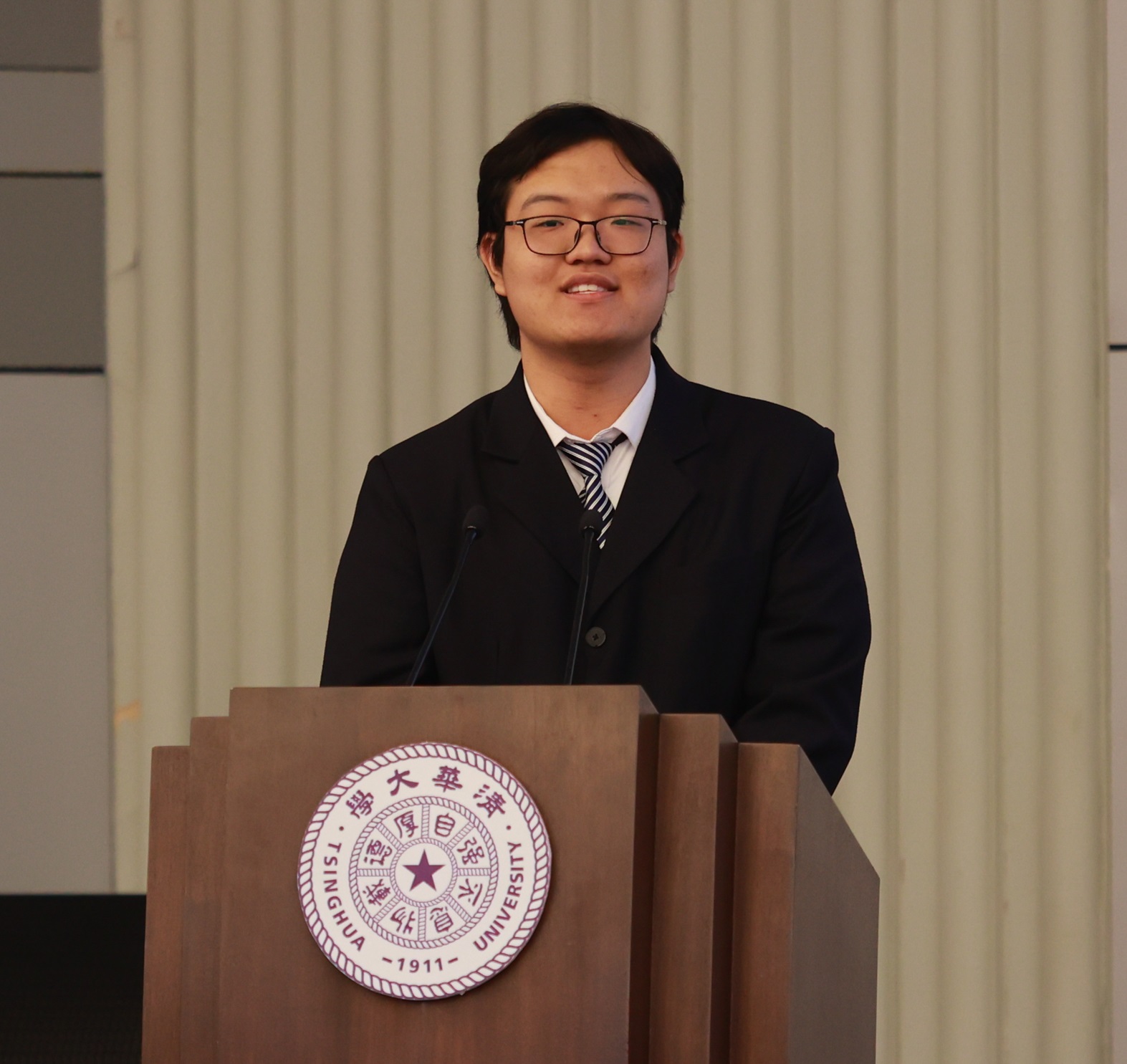}}}]{Huan-ang Gao} is a second-year PhD student at the Department of Computer Science and Technology and the Institute for AI Industry Research at Tsinghua University. He earned his bachelor's degree in engineering from the same department at Tsinghua University in 2024. His research focuses on developing generative simulation methods for evaluating and training embodied AI policies.
\end{IEEEbiography}

\vskip -3\baselineskip plus -1fil
\begin{IEEEbiography}[{\raisebox{2.2\baselineskip}{\includegraphics[width=1in,height=1.25in,clip,keepaspectratio]{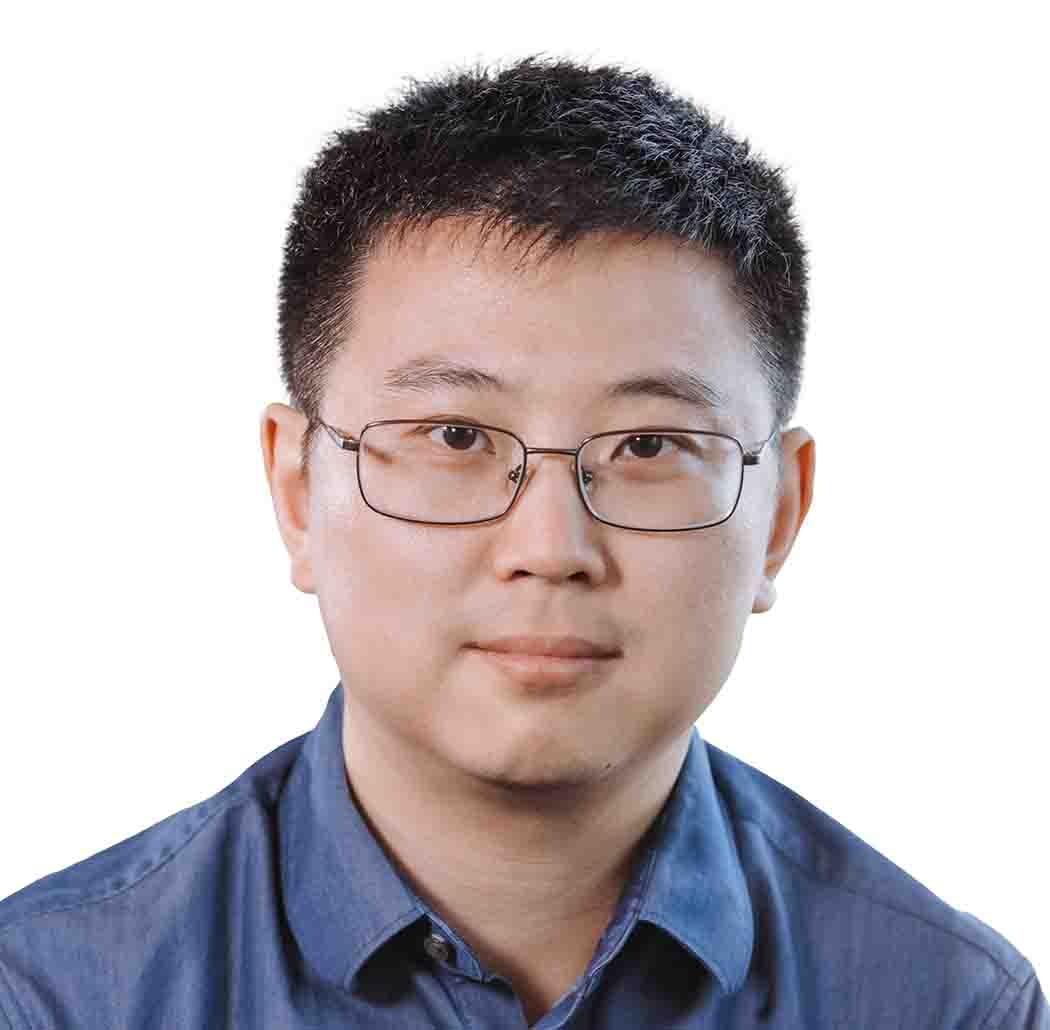}}}]{Yixin Zhu} received the Ph.D. degree in statistics from the University of California, Los Angeles (UCLA), in 2018. He is currently an Assistant Professor with Peking University, Beijing, China, jointly appointed in the School of Psychological and Cognitive Sciences and the Institute for Artificial Intelligence. His research interests include cognitive reasoning, embodied intelligence, computer vision, human-centric AI, and BCI, with a particular emphasis on integrating commonsense knowledge with multimodal perception to build interactive intelligent systems. He has published in Science, Nature Machine Intelligence, Nature Human Behaviour, Science Robotics, and Science Advances, among others.
\end{IEEEbiography}

\vskip -2.5\baselineskip plus -1fil
\begin{IEEEbiography}[{\raisebox{2.2\baselineskip}{\includegraphics[width=1in,height=1.25in,clip,keepaspectratio]{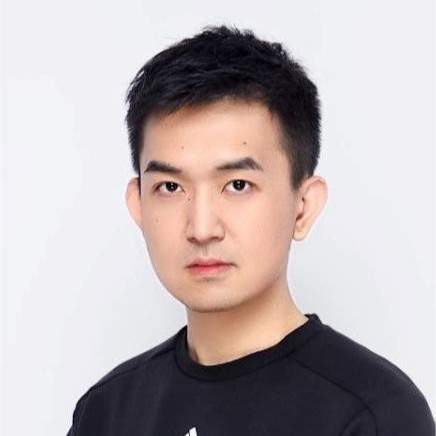}}}]{Hao Zhao} received the B.E. degree and the Ph.D. degree both from the EE department of Tsinghua University, Beijing, China. He is currently an Assistant Professor with the Institute for AI Industry Research (AIR), Tsinghua University. He was a research scientist at Intel Labs China and a joint postdoc affiliated to Peking University. His research interests cover various computer vision topics related to robotics, especially 3D scene understanding. His work has been recognized with the Best Student Paper Award at CVPR 2026 and the Best Paper Award at 3DV 2024, finalist selections at ICRA 2026 and RSS 2026, and a runner-up award at CICAI 2023.
\end{IEEEbiography}

\end{document}